\pdfoutput=1
\documentclass[11pt]{article}

\usepackage[final]{acl}

\usepackage{times}
\usepackage{latexsym}

\usepackage[T1]{fontenc}
\usepackage[utf8]{inputenc}

\usepackage{microtype}

\usepackage{inconsolata}

\usepackage{graphicx}

\usepackage{supertabular}

\newcommand{\Title}{InstructPart}

\usepackage{graphicx}
\usepackage{amsmath}
\usepackage{amssymb}
\usepackage{booktabs}

\usepackage{afterpage}
\usepackage{graphicx}
\usepackage{multirow}
\usepackage{tabularx}
\usepackage{booktabs}
\usepackage{colortbl}
\usepackage{amsfonts}
\usepackage{amssymb}
\usepackage{bbding}
\usepackage{xcolor}
\usepackage{adjustbox}
\usepackage{listings}
\usepackage{longtable}
\usepackage{afterpage}
\usepackage{stfloats}
\usepackage{supertabular}

\title{\Title{}: Task-Oriented Part Segmentation with Instruction Reasoning}

\author{
    \textbf{Zifu Wan}\quad \textbf{Yaqi Xie}\quad \textbf{Ce Zhang}\quad \textbf{Zhiqiu Lin}\quad \textbf{Zihan Wang}\vspace{0mm}\\
      \textbf{Simon Stepputtis}\quad \textbf{Deva Ramanan}\quad \textbf{Katia Sycara} \vspace{1mm}\\
    Robotics Institute, Carnegie Mellon University \\
 \small{\texttt{\{zifuw, yaqix, cezhang, zhiqiul, zihanwa3, sstepput, deva, sycara\}@andrew.cmu.edu
 }} \vspace{1mm}
 \\ 
  \url{https://zifuwan.github.io/InstructPart/}
}

\begin{document}
\maketitle

% \begin{abstract}
% Recent advancements in Vision-Language Models (VLMs) have led to their increased application in robotic tasks. 
% While the implementation of VLMs is primarily at the object level, the distinct affordances of an object's various parts — such as a knife's blade for cutting versus its handle for grasping — remain a challenge for current state-of-the-art models. 
% Our investigations reveal that these models often fail to accurately segment parts based on task instructions, a capability crucial for precise robotic interactions.
% Addressing the lack of real-world datasets to evaluate these fine-grained tasks, we introduce a comprehensive dataset that includes image observations, task descriptions, and precise annotations for object-part interactions, complemented by part segmentation masks.
% We present an evaluation of common pre-trained VLMs using this benchmark, shedding light on the models' performance in understanding and executing part-level tasks within everyday contexts.
% \end{abstract}

\begin{abstract}
%%%%%% ECCV version
% Large multimodal foundation models, particularly in the domain of language and vision, have greatly accelerated many use cases, including robotics, autonomous driving, information retrieval, and grounding. 
% However, many of these models perceive objects as indivisible, neglecting the components comprising a whole. 
% Yet, knowing about the various components and their associated affordances provides valuable insights into the utility of an object, particularly in the domain of instruction following.
% In this work, we propose a novel real-world benchmark consisting of hand-labeled object-part images and two associated tasks: a) identifying an object part given a task-oriented instruction and b) identifying an object part given a part query.  
% Through our experiments, we show that object-part segmentation remains a challenging task even for recent state-of-the-art Vision-Language Models (VLM). 
% In addition to our benchmark, we also introduce PISA -- \textit{Part Identification and Segmentation Assistant} -- an improvement of a current SOTA VLM that provides a strong baseline for future research on part-segmentation.
% With our dataset, associated benchmark, and baseline, we hope to provide a valuable resource for the vision community, advancing research in many domains, including robotics, virtual reality, information retrieval and other related disciplines. 
%%%%%%
%%%%%% Neurips Version
Large multimodal foundation models, particularly in the domains of language and vision, have significantly advanced various tasks, including robotics, autonomous driving, information retrieval, and grounding.  
However, many of these models perceive objects as indivisible, overlooking the components that constitute them.  
Understanding these components and their associated affordances provides valuable insights into an object's functionality, which is fundamental for performing a wide range of tasks.
In this work, we introduce a novel real-world benchmark, \textit{\Title{}}, comprising hand-labeled part segmentation annotations and task-oriented instructions to evaluate the performance of current models in understanding and executing part-level tasks within everyday contexts.
Through our experiments, we demonstrate that task-oriented part segmentation remains a challenging problem, even for state-of-the-art Vision-Language Models (VLMs).  
In addition to our benchmark, we introduce a simple baseline that achieves a twofold performance improvement through fine-tuning with our dataset.
With our dataset and benchmark, we aim to facilitate research on task-oriented part segmentation and enhance the applicability of VLMs across various domains, including robotics, virtual reality, information retrieval, and other related fields.
% Project website: \href{https://zifuwan.github.io/InstructPart/}{https://zifuwan.github.io/InstructPart/}.
\end{abstract}    
\section{Introduction}
\label{sec:intro}
% Give a motivating example from our daily life.\\
% Identify our task, natural language instruction + visual observation to part segments.\\
% Why this task is important. e.g., tie to robotics, home service robot, general intelligent agent.\\
% Why this task is difficult. e.g., existing works are in controlled environment, need finetuning, senstive to environment difference, lighting conditions...\\

% \textbf{Existing Literature:}
% - Existing dataset (emphasize on their limitations, small, animals, poor annotation...)\\
% - Existing methods (briefly discuss), show that this task is challenging for all existing methods

% \textbf{Our contributed dataset.}

% \textbf{Summarise our contributions: }\\
% - First instruction to part seg dataset\\
% - Evaluated on various models\\
% - ?

% }

% \yaqi{
% \begin{enumerate}
% \item Weaken oracle referring. Make it to be an auxiliary task.
% \item Ablation studies are unclear. Motivate them better and give them more precise tiles. Current ones are confusing. (something like: the first one - Models can't segment the part, can it segment the object itself? The second one - what is reasonableness, you need to define it. And why are you interested in this?)
% \item Results insights (main takeaways, design principles).
% \item Change task order in fig2. 
% \item Make fig3 shorter.
% \item Match the sections in related works with baseline models. Why reasoning segmentation methods are not discussed in the referring segmentation section?
% \end{enumerate}
% }

%%% Begin Simon's Stuff 

\begin{figure}[ht]
  \begin{center}
     \includegraphics[width=\linewidth]{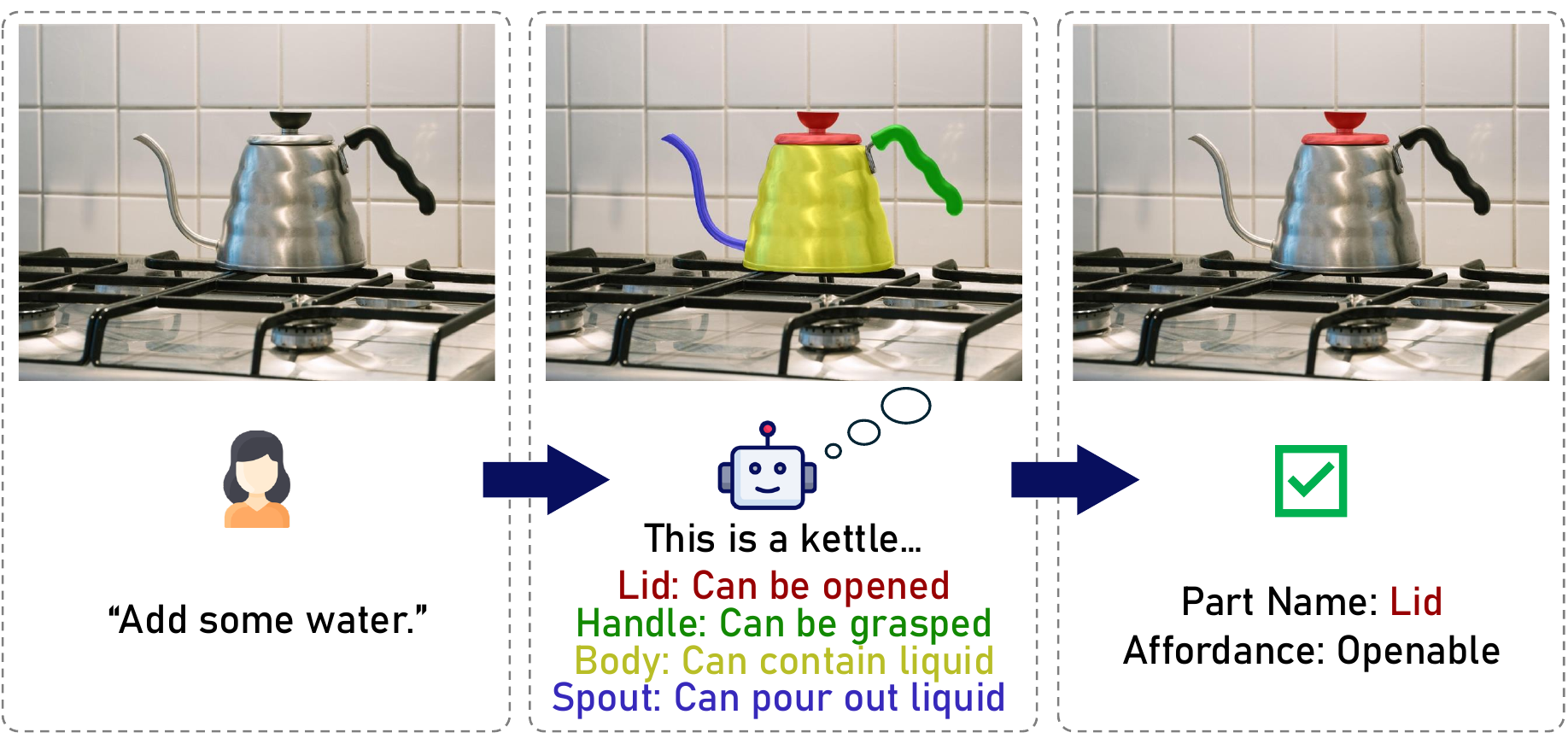}
  \end{center}
  \vspace{-10pt}
  \caption{
 The task-oriented part segmentation task: Presented with an image observation (left) and a corresponding task to add some water, the system is required to reason about specific parts to fulfill the task. 
  % Our InstructPart dataset comprises 700 images, each annotated with natural language instructions and corresponding ground truth part segmentations. This dataset presents a challenging visual understanding task for current vision-language models. \yaqi{change the caption as Katia suggestted.}
  }
  \label{fig:task illunstration}
  \vspace{-10pt}
\end{figure}
%%%%%%%%%%%%%%%%%%%%%%%%%%%%%%%%%%%%%%%%%%%%%%%%%%%%%%%%%%%
Large Vision-Language Models (LVLMs)~\cite{radford2021learning, alayrac2022flamingo, you2023ferret} have been extensively utilized across various domains, such as robotics~\cite{driess2023palm}, autonomous driving~\cite{zhou2023vision,wan2025sigma}, medical imaging~\cite{han2023multimodal}, and information retrieval~\cite{liu2021image}, owing to their strong language reasoning and perceptual capabilities.
In these cases, LVLMs are primarily employed for language grounding, enabling the identification of visual targets within a scene based on associated language descriptions.
By leveraging large datasets composed of image-text pairs, LVLMs can map visual content to textual semantic representations~\cite{radford2021learning} within joint embedding spaces. However, while this approach yields powerful models with strong text-image alignment, they often focus on understanding entire objects~\cite{liu2023referring, zou2023segment, zou2023generalized, xu2023side, liang2023open, sun2023going}, overlooking the fact that grounding is not solely about classifying whole objects but also about recognizing fine-grained parts.
As illustrated in Figure~\ref{fig:task illunstration}, given the task of adding water and a visual observation of a kettle, the system must not only identify the entire kettle but also recognize each part of the target and its corresponding affordances before grounding to task-related regions.

To advance task-oriented part segmentation, we believe that establishing a benchmark is essential for the field. However, most large-scale vision datasets primarily focus on object-level understanding~\cite{liu2023referring, zou2023segment, zou2023generalized, xu2023side, liang2023open, sun2023going}, while existing part-level recognition datasets either cover only a limited range of part categories~\cite{nguyen2017object, myers2015affordance, roy2016multi} or are derived from simulations~\cite{geng2023gapartnet, xiang2020sapien, mo2019partnet}. We attribute this primarily to the challenge of annotating part-level labels and task-related descriptions, which is both time-consuming and expensive~\cite{wan2024instructpart}.

To address this challenge, we introduce a new real-world dataset, \textbf{{\Title{}}}, consisting of 2,400 images across 48 object classes and 44 part classes, with hand-labeled segmentation masks, as well as 9,600 hand-labeled task instructions, 2,400 part queries, and 2,400 affordances.
Each image is accompanied by human-annotated and GPT-polished instructions for common household tasks and detailed part segmentation masks.
As part of our benchmark, we propose two distinct tasks: a) Task Reasoning Part Segmentation (TRPS): identifying a particular part given an instruction to fulfill a task, e.g., ``Locate the part meant for pulling to open the microwave''; and b) Oracle Referring Part Segmentation (ORPS): identifying an object part given a part query, e.g., ``handle of the microwave''. 
% We form a thorough benchmark including state-of-the-art vision-language models on both tasks.
Thorough evaluations of current vision-language models on the two tasks reveal a significant deficiency in their ability to comprehend natural language and accurately ground it across diverse objects and parts. This finding highlights the need to address a critical shortcoming in vision-language models for fine-grained segmentation.

% To facilitate the research of task-oriented part segmentation, we collect a dataset, \Title{}, 
% Notably, our dataset does not merely provide a single point on the target segment as in~\cite{Luo_2022_CVPR}, but rather offers a segmentation mask for the entire part in the query.
% We split the dataset into 1,800 images for training and 600 images for evaluation.
% With this dataset and associated benchmark, we hope to provide a valuable resource for the community, fostering the development of more advanced VLMs across various domains. 
% This ability is essential for seamless interaction with humans, especially in unstructured environments such as homes or public spaces.
% By utilizing our dataset, we envision research in robotics, particularly for assistive robots, as well as manipulation tasks, object segmentation, virtual reality, affordance learning, and other domains.
% Particularly, we demonstrate the manipulation usage of \Title{} by evaluating a model trained with \Title{} on real-world grasping data.

Finally, we explore the training potential of our dataset by proposing a simple yet effective baseline, which leads to a nearly 100\% improvement.
With our proposed benchmark, we emphasize the importance of advancing vision-language models to excel not only in object-level understanding but also in discerning fine-grained part-level details.  
By utilizing our dataset, we hope to envision advancements in robotics, particularly for assistive robots, as well as in manipulation tasks, object segmentation, virtual reality, affordance learning, and other related domains. Our contributions can be summarized as follows: 
\begin{itemize}
    \item[$\bullet$] To the best of our knowledge, we introduce the first dataset that bridges task-oriented interactions with part segmentation for common household tasks.
    \item[$\bullet$] We rigorously evaluate various vision-language models on our dataset, revealing their limitations in fine-grained recognition with language reasoning.
    \item[$\bullet$] We fine-tune a simple baseline based on a state-of-the-art model, achieving performance gains of over twofold, highlighting the quality and training potential of our dataset.

    % \item[$\bullet$] We propose the novel Task-Oriented Reasoning Part Segmentation (TRPS) task, highlighting the importance of task completion over simple part recognition.
    % \item[$\bullet$] To the best of our knowledge, we present the largest dataset of hand-labeled real-world images for object-part segmentation and associated task descriptions.
    % \item[$\bullet$] We propose a novel benchmark containing two distinct tasks: a) Task-oriented Reasoning Part Segmentation (TRPS) and b) Oracle Referring Part Segmentation (ORPS).
    % \item[$\bullet$] Through extensive experiments, we demonstrate that current VLMs struggle with such fine-grained segmentation and propose a new method -- PISA -- as a strong baseline for our benchmark task.
    % \item[$\bullet$] By training with our 1,800 samples, PISA gains a performance boost of over twofold.
    % \item[$\bullet$] We collect a real-world robotic platform grasping dataset and evaluate the models trained with our \Title{}, demonstrating \Title{}'s usage for the manipulation aspect.
\end{itemize}

\section{Related Work}
\label{sec:related work}

\begin{table}
\begin{center}
\setlength{\tabcolsep}{1.36mm}{
       \resizebox{\linewidth}{!}{
  \begin{tabular}{l|ccccc}
    \specialrule{.1em}{.05em}{.05em} 
          Dataset & \#Object & \#Part & \#Affordance & \#Action & Instruction\\
          \cline{1-6}
          PartImageNet & 11/158 & 13 & N/A & N/A  & \XSolidBrush\\
            Pascal-Part  & 20 & -- & N/A & N/A  & \XSolidBrush \\
            % Ade20K & 2693 & 476 & ... & \XSolidBrush & ... \\
            PACO & 75 & -- & N/A & N/A  & \XSolidBrush \\
            \hline
            UMD  & 17 & N/A & 7 & N/A & \XSolidBrush  \\
            NYUv2-AFF~  & 40 & N/A & 5 & N/A  & \XSolidBrush\\
            IIT-AFF & 10 & N/A & 9 & N/A  & \XSolidBrush \\
            \hline
            AGD20K$^*$ & 50 & N/A & 36 & N/A  & \XSolidBrush \\
            \cline{1-6}
            \Title{} (Ours) & 48 & 44 & 30 & 37  & \Checkmark\\
    \specialrule{.1em}{.05em}{.05em} 
    \end{tabular}}}
\vspace{-3pt}
\caption{Comparison of relevant part segmentation datasets. We show the number of object classes (\#Object), part classes (\#Part), affordances (\#Affordance), actions (\#Action), and whether instructions are included (Instruction). N/A means there is no such type of data, while -- means the data exists while no relevant information is provided. 11/158 indicates the super-class and sub-class numbers in PartImageNet. $^*$ indicates the dataset only contains point annotations instead of accurate masks for target affordances.}
  \label{tab:dataset_compare}%
  \vspace{-15pt}
\end{center}
\end{table}
\subsection{Part Segmentation}
% \looseness = -1
The problem of segmenting an object into a collection of semantic parts is not a novel problem in it itself.
Prior works mainly utilized fully supervised approaches, which need to be trained on large datsets~\cite{sun2023going}, such as PartImageNet~\cite{he2022partimagenet}, Pascal-Part~\cite{chen2014detect}, ADE20K~\cite{zhou2019semantic}, and PACO~\cite{ramanathan2023paco}.
However, these datasets contain only a limited subset relevant to human-robot interaction (e.g., PartImageNet includes just one related category: bottle), thus restricting their applicability to daily tasks. 
In robotics, part segmentation is used to understand the components of objects and their associated affordances, which are crucial for manipulation tasks~\cite{gadre2021act, yi2018deep}.
% A primary use case for part segmentation is robotics, where understanding the parts of objects and their associated affordances can be used for further manipulation~\cite{gadre2021act, yi2018deep}. 
While many datasets have been created for this domain~\cite{mo2019partnet, xiang2020sapien, geng2023gapartnet}, they are all generated from simulators, which introduces potential challenges when generalizing to real-world scenarios.
% Related datasets include PartNet~\cite{mo2019partnet}, PartNet-Mobility~\cite{xiang2020sapien}, 3D AffordanceNet~\cite{deng20213d}, and GAPartNet~\cite{geng2023gapartnet}. 
%
% However, these datasets are all generated from simulators, thus adding potential challenges when generalizing to real-world problems.
%
To address this issue, real-world affordance datasets such as UMD-Affordance~\cite{myers2015affordance}, NYUv2-Affordance~\cite{roy2016multi}, and IIT-AFF~\cite{nguyen2017object} exist. However, due to the difficulty of collecting large quantities of real-world data, these datasets are limited in the number of affordances they present.
On the other hand, AGD20K~\cite{Luo_2022_CVPR} collects egocentric and exocentric images for affordance learning. However, it only provides sparse point annotations, which can be insufficient for accurate task execution, such as manipulation.
%
% However, these datasets cover a limited set of scenes and contain less than 10 classes of affordance. 
Similarly, Where2Act~\cite{mo2021where2act} extracts actionable information from articulated objects with movable parts but is limited to six action types and a single contact point, which may be sub-optimal.
%
% Besides, using a simple word or phrase can be insufficient to represent affordance sometimes. 
Furthermore, the aforementioned datasets only contain simple word phrases outlining the target; however, full language comprehension is crucial in a human-robot interaction task. 
Understanding language can be ambiguous even for simple objects like a light switch, which can be ``turned on'', ``pressed'' or ``twisted'' depending on the switch's type, and people tend to refer to such objects as parts of larger task descriptions instead of a single word. 
% In real-world situations, people tend to refer to a part using an instructional sentence instead of a single word.  
%
Motivated by this, we construct a comprehensive dataset with task descriptions and object-part classes, as shown in Tab.~\ref{tab:dataset_compare}. 
% We also annotate affordances and actions for future research usage.
%
% A comparison of relevant real-world part segmentation datasets is shown in Tab.~\ref{tab:dataset_compare}.
%
% We hope the dataset can provide more insights into future work about the interaction between robots and the environment.

\subsection{Open-Vocabulary Segmentation} 
% Traditional fully-supervised recognition methods have limited generalization ability, leading to more attention to recognizing open-world classes beyond the training phase. 
% Open-vocabulary segmentation and object detection have become popular in recent years due to the development of foundational models that align general-purpose language with internet-scale image datasets by aligning the latent representations of both modalities with each other, allowing for intricate reasoning even in a zero-shot manner. One such model is CLIP~\cite{radford2021learning}, having accelerated the development of vision-language models. 
Open-vocabulary segmentation aims to perform zero-shot segmentation with the assistance of vision-language foundation models, such as CLIP~\cite{radford2021learning}.
%
% Early works propose the concept of zero-shot learning, which generalizes to similar classes by learning the attribute representations~\cite{romera2015embarrassingly,xian2017zero}. These methods incorporate external knowledge to recognize out-of-distribution classes, such as image-caption pairs~\cite{zareian2021open}, word embeddings~\cite{bucher2019zero, xian2019semantic, han2021contrastive}, etc.
%
% Recently, the pre-trained vision language model, CLIP~\cite{radford2021learning}, which was trained on 400 million image-caption pairs, aligns the gap between embedding space of vision and language and demonstrates valuable potential in open-vocabulary segmentation.
%
% Recently, CLIP has been used as the backbone for complex models, e.g., MaskFormer~\cite{cheng2022masked} or SAM~\cite{kirillov2023segment}, and apply CLIP for classification due to its generalizability and ability to be fine-tuned towards novel tasks.
% The key to these methods is to align the embedding space of the image encoder with that of CLIP.
%
For example,
%, LSeg~\cite{liu2022open} aligns the spaces by maximizing the correlation between the text embedding and the image pixel embedding during training, and OpenSeg~\cite{ghiasi2022scaling} aligns the visual features with text embedding within different regions.
%
OVSeg~\cite{liang2023open} proposes to crop the region proposals and finetune CLIP using a mask prompt tuning mechanism.
%
% FC-CLIP~\cite{yu2023convolutions} uses a frozen convolutional CLIP backbone for open-vocabulary recognition.
%
SAN~\cite{xu2023side} applies a side adapter network to a frozen CLIP to get the class of masks.
Going beyond object-level segmentation, 
VLPart~\cite{sun2023going} performs open-vocabulary part segmentation by parsing the novel object into parts using its semantic correspondence with the base object and classifies it with CLIP.
%
% OPS~\cite{pan2023towards} generates pseudo labels for unlabeled data with a trained object detector and clusters between different parts in a self-supervised procedure.
% %
% However, OPS cannot predict semantic labels for the segments and is not suitable for our task.

Although these open-world recognition methods demonstrate potential in recognizing out-of-distribution classes, they have limited reasoning ability to understand complex instructional sentences, prohibiting their wider usage in daily tasks requiring complex language comprehension.

\subsection{Referring Expression Segmentation}
Referring expression segmentation aims to generate a segmentation mask from a given language expression~\cite{hu2016segmentation}.
% , and various datasets such as ReferIt~\cite{kazemzadeh2014referitgame}, CLEVR-Ref+~\cite{liu2019clevr}, refCOCO~\cite{yu2016modeling}, refCOCOg~\cite{mao2016generation}, gRefCOCO~\cite{liu2023gres} are collected to support the task.
%
% These datasets contain image-expression pairs and the masks of objects being referred to.
%
Popular referring segmentation methods use a visual and a language encoder to extract features from the two modalities respectively, and design attention mechanisms to incorporate the features and assemble classes for region masks~\cite{yang2022lavt, liu2023gres, ouyang23slvit, liu2023referring}.
Recently, more works have applied pre-trained foundation models, e.g., SAM~\cite{kirillov2023segment} and CLIP~\cite{radford2021learning} as the encoder and focused on the design of the decoder, such as X-Decoder~\cite{zou2023generalized} and SEEM~\cite{zou2023segment}.
Furthermore, ManipVQA~\cite{huang2024manipvqa} applies VLMs with manipulation-centric knowledge to detect tools and affordances.
However, the referring expression task only takes short phrases as input and does not consider complex reasoning, for example, when the target name does not appear directly in the expression.

\subsection{Reasoning Segmentation}
\looseness = -1
On the other hand, remarkable advances have been made in large language models (LLMs), which can understand complex language inputs and have the potential for more complex referring segmentation.
% Recent advances in large language models (LLMs) have been extended to the development of multi-modal LLMs for visual understanding, showcasing their capabilities through tasks such as image captioning and visual question answering~\cite{li2023blip,liu2023improved,zhu2023minigpt,alayrac2022flamingo,yang2023dawn}.
%
Models such as BLIP-2~\cite{li2023blip}, LLaVA-1.5~\cite{liu2023improved}, MiniGPT-4~\cite{zhu2023minigpt}, and Flamingo~\cite{alayrac2022flamingo} have explored the design of multi-modal LLMs for visual understanding and demonstrate their ability through tasks such as image captioning, visual question answering (VQA), etc.
To enable the grounding ability of multimodal LLMs, 
% VisionLLM~\cite{wang2023visionllm} proposes an open-ended task decoder with an LLM and returns the coordinates of object polygons. 
%
Shikra~\cite{chen2023shikra} and MiniGPT-v2~\cite{chen2023minigpt} process object coordinates as input and enable the localization ability by returning coordinates.
However, these methods cannot produce segmentation masks and can only implicitly generate texts using LLMs rather than using a visual decoder for localization directly, which can be counterintuitive for image segmentation.

\looseness=-1
Recently, LISA~\cite{lai2023lisa} integrated a multi-modal LLM~\cite{liu2023improved} with a vision backbone and jointly trained a decoder to produce segmentation masks from language input. 
Despite using only 239 collected samples, LISA shows significant improvement in the reasoning process. However, its data is limited to entire objects, making it challenging for LISA to perform more fine-grained grounding.
Motivated by this limitation, we introduce the \textit{InstructPart} dataset, which contains instruction-part pairs, high-level affordance, low-level action, and part segmentation masks. With this dataset, we broaden the applicability of VLMs to various domains, such as manipulation, by enhancing their part grounding ability.
% They propose a new task -- reasoning segmentation, which requires the model to ground an area after comprehending a complex input sentence. 
%
% \simon{[We don't discuss results in the related work]}
% However, we find that LISA performs poorly when instructions are used as inputs for referencing specific parts. This limitation is particularly significant in the context of robotic applications, where the precise identification of parts is essential for effective interaction.
% Given the capabilities of the LISA approach, we propose a modification, named PISA (Part Identification and Segmentation Assistant), in order to reason over object parts from instructions, thus broadening the applicability of such approaches to various domains, including robotics and manipulation.
% This model is introduced alongside our \textit{InstructPart} dataset, which contains instruction-part pairs, object-part names, high-level affordance, low-level action, and the part segmentation mask.
% Given this dataset, we test and outline the potential of current state-of-the-art models with respect to the accuracy of identifying semantically meaningful object parts given an instruction of the desired task.

\section{The \Title{} Benchmark and Baseline Models}
\label{\Title{}}
In this section, we describe our \textit{\Title{}} benchmark in detail and introduce a simple baseline method for our benchmark.

\begin{figure*}[ht]
\vspace{1mm}
  \begin{center}
     \includegraphics[width=0.996\linewidth]{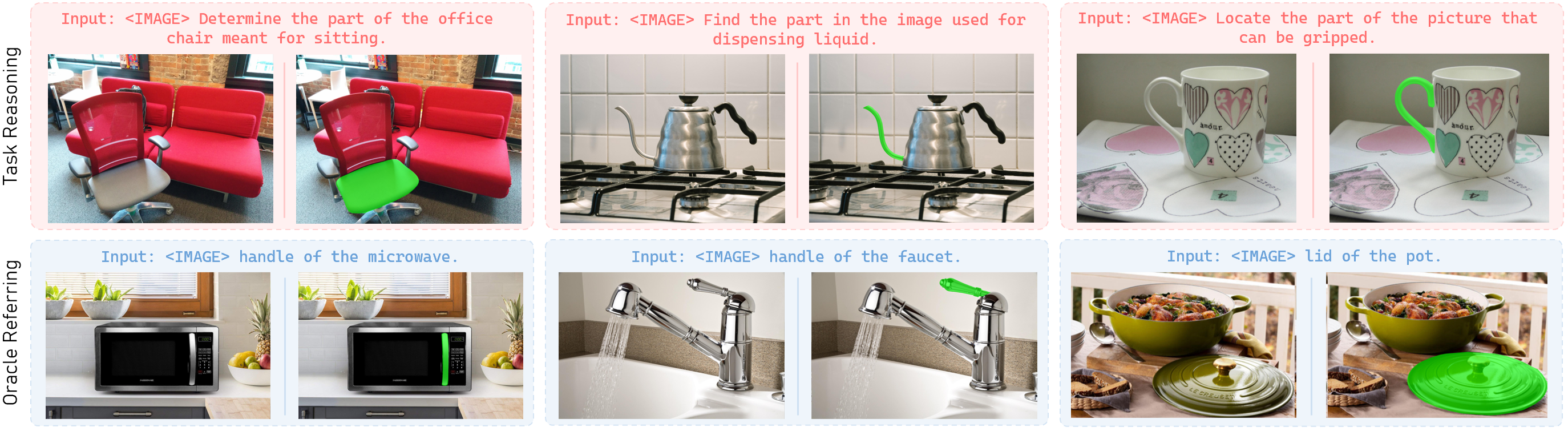}
  \end{center}
 \vspace{-0.18in}
  \caption{Examples from our \Title{} dataset are illustrated as follows: instruction queries are denoted in red text, while object and part names are indicated in blue. Each example includes an observation image (left), with the corresponding ground truth part segments (right), highlighted with a green mask.}
  \vspace{-0.16in}
  \label{fig:dataset_example}
\end{figure*}

\subsection{\Title{} Task Definition}
\label{sec::task definition}

Motivated by scenarios where agents need to localize areas based on task-specific queries, we define two tasks. The first, \textbf{Task Reasoning Part Segmentation (TRPS)}, challenges models to combine linguistic reasoning with visual grounding. The second, \textbf{Oracle Referring Part Segmentation (ORPS)}, focuses exclusively on evaluating visual grounding using oracle information about the designated object and part.

% \noindent
\looseness = -1
\textbf{TRPS.}
The TRPS task, illustrated in the first row of Fig.~\ref{fig:dataset_example}, is designed to explore the model's reasoning and part grounding abilities. The input is an instruction-image pair, and the goal is to identify the referred part's segmentation mask, as shown in green masks in Fig.~\ref{fig:dataset_example}. This task challenges the model to comprehend the instruction, analyze the image, and locate the corresponding part, Formally, the task is defined as: 
\(
    \mathcal{F}(I_{\text{instruction}}, I_{\text{image}})\Rightarrow M,
    \label{eq:instruction}
\)
% To explore the reasoning and part grounding ability of current models, we propose the TRPS task, which is shown in the first row of Fig.~\ref{fig:dataset_example}.
% %
% The models should only take an instruction-image pair as the input and find the part segmentation masks being referred to, which are shown in green masks in Fig.~\ref{fig:dataset_example}.
% %
% This requires the model to possess the ability to understand the instruction, analyze the image, and refer to the corresponding part area.
% %
% The task can be formulated as:
% \(
%     \mathcal{F}(I_{\text{instruction}}, I_{\text{image}})\Rightarrow M,
%     \label{eq:instruction}
% \)
where $\mathcal{F}$ represents the evaluated model, and \(I_{\text{instruction}} \in \{I_{\text{human}}, I_{\text{GPT}} \} \) is the instruction input that can either be annotated by human experts or rewritten by GPT-4.

% \noindent
\textbf{ORPS.}
In the ORPS task, shown in the second row of Fig.~\ref{fig:dataset_example}, the model is provided with direct part names to ensure accurate textual input. The task can be formulated in two ways:
We formulate the ORPS task in two formats:
\begin{enumerate}
    \item Including both the part name and the object name, e.g., \textit{the handle of the faucet}:
    \(
    \mathcal{F}(P \, \text{of} \, O, I_{\text{image}})\Rightarrow M.
    \label{eq:oracle1}
    \)
    \item Incorporating the affordance, e.g., \textit{the handle of the cup that can be held}, which could assist the model in identifying the part:
    \(
    \mathcal{F}(P \, \text{of} \, O \, \text{that} \, A_\text{a}, I_{\text{image}})\Rightarrow M,
    \label{eq:oracle2}
    \)
    where $A_\text{a}$ refers to the affordance.
    We manually adjust the active and passive voice of the affordance according to ensure grammatical precision.
    % \simon{[What's going on here? This equation looks so strange.]}\zifu{changed}
    % where $A_m$ is either the passive or the active voice of the affordance.
    % %
    % We first use a template to constitute $A_m$ with the passive voice of the affordance, such as \textit{the stem of the wine glass that can be held}.
    % %
    % Then, we manually adjust some affordances that should be an active voice, such as \textit{the shelves of the cabinet that can support}.
    % %
    % However, there are some exceptional circumstances. For example, the shelves of the cabinet have the affordance to support. In this case, we cannot simply include the passive voice of "support" in the template. As a result, we manually go through the entire dataset and correct inappropriate cases.

\end{enumerate}

%%%%%%%%%%%%%%%%%%%%%%%%%%%%%%%%%%

\subsection{\Title{} Dataset}
\label{sec:ip}
% %%%%%%%%%%%%
% %%%fig
% %%%%%%%%%%%%%
% \begin{figure}[htb]
%   \begin{center}
%      \includegraphics[width=1\linewidth]{figs/wordclouds.png}
%   \end{center}
%  \vspace{-0.26in}
%   \caption{\Title{} dataset object and part classes. \yaqi{Remove this figure. It doesn't tell us anything besides the un-balanceness}}
%   \vspace{-0.15in}
%   \label{fig:data classes}
% \end{figure}
% \looseness = -1
In line with our proposed tasks, we collect data to create the \Title{} dataset. This dataset is designed to evaluate the effectiveness of current models in understanding natural language and their ability to ground to specific parts.
It comprises 2,400 images, carefully selected to align with everyday household tasks.
% with 700 sourced from Flickr and 1,700 from AGD20K~\cite{Luo_2022_CVPR}, 
Specifically, \Title{} includes 48 object classes, 44 part categories, 30 affordances, and 37 actions. During data selection, a uniform distribution of object classes is ensured to create a balanced dataset. More details are included in Appendix~\ref{appendix: dataset detail}.

In the first row of Fig.~\ref{fig:dataset_example}, we show annotated examples for the TRPS task.
For each image sample, we manually design a task description based on the observed environment and the potential intention of an agent to interact with the object.
For each sample, we annotate all the fine-grained segmentation masks relevant to the task description as the ground truth. These masks are human-labeled to ensure accuracy and alignment with human understanding of object parts, maintaining the high quality of our dataset.
% \looseness = -1
We deliberately avoid specific part names in the instructions to better adapt to real-world scenarios. For example, commonly used expressions such as \textit{``Flush the toilet''} or \textit{``Turn on the faucet''} are preferred over more detailed directives such as \textit{``Press the toilet handle''} or \textit{``Lift the faucet handle''}.
The selection of these task descriptions aims to train models that are better at reasoning about object parts and their affordances, rather than simply identifying the part name that would solve the task.
By avoiding part names, our dataset more effectively analyzes the reasoning ability of models, requiring them to infer parts from implicit descriptions.
We engaged six human experts to create free-form natural language task instructions, which were then refined using GPT-4 for grammatical precision and sentence diversity. This was followed by thorough human verification to prevent hallucinations or other issues that can arise from using large language models for phrase diversification.
%
% We conduct a thorough review of each data piece fine-tuned on GPT-4 to prevent any unexpected errors in generation.
%
For the ORPS task, we use the part name and object name as the language input to evaluate the model's ability to directly ground to the part.

In addition to the instruction-image pairs, we provide the names of objects and parts relevant to the image, such as \textit{seat of the chair}, \textit{spout of the kettle}, \textit{handle of the cup}.
%
% For example, instead of referring to the \textit{switch of a toilet}, we annotate it by \textit{handle} or \textit{button} according to its real appearance in the image.
%
% These carefully chosen names can help models better refer to the specific parts.
%
We also include a corresponding affordance and action for each instruction.
Specifically, affordances refer to low-level actions performed to a specific part, like \textit{``pull''}, \textit{``push''}, or \textit{``twist''}, while actions refer to the high-level function to be achieved, such as \textit{``turn on''}, \textit{``pick up''}, or \textit{``open''}.
Note that the affordance and action could be identical sometimes, e.g., \textit{ ``pour''}, \textit{``cut''}, etc.
In the examples shown in the first row of Fig.~\ref{fig:dataset_example}, the affordances are \textit{``support''}, \textit{``pour''}, \textit{``grip''}, and the actions are \textit{``sit''}, \textit{``pour''} and \textit{``pick up''}.
This allows us to categorize affordances into two levels, addressing the ambiguity in definitions as noted in previous studies~\cite{nguyen2017object, roy2016multi, myers2015affordance}.
Note that in this work, we use the task descriptions and part names as the text input, while the affordance and action labels are reserved for future research.

In summary, the annotation for each of the samples in \Title{} can be represented as:
\(
(I_{\text{task}}, I_{\text{image}}, O, P, M, A_{\text{affordance}}, A_{\text{action}}),
\)
where these items refer to task instruction $I_{\text{task}}$, image observation $I_{\text{image}}$, object name $O$, part name $P$, segmentation mask $M$, affordance name $A_{\text{affordance}}$, and action name $A_{\text{action}})$.
Note that \(I_{\text{task}} \in \{I_{\text{human}}, I_{\text{GPT}} \} \), which means the text instruction is either directly annotated by humans or rewritten by GPT-4.
More annotated examples can be found in Appendix~\ref{appendix: annotation example} and \ref{appendix: more annotations}.

\subsection{Baseline Method}
\label{sec:pisa}
For our \Title{} benchmark, we build a simple yet effective baseline model: Part Identification and Segmentation Assistant (PISA). 
PISA originates from LISA~\cite{lai2023lisa}, which demonstrates superior capability in object-level reasoning segmentation.
Motivated by \cite{li2023one}, which shows the effectiveness of DINOv2~\cite{oquab2024dinov2} in extracting correspondence information among various parts, we improve LISA with a frozen DINOv2 backbone for feature extraction. As suggested by \cite{li2023one}, we use linear layers to integrate multi-level features from DINOv2 for various granularity information fusion.
The fused features are sent to an image decoder derived from SAM~\cite{kirillov2023segment}, where we apply Transpose Convolution and up-sampling for decoding in an alternating manner.
% PISA is motivated by \cite{li2023one}, showing the effectiveness of the self-supervised learning with DINOv2~\cite{oquab2024dinov2}, in the realm of extracting correspondence information among various images.
%
% We imbue a pre-trained LlaVA~\cite{liu2024visual} model with a frozen DINOv2 backbone for feature extraction.
%
% In our novel PISA pipeline, this setup is trained to extract and match features from vision and language input.
% Features from the two foundation models are sent to an image decoder derived from SAM~\cite{kirillov2023segment}, where we apply Transpose Convolution and up-sampling for decoding in an alternating manner.
% Utilizing our dataset, we fine-tune the decoder on our TRPS task, demonstrating superior performance to other methods, as shown in section~\ref{sec:experiments}.
% Since we freeze parameters of the vision and language encoders, we inject trainable rank decomposition~\cite{hu2021lora} matrices to the models for better optimization of our dataset.
% While utilizing a mixture of pre-existing components, the novelty of PISA is the integration of two large pre-trained methods, including both supervised learning and self-supervised learning, for better visual-language understanding.
\section{Experiments}
\label{sec:experiments}
% In this section, we first introduce our evaluation metrics (see Sec.~\ref{sec:metrics}) and evaluated methods (see Sec.~\ref{sec:eval_methods})). 
% Then we benchmark the performance of current state-of-the-art segmentation models on our novel object-part dataset, across the TRPS and ORPS tasks (see Sec.~\ref{sec:quant}). 
% Finally, we showcase the fine-tuning potential of our \Title{} dataset (see Sec.~\ref{sec:finetune}) before providing qualitative comparison across various methods (see Sec.~\ref{sec:qual}).

%%%%%%%%%%%%%%%%%%%%%%%%%%%%%%
%%%Tabel
%%%%%%%%%%%%%%%%%%%%%%%%%%%%%%
\begin{table*}[t]
  \centering
  \footnotesize
  \setlength{\tabcolsep}{1mm}{
  \resizebox{\linewidth}{!}{
  \begin{tabularx}{\textwidth}{c|c|*{4}{>{\centering\arraybackslash}X}|*{4}{>{\centering\arraybackslash}X}|*{4}{>{\centering\arraybackslash}X}|*{4}{>{\centering\arraybackslash}X}}
    \toprule
    & \multirow{3}{*}[-1.3ex]{Methods} & \multicolumn{8}{c|}{Oracle Referring Part Segmentation} & \multicolumn{8}{c}{Task Reasoning Part Segmentation} \\
    \cmidrule{3-18}
    & & \multicolumn{4}{c|}{Object-Part} & \multicolumn{4}{c|}{Object-Part-Affordance}  & \multicolumn{4}{c|}{Human-Annotated} & \multicolumn{4}{c}{GPT-4-Rewritten}\\
    \cmidrule{3-18}
    & & gIoU  & cIoU  & $\text{P}_{\text{50-95}}$   & $\text{P}_\text{50}$  & gIoU   & cIoU  &  $\text{P}_{\text{50-95}}$  & $\text{P}_\text{50}$ & gIoU   & cIoU  & $\text{P}_{\text{50-95}}$   & $\text{P}_\text{50}$  & gIoU   & cIoU  &  $\text{P}_{\text{50-95}}$  & $\text{P}_\text{50}$ \\
    \midrule\midrule
    % \hline\hline
      \multirow{3}{*}[-0.2ex]{\centering \rotatebox{90}{OVS}}
    & VLPart & 22.06 & 21.78 & 16.02 & 22.50 & 15.32 & 12.78 & 11.83 & 15.33 & 0.39 & 1.16 & 0.00 & 0.00 & 0.76 & 0.84 & 0.20 & 0.50  \\
      & OVSeg & 28.58 & 20.49 & 10.37 & 22.33 & 28.60 & 20.99 & 10.87 & 22.50 & 22.44 & 14.11 & 7.07 & 15.33 & 23.14 & 15.51 & 7.13 & 15.17  \\
      & SAN & 10.51 & 20.24 & 4.72 & 10.17 & 12.11 & 20.37 & 5.48 & 12.00 & 9.08 & 13.56 & 2.62 & 6.67 & 6.96 & 14.69 & 1.90 & 5.17  \\
      \midrule
      & X-Decoder & 18.96 & 15.65 & 8.52 & 14.83 & 18.96 & 15.65 & 8.52 & 14.83 & 17.48 & 13.61 & 7.00 & 13.17 & 17.38 & 12.76 & 6.90 & 13.17  \\
      \multirow{4}{*}[2.4ex]{\centering \rotatebox{90}{RES}}
      & SEEM & 13.54 & 14.63 & 6.33 & 10.50 & 13.54 & 14.63 & 6.33 & 10.50 & 13.52 & 14.09 & 4.97 & 9.83 & 14.53 & 14.19 & 4.57 & 10.67  \\
      & TRIS & 23.02 & 19.90 & 6.97 & 17.50 & 23.11 & 19.65 & 6.98 & 18.50 & 21.97 & 17.83 & 6.68 & 15.00 & 22.66 & 18.52 & 7.03 & 16.83  \\
      & G-SAM & 34.33 & 24.83 & 15.03 & 28.83 & 33.63 & 24.79 & 14.42 & 27.83 & \underline{29.95} & \underline{21.45} & \underline{11.98} & \underline{25.17} & 29.57 & \underline{21.88} & 11.60 & 23.00  \\
      \midrule
      \multirow{3}{*}[-0.2ex]{\centering \rotatebox{90}{RS}}
      &  LISA & \underline{34.46} & \textbf{39.44} & \underline{17.48} & \underline{32.67} & \textbf{35.77} & \textbf{39.62} & \textbf{18.78} & \textbf{34.50} & \textbf{32.11} & \textbf{30.25} & \textbf{16.98} & \textbf{30.00} & \textbf{29.75} & \textbf{27.44} & \textbf{15.08} & \textbf{27.83}  \\   
      & Shikra & 4.50 & 7.20 & 2.67 & 4.17 & 9.36 & 15.37 & 4.92 & 7.83 & 1.70 & 3.48 & 0.83 & 1.50 & 14.65 & 12.95 & 8.40 & 13.33  \\
      & MiniGPT-v2 & \textbf{35.65} & \underline{36.05} & \textbf{18.38} & \textbf{33.17} & \underline{34.58} & \underline{35.11} & \underline{18.50} & \underline{34.27} & 26.29 & 19.46 & 13.00 & 24.00 & \underline{29.67} & 21.37 & \underline{15.07} & \underline{24.17}  \\
      % & SAM-GPT-4V  & -- & -- & -- & -- & -- & -- & -- & -- & -- & -- & -- & -- & -- & -- & -- & -- \\
      \midrule
      &  Average & 22.56 & 22.02 & 10.65 & 19.67 & 21.95 & 21.10 & 10.66 & 19.81 & 17.49 & 14.90 & 7.11 & 14.44 & 17.59 & 16.02 & 7.79 & 14.98  \\
    \bottomrule
  \end{tabularx}
  }
  \vspace{-6pt}
  }
  \caption{Results on ORPS (left) and TRPS (right) tasks. We divide the methods into three categories, namely, open-vocabulary segmentation (OVS), referring expression segmentation (RES), and reasoning segmentation (RS). The best results are \textbf{bolded}, and the second-best are \underline{underlined}.}
  \label{tab:combined_refer}
  \vspace{-10pt}
\end{table*}

\subsection{Metrics}
\label{sec:metrics}
% \looseness=-1
To evaluate our approach, we use standard metrics in LISA~\cite{lai2023lisa}, namely gIoU and cIoU. 
gIoU reflects the average of all per-image Intersection-over-Unions (IoUs), while cIoU is defined by the cumulative intersection over the cumulative union.
To evaluate the precision of the models, we adopt Precision@50 (P@50) metric as the previous referring segmentation works~\cite{liu2023referring, mao2016generation} and develop a Precision@50:95 (P@50:95) metric according to COCO~\cite{lin2014microsoft}.
The P@50 metric considers a mask to be a true positive when the IoU ratio exceeds 0.5, and P@50:95 calculates across a range of IoU thresholds from 0.50 to 0.95 with increments of 0.05, then averages across all the thresholds.
The P@50:95 metric requires a higher least IoU for the prediction; hence, it is always lower than the P@50 metric.
For the two metric types, IoU and Precision, the latter metric only counts those results greater than a threshold, hence can pose more challenges to the model and fairly evaluate the results with a high recall rate.

\subsection{Evaluated Methods}
\label{sec:eval_methods}
Here, we introduce the set of baseline models utilized in our experiments. 
More details about the model settings can be found in Appendix~\ref{appenix: model details}.

% \noindent
\textbf{Open-vocabulary Segmentation Models.}
The open-vocabulary part segmentation model, i.e., VLPart~\cite{sun2023going}, is intuitively suitable for our tasks since plentiful part segments were used for training. 
%
% As a result, we would like to know whether they can perform well in our dataset collected especially for robotic daily tasks.
%
We also choose OVSeg~\cite{liang2023open} and SAN~\cite{xu2023side} to discover the performance of the open-vocabulary object segmentation methods on our task.
We select the best-reported models for the three methods.
% , \textit{ovseg\_swinbase\_vitL14\_ft\_mpt.pth} and \textit{san\_vit\_large\_14.pth} respectively.

% \noindent
\looseness = -1
\textbf{Refering Segmentation Models.}
We conduct experiments with off-the-shelf models including X-Decoder~\cite{zou2023generalized}, SEEM~\cite{zou2023segment}, and TRIS~\cite{liu2023referring}. 
%
% We adopt \textit{xdecoder\_focalt\_last.pt}, \textit{seem\_focall\_v1.pt}, and \textit{stage2\_refcocog\_google.pth} for the three models respectively.
%
Besides, we also evaluate Grounding-DINO~\cite{liu2023grounding}, which has provided a great open-vocabulary referring detection ability and has been integrated with SAM~\cite{kirillov2023segment}, namely Grounded-SAM.
%\footnote{\url{https://github.com/IDEA-Research/Grounded-Segment-Anything}}.
%
We adopt the best models for these methods.

% \noindent
\textbf{Reasoning Segmentation Models.}
For our tasks, LISA~\cite{lai2023lisa} is a natural choice since it can return masks and has been trained on several part segmentation datasets~\cite{he2022partimagenet, chen2014detect, ramanathan2023paco}.
% As a result, it is interesting to explore whether it possesses the ability to understand instructions and find part segments.
%
Other multi-modal LLMs, including Shikra~\cite{chen2023shikra} and MiniGPT-v2~\cite{chen2023minigpt} also have localization ability and have been chosen for our evaluation. 
Since they can only return bounding box outputs, we use the results as box prompts for SAM~\cite{kirillov2023segment} to get a mask output for fair comparison.
%
% However, we cannot test VisionLLM since its code has not been released.
%
% To prompt LISA, we follow its original setting to add \textit{``Please output the segmentation mask.''} at the end of each instruction.
% %
% Besides, in order to formulate a query for the oracle referring part segmentation task, we embed the object and part name in a format of: ``Where is the {$I_\text{text}$} in the image'', where $I_\text{text}$ stands for the text input mentioned in Sec.~\ref{sec::task definition}.
%
% To prompt Shikra, we integrate our instruction in its original template as follows:
% \begin{itemize}
%     \item Instruction reasoning part segmentation: \\ \textless $I_{\text{text}}$\textgreater. Can you point out all the related parts in the image \textless $I_{\text{image}}$\textgreater \, and provide the coordinates of their locations?
%     \item Oracle referring part segmentation: \\Can you point out all the \textless $I_{\text{text}}$\textgreater \, in the image \textless $I_{\text{image}}$\textgreater \, and provide the coordinates of their locations?
% \end{itemize}
%
% We adopt LISA-7B-v1~\cite{lai2023lisa} model that has been fine-tuned on both training and validation data of LISA's dataset.
% %
% Additionally, we select the Shikra-7B-delta-v1-0708 for Shikra and the stage-3 model for MiniGPT-v2.

% \noindent
\textbf{Grid-based GPT-4V.}
The recent release of GPT-4V has demonstrated remarkable advancements in complex visio-linguistic reasoning~\cite{yang2023dawn}.
% , outperforming its predecessors in previous challenges, such as Winoground~\cite{thrush2022winoground}.
%
% As a natural thought, we wonder if GPT-4V can also succeed in understanding and grounding object parts.
%
However, GPT-4V API cannot return segmentation mask output directly, and our preliminary experiments showed that GPT-4V performs poorly when it is asked to generate text coordinates.
As a result, we first use Grounding-DINO~\cite{liu2023grounding} to find the bounding box of the entire object and crop it, then ask GPT-4V to virtually divide the box to \(7\times7\) grids and identify the grids including the desirable parts.
Afterward, the coordinates of the grids are used as a prompt for SAM~\cite{kirillov2023segment} to obtain the segmentation mask.

% \noindent
\textbf{SoM-based GPT-4V.}
SoM~\cite{yang2023set} proposes to label the masks obtained by SAM~\cite{kirillov2023segment} with numbers in the center of each object. 
As it proves that precise referring can boost the performance of GPT-4V, we apply a similar manner for our part segmentation task.
% In this case, we also use Grounding-DINO~\cite{liu2023grounding} first and apply SoM to the object patches instead of entire image.
%
% Although SAM~\cite{kirillov2023segment} can be a superior choice to obtain masks at multiple granularities~\cite{zou2023segment}, it is prone to failing in part segmentation for small objects.
% %
% As a result, we add Grounding-DINO~\cite{liu2023grounding} to detect the object first, then apply SoM to the object patch instead of entire image.

% \looseness=-1
\textbf{PISA and Fine-tuning.}
To evaluate our proposed method, we use all training data of LISA~\cite{lai2023lisa} for pertaining and fine-tuning with 1,800 samples of our data. As a comparison, we also fine-tune LISA with the same data.
Besides, we also train the models with multiple numbers of samples. More results can be found in Appendix~\ref{appendix: training samples}.

\subsection{Quantitative Results of SOTA VLMs}
\label{sec:quant}

% \noindent
\textbf{Open-sourced VLMs Results.}
\looseness = -1
The left part of Tab.~\ref{tab:combined_refer} shows the result of our ORPS task, where object and part names are explicitly embedded into a template, mitigating the need for models' reasoning ability. 
% The \textit{Object-Part} column stands for the results using the template mentioned in Eq.~\ref{eq:oracle1}, while the \textit{Object-Part-Affordance} column incorporates the affordance according to Eq.~\ref{eq:oracle2}.
%
The right part of Tab.~\ref{tab:combined_refer} shows the result of TRPS, where part names are not present in the instruction and require more reasoning ability to understand the implicit meaning.
\looseness = -1
Comparing the left and right parts of Tab.~\ref{tab:combined_refer} we can find that the performance of oracle referring task is generally better than that of task reasoning. 
This demonstrates that current models lack the reasoning ability to infer from a task-image pair to the correct interactive part.
For the ORPS task, incorporating the affordance in the instruction leads to no apparent increase in the average performance.
%
% While LISA and X-Decoder achieve some increase, other models are impacted by the affordance.
%
This indicates that most models may not possess the common sense to relate a part to an affordance, suggesting the potential of \Title{} for affordance learning.
Besides, for the TRPS task, we can find that GPT-4 rewritten instructions lead to overall better performances.
This indicates that the precise instruction descriptions generated by GPT-4 align more effectively with the language embedding space of multimodal LLMs, enhancing the reasoning capabilities of vision-language models for handling instructions.
\textbf{GPT-4V Based Methods Results.}
Tab.~\ref{tab:gpt4v} shows the results of two GPT-4V segmentation methods.
%
% However, due to the quota restriction, we are not able to frequently call GPT-4V API.
%
We test the two methods on the oracle referring task to explore GPT-4V's localization ability.
We select a subset consisting of 226 samples from the dataset according to the original category distribution.
Although the results cannot be fairly compared with other methods in Tab.~\ref{tab:combined_refer}, it still reveals the poor performance of GPT-4V.
%
% For the two methods, the SoM-based method performs better, demonstrating that GPT-4V has difficulties directly localizing the part targets and has to choose from a set.
% %
% However, compared to the satisfactory results in \cite{yang2023set}, SoM fails to handle our dataset.
%
Two reasons may explain this: 
1) While GPT-4V can localize objects~\cite{yang2023set}, we hypothesize that it is not trained directly on fine-grained part data.
2) Labeling numbers in the center of fine-grained parts may lead to overlapping and ambiguity in referring.
\begin{table}[h]
  \renewcommand\arraystretch{0.9}
  \centering
  \footnotesize
  \centering
     \setlength{\tabcolsep}{1.5mm}{
     \resizebox{0.95\linewidth}{!}{
     \begin{tabular}{c|cccc}
          % \specialrule{.1em}{.05em}{.05em}
          \toprule
          \multirow{2}{*}[-0.7ex]{Methods}& \multicolumn{4}{c}{Object-Part} \\
            \cmidrule{2-5}
                  & gIoU  & cIoU  & $\text{P}_{\text{50-95}}$   & $\text{P}_\text{50}$  \\
            \midrule\midrule
            Grid-based GPT-4V   & 14.14 & 17.15 & 5.67 & 12.37 \\
            \midrule
            SoM-based GPT-4V   & 25.41 & 26.82 & 17.90 & 25.81 \\
            % SAN~\cite{xu2023side}  & 13.45 & 19.70 & 5.36 & 10.74  \\
            % \specialrule{.1em}{.05em}{.05em} 
            \bottomrule
    \end{tabular}}
}
\vspace{-4pt}
  \caption{GPT-4's performance in the object-part oracle referring part segmentation task, as applied to a subset of InstructPart.}
  \label{tab:gpt4v}
  \vspace{-7pt}
\end{table}

%%%%%%%%%%%%%%%%%%%%%%%%%%%%%%
%%%Tabel
%%%%%%%%%%%%%%%%%%%%%%%%%%%%%%

% \looseness = -1
\subsection{Quantitative Results of Fine-tuning with \Title{}} 
\label{sec:finetune}
Tab.~\ref{tab:finetune} shows the results of TRPS task with human-annotated instructions. The pre-trained PISA outperforms LISA by a large margin, demonstrating its strong reasoning part segmentation ability.
After fine-tuning, both LISA and PISA gain great improvement in all metrics, indicating the exceptional quality and training utility of our data.

\begin{table}[h]
  \renewcommand\arraystretch{0.95}
  \centering
  \footnotesize
  % \vspace{2mm}
  \centering
     \setlength{\tabcolsep}{1.5mm}{
     \resizebox{0.95\linewidth}{!}{
     \begin{tabular}{c|cccc}
          % \specialrule{.1em}{.05em}{.05em}
          \toprule
          Methods & gIoU  & cIoU  & $\text{P}_{\text{50-95}}$ & $\text{P}_\text{50}$  \\
            \midrule\midrule
            LISA-Pretrained & 32.11 & 30.25 & 16.98 & 30.00 \\
            PISA-Pretrained & 43.46 & 46.76 & 20.00 & 44.50\\
            LISA-Tuned & 71.26 & 72.14 & 57.73 & 79.33\\
            PISA-Tuned & \textbf{76.19} & \textbf{78.39} & \textbf{62.20} & \textbf{87.00}\\
            \bottomrule
    \end{tabular}}
}
\vspace{-4pt}
  \caption{Comparison of pre-training and fine-tuning results. We use all datasets that LISA was trained on to get the pre-trained model. Fine-tuned models are trained with 1,800 samples in \Title{}.}
  \vspace{-6pt}
  \label{tab:finetune}
\end{table}

\afterpage{%
\begin{figure*}[!t]
	\centering
	\vspace{-4mm}
	\resizebox{1\textwidth}{!}
	{
		\begin{tabular}{@{}c@{}c@{}c@{}c@{}c@{}c@{}c@{}c@{}c@{}c@{}c@{}c}

% \vspace{-2mm}
% \includegraphics[width=1.8cm,height=1.8cm]{figs/selected_imgs_neurips/gt/1009786005_d4a02fd811_o-faucet-handle.png}\hspace{1mm} &
% \includegraphics[width=1.8cm,height=1.8cm]{figs/selected_imgs_neurips/pred_mask_human_xdecoder/1009786005_d4a02fd811_o-faucet-handle.png}\hspace{1mm} &
% \includegraphics[width=1.8cm,height=1.8cm]{figs/selected_imgs_neurips/pred_mask_human_seem/1009786005_d4a02fd811_o-faucet-handle.png}\hspace{1mm} &
% \includegraphics[width=1.8cm,height=1.8cm]{figs/selected_imgs_neurips/pred_mask_human_tris/1009786005_d4a02fd811_o-faucet-handle.png}\hspace{1mm} &
% \includegraphics[width=1.8cm,height=1.8cm]{figs/selected_imgs_neurips/pred_mask_human_groundedsam/1009786005_d4a02fd811_o-faucet-handle.png}\hspace{1mm} &
% \includegraphics[width=1.8cm,height=1.8cm]{figs/selected_imgs_neurips/pred_mask_human_minigpt/1009786005_d4a02fd811_o-faucet-handle.png}\hspace{1mm} &
% \includegraphics[width=1.8cm,height=1.8cm]{figs/selected_imgs_neurips/pred_lisa-untrain-test/1009786005_d4a02fd811_o-faucet-handle.png}\hspace{1mm} &
% \includegraphics[width=1.8cm,height=1.8cm]{figs/selected_imgs_neurips/pred_lisa-train1800/1009786005_d4a02fd811_o-faucet-handle.png}\hspace{1mm} &
% \includegraphics[width=1.8cm,height=1.8cm]{figs/selected_imgs_neurips/pred_pisa_pretrain-v1-dinodecoder-train1800/1009786005_d4a02fd811_o-faucet-handle.png}\hspace{1mm} \\
\vspace{-2mm}
\includegraphics[width=1.8cm,height=1.8cm]{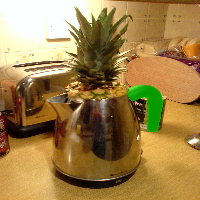}\hspace{1mm} &
\includegraphics[width=1.8cm,height=1.8cm]{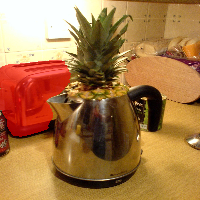}\hspace{1mm} &
\includegraphics[width=1.8cm,height=1.8cm]{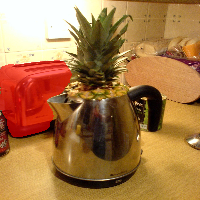}\hspace{1mm} &
\includegraphics[width=1.8cm,height=1.8cm]{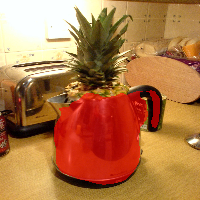}\hspace{1mm} &
\includegraphics[width=1.8cm,height=1.8cm]{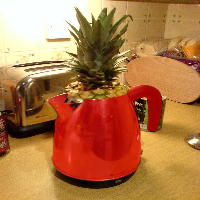}\hspace{1mm} &
\includegraphics[width=1.8cm,height=1.8cm]{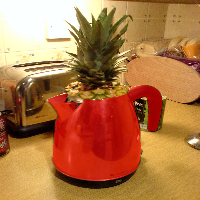}\hspace{1mm} &
\includegraphics[width=1.8cm,height=1.8cm]{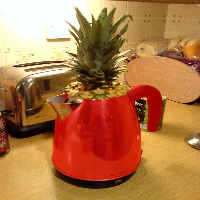}\hspace{1mm} &
\includegraphics[width=1.8cm,height=1.8cm]{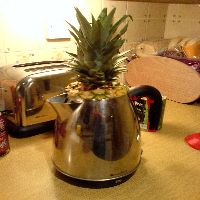}\hspace{1mm} &
\includegraphics[width=1.8cm,height=1.8cm]{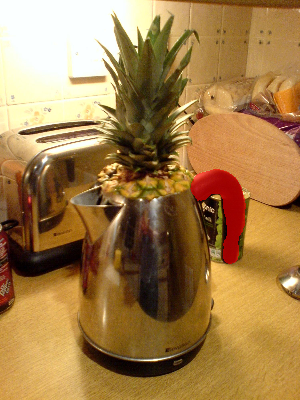}\hspace{1mm} \\
   \vspace{-2mm}
			%-----------------------------------------------------------------------------------------
			\includegraphics[width=1.8cm,height=1.8cm]{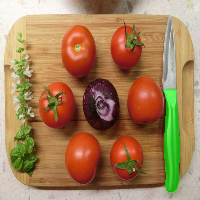}\hspace{1mm} &
                \includegraphics[width=1.8cm,height=1.8cm]{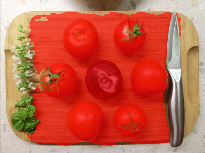}\hspace{1mm} &
			\includegraphics[width=1.8cm,height=1.8cm]{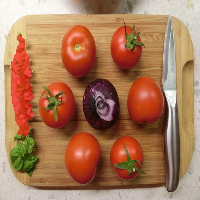}\hspace{1mm} &
			\includegraphics[width=1.8cm,height=1.8cm]{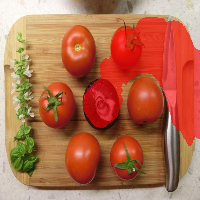}\hspace{1mm} &
			\includegraphics[width=1.8cm,height=1.8cm]{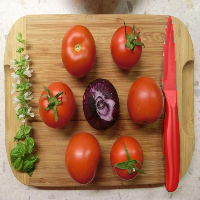}\hspace{1mm} &
			\includegraphics[width=1.8cm,height=1.8cm]{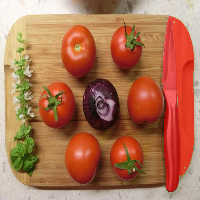}\hspace{1mm} &
			\includegraphics[width=1.8cm,height=1.8cm]{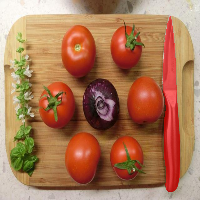}\hspace{1mm} &
   			\includegraphics[width=1.8cm,height=1.8cm]{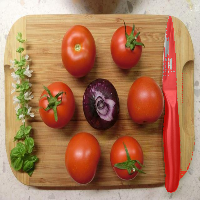}\hspace{1mm} &
			\includegraphics[width=1.8cm,height=1.8cm]{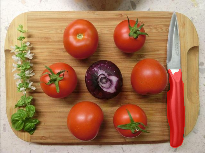}\hspace{1mm} &\\[1pt]
                %-----------------------------------------------------------------------------------------
			{\scriptsize Ground Truth} & {\scriptsize X-Decoder} & {\scriptsize SEEM} & {\scriptsize TRIS} & {\scriptsize G-SAM} & {\scriptsize MiniGPT-v2} & {\scriptsize LISA-Pretrain} & {\scriptsize LISA-Finetune} & {\scriptsize \textcolor{red}{PISA-Finetune}}\ \\
		\end{tabular}
	}
	\vspace{-3mm}
	\caption{Qualitative comparison of different VLMs and the fine-tuned models. In these examples, the pre-trained LISA falls short of recognizing the correct part. After fine-tuning, PISA shows better potential for part understanding than LISA. More results can be found in Figure~\ref{fig:qualitative supplementary results 3}.}
	\label{fig:qualitative results 3}
	\vspace{-4mm}
\end{figure*}
%%%%%%%%%%%%%%%%%%%%%%%%%%%%%%%%%%%5
%% both tune good
%%%%%%%%%%%%%%%%%%%%%%%%%%%%%%%%%%%%

\begin{figure*}[!t]
	\centering
	% \vspace{-4mm}
	\resizebox{1\textwidth}{!}
	{
		\begin{tabular}{@{}c@{}c@{}c@{}c@{}c@{}c@{}c@{}c@{}c@{}c@{}c@{}c}
 \vspace{-2mm}
\includegraphics[width=1.8cm,height=1.8cm]{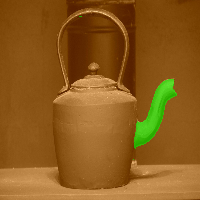}\hspace{1mm} &
\includegraphics[width=1.8cm,height=1.8cm]{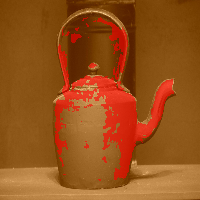}\hspace{1mm} &
\includegraphics[width=1.8cm,height=1.8cm]{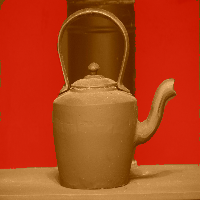}\hspace{1mm} &
\includegraphics[width=1.8cm,height=1.8cm]{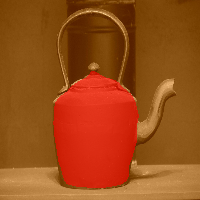}\hspace{1mm} &
\includegraphics[width=1.8cm,height=1.8cm]{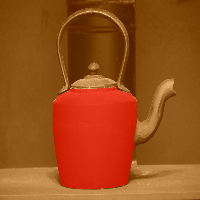}\hspace{1mm} &
\includegraphics[width=1.8cm,height=1.8cm]{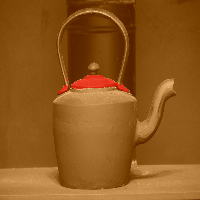}\hspace{1mm} &
\includegraphics[width=1.8cm,height=1.8cm]{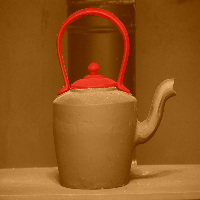}\hspace{1mm} &
\includegraphics[width=1.8cm,height=1.8cm]{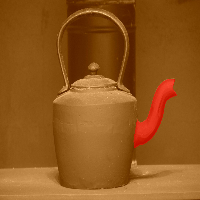}\hspace{1mm} &
\includegraphics[width=1.8cm,height=1.8cm]{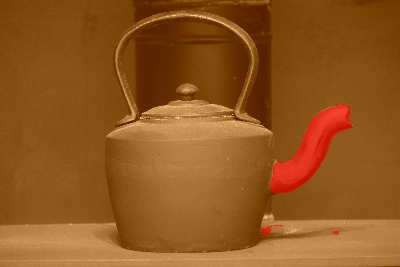}\hspace{1mm} \\
 \vspace{-2mm}

\includegraphics[width=1.8cm,height=1.8cm]{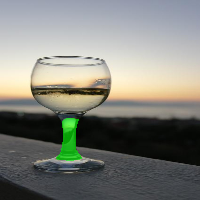}\hspace{1mm} &
\includegraphics[width=1.8cm,height=1.8cm]{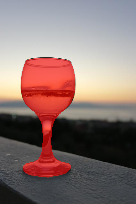}\hspace{1mm} &
\includegraphics[width=1.8cm,height=1.8cm]{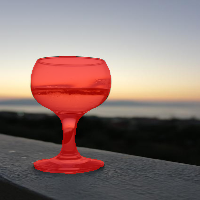}\hspace{1mm} &
\includegraphics[width=1.8cm,height=1.8cm]{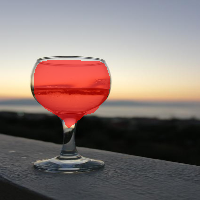}\hspace{1mm} &
\includegraphics[width=1.8cm,height=1.8cm]{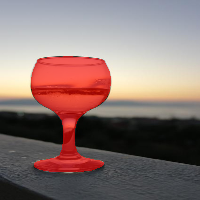}\hspace{1mm} &
\includegraphics[width=1.8cm,height=1.8cm]{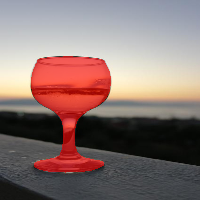}\hspace{1mm} &
\includegraphics[width=1.8cm,height=1.8cm]{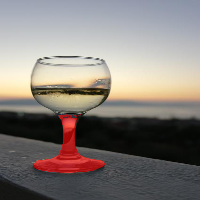}\hspace{1mm} &
\includegraphics[width=1.8cm,height=1.8cm]{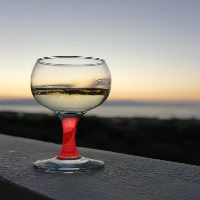}\hspace{1mm} &
\includegraphics[width=1.8cm,height=1.8cm]{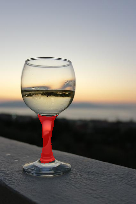}\hspace{1mm} &\\[1pt]
                %-----------------------------------------------------------------------------------------
			{\scriptsize Ground Truth} & {\scriptsize X-Decoder} & {\scriptsize SEEM} & {\scriptsize TRIS} & {\scriptsize G-SAM} & {\scriptsize MiniGPT-v2} & {\scriptsize LISA-Pretrain} & {\scriptsize \textcolor{red}{LISA-Finetune}} & {\scriptsize \textcolor{red}{PISA-Finetune}}\ \\
		\end{tabular}
	}
	\vspace{-3mm}
	\caption{Qualitative comparison of different VLMs and the fine-tuned models. In these examples, the pre-trained LISA falls short of recognizing the correct part. After fine-tuning, both LISA and PISA perform well on the part identification. More results can be found in Figure~\ref{fig:qualitative supplementary results 2}.}
	\label{fig:qualitative results 2}
	\vspace{-4mm}
\end{figure*}
%%%%%%%%%%%%%%%%%%%%%%%%%%%%%%%
% page 3
%%%%%%%%%%%%%%%%%%%%%%%%%%%%%%%

\begin{figure*}[!t]
	\centering
	% \vspace{-4mm}
	\resizebox{1\textwidth}{!}
	{
		\begin{tabular}{@{}c@{}c@{}c@{}c@{}c@{}c@{}c@{}c@{}c@{}c@{}c@{}c}
  %%%%%%%%%%%%%lisa already good
%  \vspace{-2mm}

% \includegraphics[width=1.8cm,height=1.8cm]{figs/selected_imgs_neurips/gt/knife_000530-knife-blade.png}\hspace{1mm} &
% \includegraphics[width=1.8cm,height=1.8cm]{figs/selected_imgs_neurips/pred_mask_human_xdecoder/knife_000530-knife-blade.png}\hspace{1mm} &
% \includegraphics[width=1.8cm,height=1.8cm]{figs/selected_imgs_neurips/pred_mask_human_seem/knife_000530-knife-blade.png}\hspace{1mm} &
% \includegraphics[width=1.8cm,height=1.8cm]{figs/selected_imgs_neurips/pred_mask_human_tris/knife_000530-knife-blade.png}\hspace{1mm} &
% \includegraphics[width=1.8cm,height=1.8cm]{figs/selected_imgs_neurips/pred_mask_human_groundedsam/knife_000530-knife-blade.png}\hspace{1mm} &
% \includegraphics[width=1.8cm,height=1.8cm]{figs/selected_imgs_neurips/pred_mask_human_minigpt/knife_000530-knife-blade.png}\hspace{1mm} &
% \includegraphics[width=1.8cm,height=1.8cm]{figs/selected_imgs_neurips/pred_lisa-untrain-test/knife_000530-knife-blade.png}\hspace{1mm} &
% \includegraphics[width=1.8cm,height=1.8cm]{figs/selected_imgs_neurips/pred_lisa-train1800/knife_000530-knife-blade.png}\hspace{1mm} &
% \includegraphics[width=1.8cm,height=1.8cm]{figs/selected_imgs_neurips/pred_pisa_pretrain-v1-dinodecoder-train1800/knife_000530-knife-blade.png}\hspace{1mm} \\
 \vspace{-2mm}

\includegraphics[width=1.8cm,height=1.8cm]{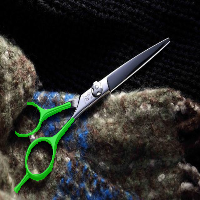}\hspace{1mm} &
\includegraphics[width=1.8cm,height=1.8cm]{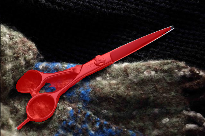}\hspace{1mm} &
\includegraphics[width=1.8cm,height=1.8cm]{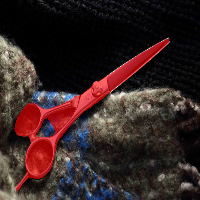}\hspace{1mm} &
\includegraphics[width=1.8cm,height=1.8cm]{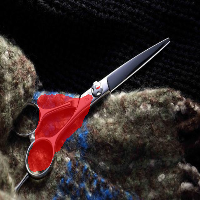}\hspace{1mm} &
\includegraphics[width=1.8cm,height=1.8cm]{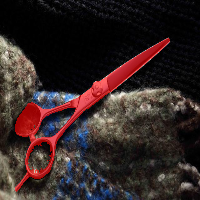}\hspace{1mm} &
\includegraphics[width=1.8cm,height=1.8cm]{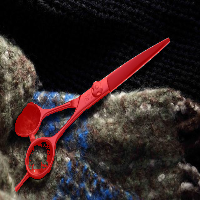}\hspace{1mm} &
\includegraphics[width=1.8cm,height=1.8cm]{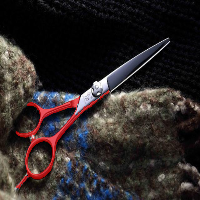}\hspace{1mm} &
\includegraphics[width=1.8cm,height=1.8cm]{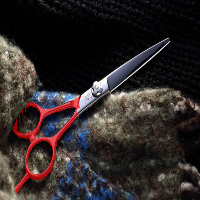}\hspace{1mm} &
\includegraphics[width=1.8cm,height=1.8cm]{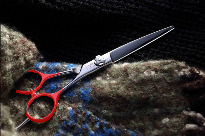}\hspace{1mm} \\
%  \vspace{-2mm}
 \vspace{-2mm}

\includegraphics[width=1.8cm,height=1.8cm]{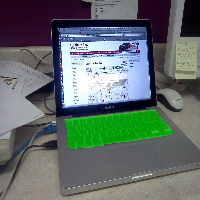}\hspace{1mm} &
\includegraphics[width=1.8cm,height=1.8cm]{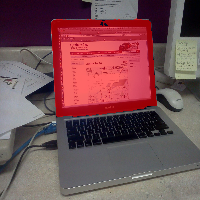}\hspace{1mm} &
\includegraphics[width=1.8cm,height=1.8cm]{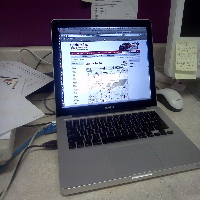}\hspace{1mm} &
\includegraphics[width=1.8cm,height=1.8cm]{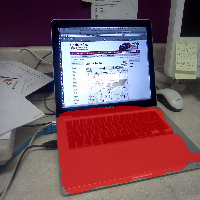}\hspace{1mm} &
\includegraphics[width=1.8cm,height=1.8cm]{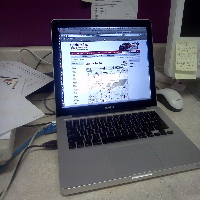}\hspace{1mm} &
\includegraphics[width=1.8cm,height=1.8cm]{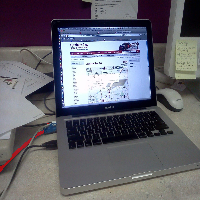}\hspace{1mm} &
\includegraphics[width=1.8cm,height=1.8cm]{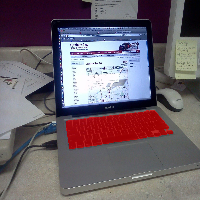}\hspace{1mm} &
\includegraphics[width=1.8cm,height=1.8cm]{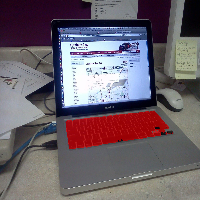}\hspace{1mm} &
\includegraphics[width=1.8cm,height=1.8cm]{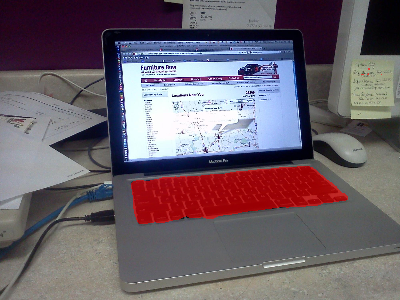}\hspace{1mm} &\\[1pt]
                %-----------------------------------------------------------------------------------------
			{\scriptsize Ground Truth} & {\scriptsize X-Decoder} & {\scriptsize SEEM} & {\scriptsize TRIS} & {\scriptsize G-SAM} & {\scriptsize MiniGPT-v2} & {\scriptsize \textcolor{red}{LISA-Pretrain}} & {\scriptsize LISA-Finetune} & {\scriptsize PISA-Finetune}\ \\
		\end{tabular}
	}
	\vspace{-3mm}
	\caption{Qualitative comparison of different VLMs and the fine-tuned models. In these examples, the pre-trained LISA already delivers good identification of the target parts. More results can be found in Figure~\ref{fig:qualitative supplementary results 2}.}
	\label{fig:qualitative results 1}
	\vspace{-5mm}
\end{figure*}

}

\subsection{Qualitative Results}
\label{sec:qual}
Fig.~\ref{fig:qualitative results 3},\ref{fig:qualitative results 2},\ref{fig:qualitative results 1} shows the visualization results on the TRPS task.
The first column depicts the ground truth labels, and the remaining columns include the results of off-the-shelf VLMs: 
X-Decoder
~\cite{zou2023generalized}
, SEEM
~\cite{zou2023segment}
, TRIS
~\cite{liu2023referring}
, Grounded-SAM
~\cite{kirillov2023segment,liu2023grounding}
, MiniGPT-v2
~\cite{chen2023minigpt}
, LISA
~\cite{lai2023lisa}.
The last two columns show the results of fine-tuned LISA and PISA models.
As shown by the examples, most VLMs tend to either obtain the entire object area or miss the correct regions, demonstrating the challenging tasks provided by \Title{}.
In Fig.~\ref{fig:qualitative results 3}, we present examples where the fine-tuned PISA shows superior visual part segmentation results, demonstrating the effectiveness of our proposed method. Besides, both the pre-trained and fine-tuned LISA models also demonstrate great potential in part grounding. Here, we visualize additional results of the VLMs and fine-tuned models. As shown in Fig.~\ref{fig:qualitative results 1}, the pre-trained LISA~\cite{lai2023lisa} can better identify desired parts compared to other VLMs. This indicates the evaluation usage of our \Title{} dataset, where all the advanced VLMs can be evaluated and compared. Furthermore, in Fig.~\ref{fig:qualitative results 2}, the pre-trained LISA fails to recognize target parts, similar to other VLMs, while both fine-tuned models significantly improve the results.
More visualizations are available in Appendix~\ref{appendix: more qualitative}.

\section{Discussion}
% \vspace{-3pt}
\textbf{Scale of \Title{} dataset.}
% Concerns may arise regarding the size of \Title{}, but we offer the following explanations:
We consider \Title{} a sufficient task-oriented part segmentation dataset for the following reasons:
\textbf{\textit{1)}} The size of \Title{} already exceeds that of several recent Vision-Language evaluation datasets, such as MMStar~\cite{chen2024we} (1500 samples, \textit{NeurIPS'24}), VisIT-Bench~\cite{bitton2024visit} (592 images, \textit{NeurIPS'23}), WHOOPS!~\cite{bitton2023breaking} (500 images, \textit{ICCV'23}), and TIFA160~\cite{hu2023tifa} (800 generated images, \textit{ICCV'23}).
We believe that our data are adequate for thorough evaluations of current models.
\textbf{\textit{2)}} \Title{} addresses a gap in data related to reasoning about robot-object interaction and part segmentation (e.g., PartImageNet includes only one relevant category: bottle).
\textbf{\textit{3)}} Fine-tuning LISA with a small subset of our dataset (200 samples) can lead to a nearly 100\% performance increase (results included in the Appendix~\ref{appendix: training samples}), demonstrating the exceptional quality and utility of our dataset.
% Although \Title{} dataset has fewer images compared to previous segmentation datasets, it focuses more on the reasoning of complex instructional data and fine-grained part segments which are hard to collect and annotate. 
% As a reference, the size of \Title{} is already larger than some recent Vision-Language evaluation datasets, such as MMStar~\cite{chen2024we} (1500 samples, Arxiv '24), VisIT-Bench~\cite{bitton2024visit} (592 images, NeurIPS '23), WHOOPS!~\cite{bitton2023breaking} (500 images, ICCV '23), and TIFA160~\cite{hu2023tifa} (800 generated images, ICCV '23). 
% Besides, fine-tuning models with a small subset of our dataset (200 samples) can lead to a nearly 100\% performance increase (results included in the supplementary material), demonstrating the exceptional quality and utility of our data.

\textbf{Novelty of \Title{}.} The novelty of \Title{} lies not in our baseline method but in our comprehensive evaluation of SOTA VLMs, revealing their limitations in complex language reasoning and part-grounding.  
We hope that the established benchmark will foster progress in VLM-based part grounding, ultimately enhancing the real-world applicability of VLMs across various scenarios.  
Our proposed baseline is simple yet demonstrates the superior quality and training potential of our dataset.  
Additionally, we conduct a case study on real-world grasping data (see Appendix~\ref{appendix: case study}), showing the potential of \Title{} for broader applications.

\textbf{Potential Applications.}
Our dataset contains samples in various scenarios, including kitchen, living room, outdoor, etc., and can be used for robot manipulation and visual question answering. Besides, our dataset can provide data for affordance learning and semantic understanding. For benchmarking usage, one can also use the entire 2,400 images to evaluate current advanced VLMs.

\vspace{-2mm}
\section{Conclusion}
\vspace{-2mm}
\label{sec:conclusion and limitation}
\looseness=-1
In this work, we introduce a new benchmark, \Title{}, a novel dataset containing part annotations for common household objects as well as two tasks: task reasoning and oracle referring segmentation.
We showed that even the most advanced vision-language models struggle with tasks that link specific affordances to the corresponding parts of an object when given high-level instructions.
By fine-tuning a simple baseline with our dataset, we achieve a twofold improvement in part segmentation, showcasing the quality and training utility of our data.
% To address this issue, we introduce PISA, a strong baseline for part segmentation given only a high-level instruction of a task. 
% A case study about grasping is further conducted to demonstrate the advantage of our dataset on task-oriented grounding.
Through our work, we highlight a significant gap in foundation models for task-oriented part segmentation and hope that with our dataset, we can pave the way for further research into object-part reasoning.

% \vspace{3pt}
\textbf{Limitations.}
In this work, we propose a baseline method that achieves significant performance improvements. However, we have not fully explored the potential of our dataset, as the affordance labels were not utilized during training. An intriguing direction for future research is to combine affordance learning with language reasoning to further enhance performance.

\section*{Acknowledgements}
This work has been funded in part by the Army Research Laboratory (ARL) award W911NF-23-2-0007 and W911QX-24-F-0049, DARPA award FA8750-23-2-1015, and ONR award N00014-23-1-2840.

\bibliography{egbib}

\clearpage
\clearpage
% \appendix
\appendix

\setcounter{page}{1}

\renewcommand{\thesection}{\Alph{section}}
\renewcommand\thefigure{\Alph{section}\arabic{figure}} 
\renewcommand\thetable{\Alph{section}\arabic{table}}

\twocolumn[
\begin{center}
    \section*{Appendix}
    \addcontentsline{toc}{section}{Supplementary Material} % Add to table of contents
\end{center}
]
\setcounter{page}{1}

%%%%%%%%%%%%%%%%%%%%%%%%%%%%%%%%%%%%
%%%%%%%%For Neurips appendix
% dataset distribution
% annotation example
% pisa training curve
% more qualitative results: pisa, gpt4v
% detailed analysis of quantitative results
% 2400 images evaluation result
% grasping experiment: details and qualitative result.

% %%%%%%%%%%%%
% %%%fig
% %%%%%%%%%%%%%
\begin{figure*}[h]
  \begin{center}
    \hspace{-5mm}
     \makebox[\textwidth]{\includegraphics[width=1.05\textwidth]{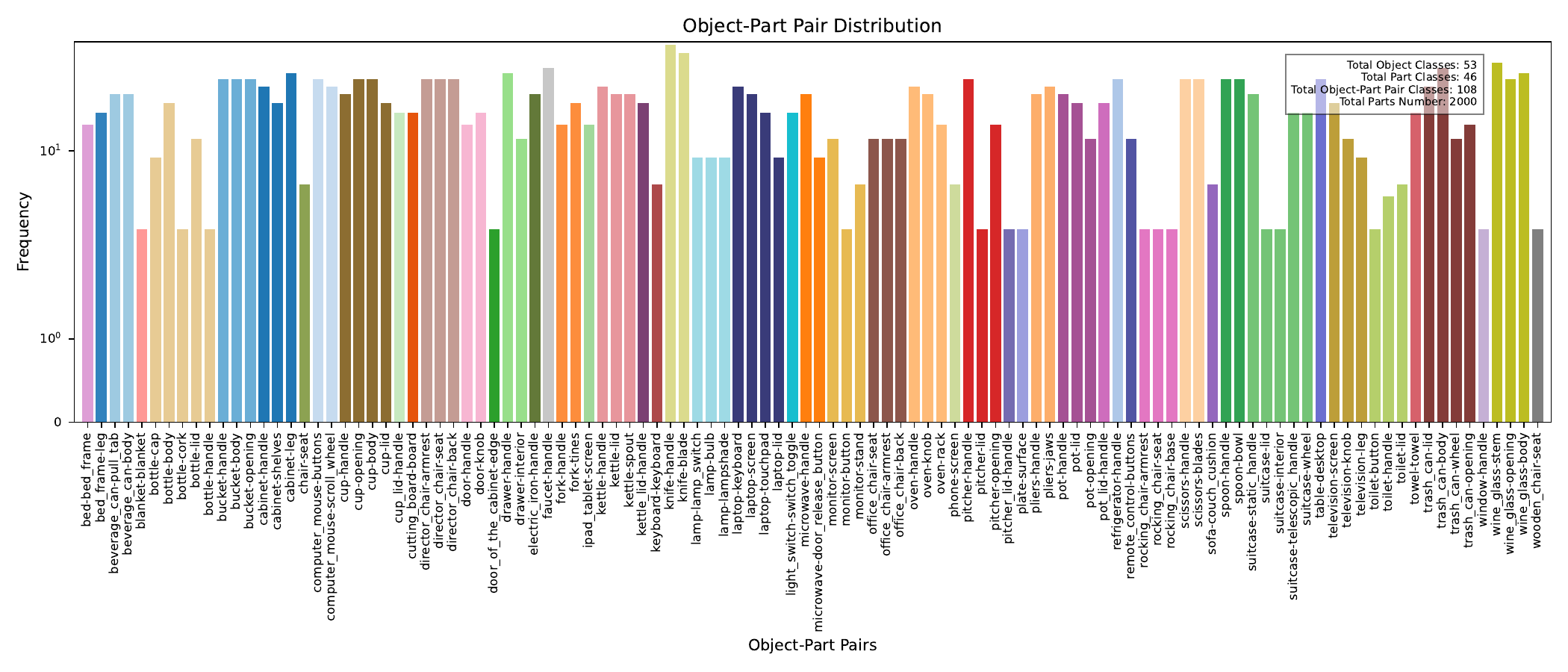}}
  \end{center}
 \vspace{-0.20in}
  \caption{Object-part pair distribution. We collect 2,400 data pieces in total, containing 48 object classes and 44 part classes, constituting 98 different object-part pair classes. The x-axis shows the name of the object-part pairs, and the y-axis shows the frequency of each item. The parts belonging to the same object classes are highlighted with the same color in the bar chart.}
  \vspace{-0.10in}
  \label{fig:dataset distribution}
\end{figure*}

\section{Dataset Details}
% \todo[inline]{update this fig}
% \vspace{-15pt}
\label{appendix: dataset detail}

\begin{figure*}[h]
  \begin{center}
     \includegraphics[width=0.8\linewidth]{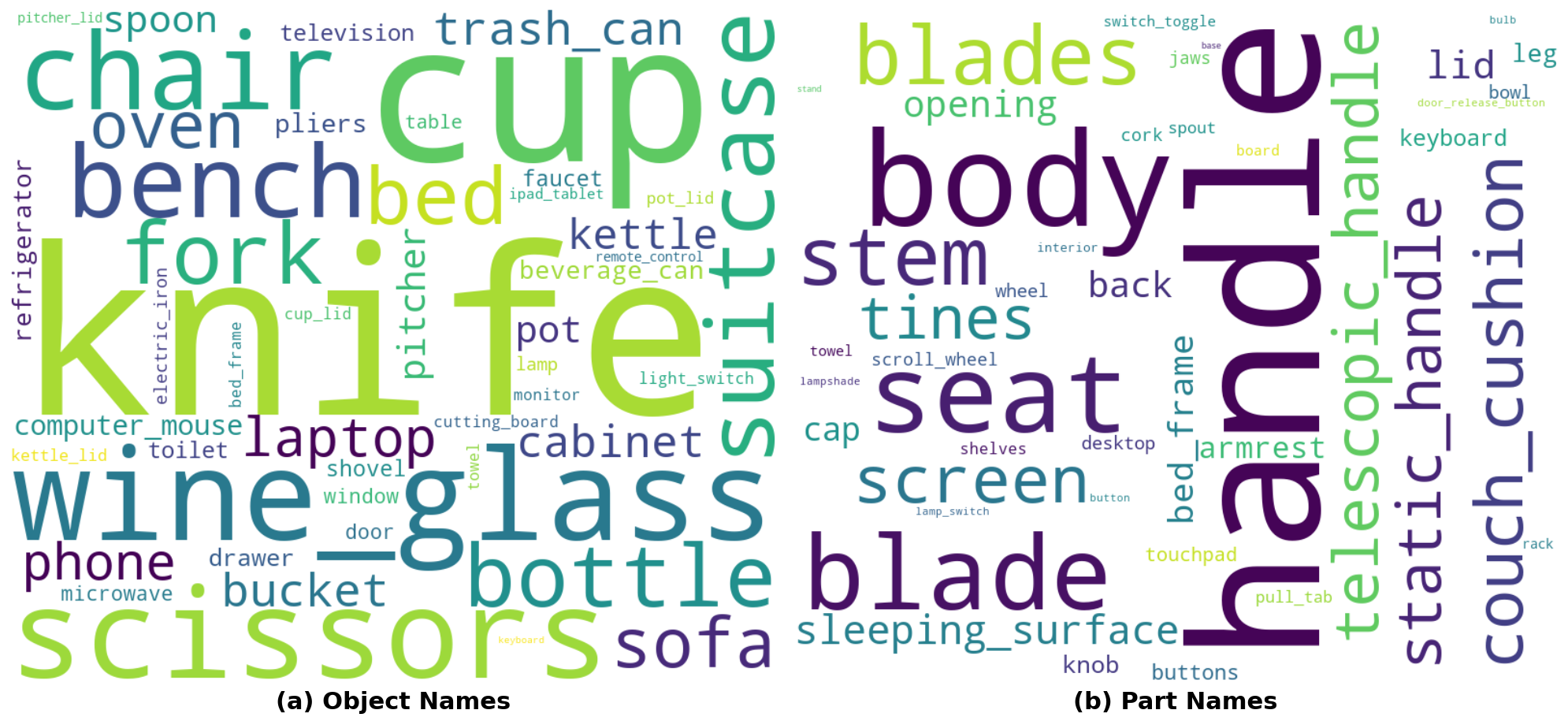}
  \end{center}
 \vspace{-0.2in}
  \caption{\Title{} dataset object and part classes. The left part shows the object class names and the right part shows the part class names.}
  \vspace{-3pt}
  \label{fig:object-part wordcloud classes}
\end{figure*}

\begin{figure*}[h]
  \begin{center}
     \includegraphics[width=0.8\linewidth]{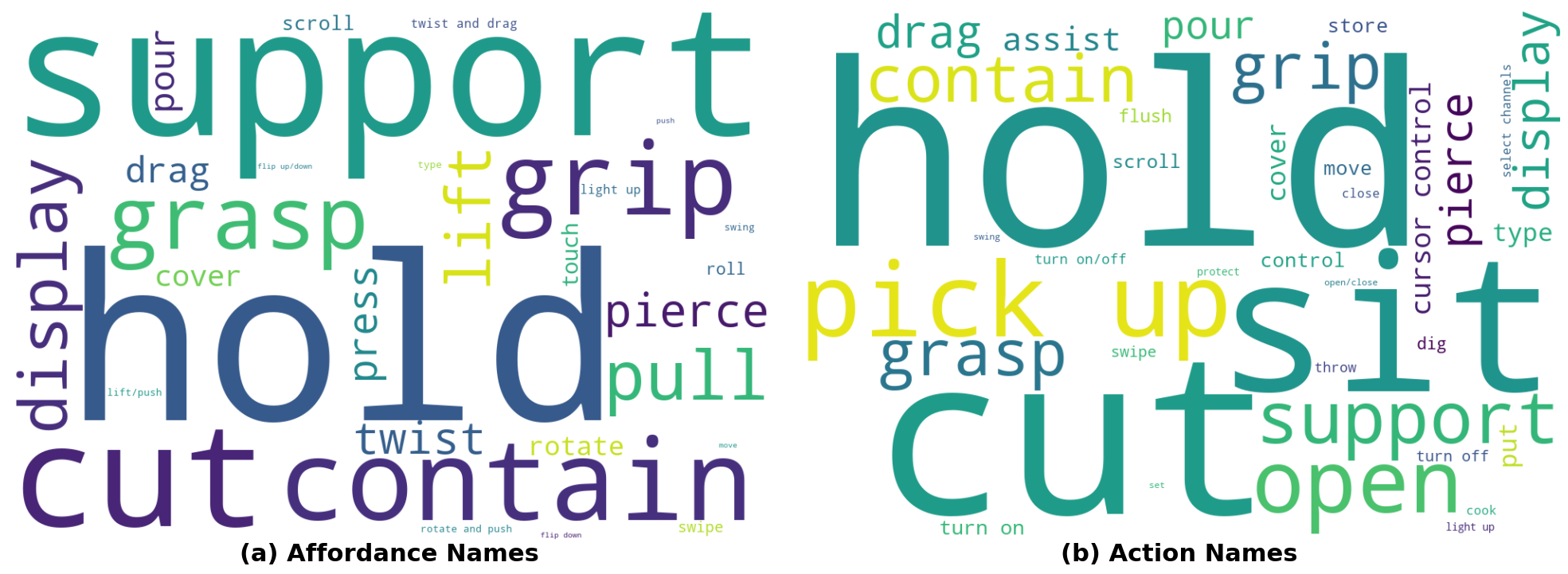}
  \end{center}
 \vspace{-0.2in}
  \caption{\Title{} dataset affordance and action categories. The left part shows the affordance names and the right part shows the action names. Specifically, affordances refer to low-level actions performed to a specific part, while actions refer to the high-level function to be achieved.}
  \vspace{-3pt}
  \label{fig:affordance-action wordcloud classes}
\end{figure*}

\Title{} dataset is collected from Flickr\footnote{https://www.flickr.com/} website and AGD20K~\cite{Luo_2022_CVPR}, where we selected free-licensed images from both sources. To better understand the categories of our dataset, we follow ADE20K~\cite{zhou2019semantic} to provide the distribution of objects and parts within \Title{}. As shown in Fig.~\ref{fig:dataset distribution}, the dataset comprises 2,400 data items, encompassing 48 object classes and 44 part classes, which together form 98 distinct object-part pair classes. Besides, we also provide a word cloud to visualize the object-part classes and affordance-action categories, as depicted in Fig.~\ref{fig:object-part wordcloud classes} and Fig.~\ref{fig:affordance-action wordcloud classes}, respectively. This diversity in classes indicates our dataset's wide coverage of various daily scenes, offering robust criteria for comprehensively analyzing the proficiency of current models in understanding task instructions and segmenting parts. Furthermore, this suggests that our dataset can be valuable for broad areas, including semantic segmentation, robot manipulation, visual question answering, and more.

% %%%%%%%%%%%%
% %%%fig
% %%%%%%%%%%%%%

% %%%%%%%%%%%%
% %%%fig
% %%%%%%%%%%%%%

\section{Annotation Example}
\label{appendix: annotation example}
% %%%%%%%%%%%%
% %%%fig
% %%%%%%%%%%%%%
\begin{figure*}[h]
  \begin{center}
     \includegraphics[width=0.996\linewidth]{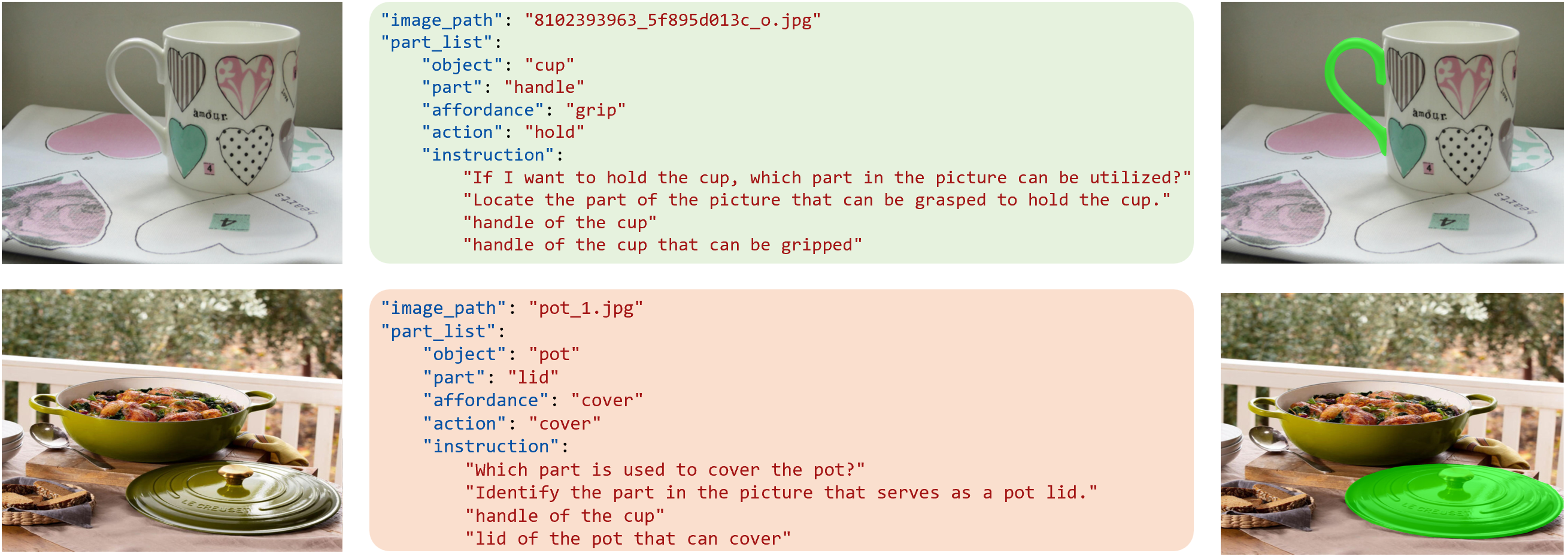}
  \end{center}
 \vspace{-0.16in}
  \caption{Annotation Example: Each data item is represented by a JSON dictionary, which details the components involved. This includes the object to which these parts belong, the name of each part, a specific instruction related to these parts, a low-level affordance associated with the instruction, and a high-level action performed on the parts. Corresponding parts are highlighted in green in the images on the right.}
  \vspace{0.15in}
  \label{fig:annotation example}
\end{figure*}
Fig.~\ref{fig:annotation example} presents two examples of annotations from our \Title{} dataset, focusing on the handle of a cup and the lid of a pod, respectively. In each JSON dictionary, the names of the object and its specific part are noted, aligned with a task instruction that pertains to a particular part shown in the image. Additionally, both a low-level affordance name and a high-level action name are provided in relation to the instruction.

\begin{figure*}[h]
    \centering
    \vspace{-3mm}
    \includegraphics[width=0.90\linewidth]{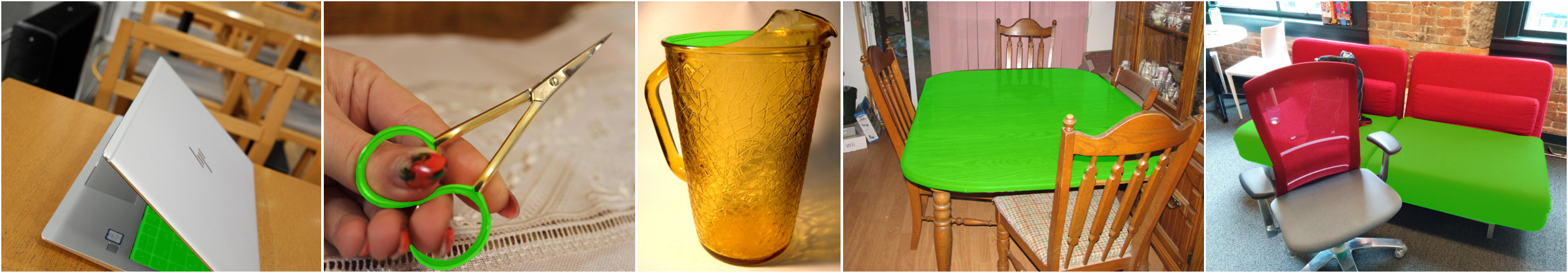}
    \vspace{-5pt}
    \caption{More complex examples in \Title{}, including occlusions and human-object interactions.}
    \vspace{-3pt}
    \label{fig:complex scenes} 
\end{figure*}

Besides, in Fig.~\ref{fig:complex scenes}, we provide more examples that contain occlusions and human interactions to showcase the complexity of our dataset.
%
% However, human-part interaction data is not specifically collected as \Title{} is designed from the viewpoint of robots.

\section{Evaluated Model Details}
\label{appenix: model details}
\textbf{Open-vocabulary segmentation models.}
We choose OVSeg~\cite{liang2023open} and SAN~\cite{xu2023side} to discover the performance of the open-vocabulary object segmentation methods on our task.
We select the best-reported models for the two methods, \textit{ovseg\_swinbase\_vitL14\_ft\_mpt.pth} and \textit{san\_vit\_large\_14.pth} respectively.

% \noindent
\textbf{Refering expression segmentation.}
We conduct experiments with off-the-shelf models including X-Decoder~\cite{zou2023generalized}, SEEM~\cite{zou2023segment}, and TRIS~\cite{liu2023referring}. 
We adopt \textit{xdecoder\_focalt\_last.pt}, \textit{seem\_focall\_v1.pt}, and \textit{stage2\_refcocog\_google.pth} for the three models respectively.
Besides, we also evaluate Grounding-DINO~\cite{liu2023grounding}, which has witnessed a great open-vocabulary referring detection ability and been integrated with SAM~\cite{kirillov2023segment} to a project, Grounded-SAM\footnote{https://github.com/IDEA-Research/Grounded-Segment-Anything}.

% \noindent
\textbf{Reasoning segmentation.}
For our tasks, LISA~\cite{lai2023lisa} can naturally be a good choice since it can return masks and has been trained on several part segmentation datasets. As a result, it is interesting to explore whether it possesses the ability to understand instructions and find part segments.
Other multi-modal LLMs, including VisionLLM~\cite{wang2023visionllm}, Shikra~\cite{chen2023shikra},
% and MiniGPT-v2~\cite{chen2023minigpt}, 
also have localization ability. 
Since they can only return bounding box outputs, we use the results as box prompts for SAM to get a mask output for fair comparison.
However, we cannot test on VisionLLM since it has not release code.

To prompt LISA, we follow its original setting to add \textit{"Please output the segmentation mask."} at the end of each instruction.
Besides, in order to formulate a query for the oracle referring part segmentation task, we embed the object and part name in a format of: ``Where is the {$I_\text{text}$} in the image'', where $I_\text{text}$ stands for the text input.

To prompt Shikra, we integrate our instruction in its original template as follows:
\begin{itemize}
    \item Instruction referring part segmentation: \\ \textless $I_{\text{text}}$\textgreater. Can you point out all the related parts in the image \textless $I_{\text{image}}$\textgreater \, and provide the coordinates of their locations?
    \item Oracle referring part segmentation: \\Can you point out all the \textless $I_{\text{text}}$\textgreater \, in the image \textless $I_{\text{image}}$\textgreater \, and provide the coordinates of their locations?
\end{itemize}

We adopt LISA-7B-v1~\cite{lai2023lisa} model that has been fine-tuned on both training and validation data of LISA's dataset.
As for Shikra, we select the frequently updated model, Shikra-7B-delta-v1-0708.

\clearpage
\clearpage
\section{Effect of Training Samples}
\label{appendix: training samples}
\begin{figure}[h]
    \centering
    % \vspace{-3mm}
    \includegraphics[width=\linewidth]{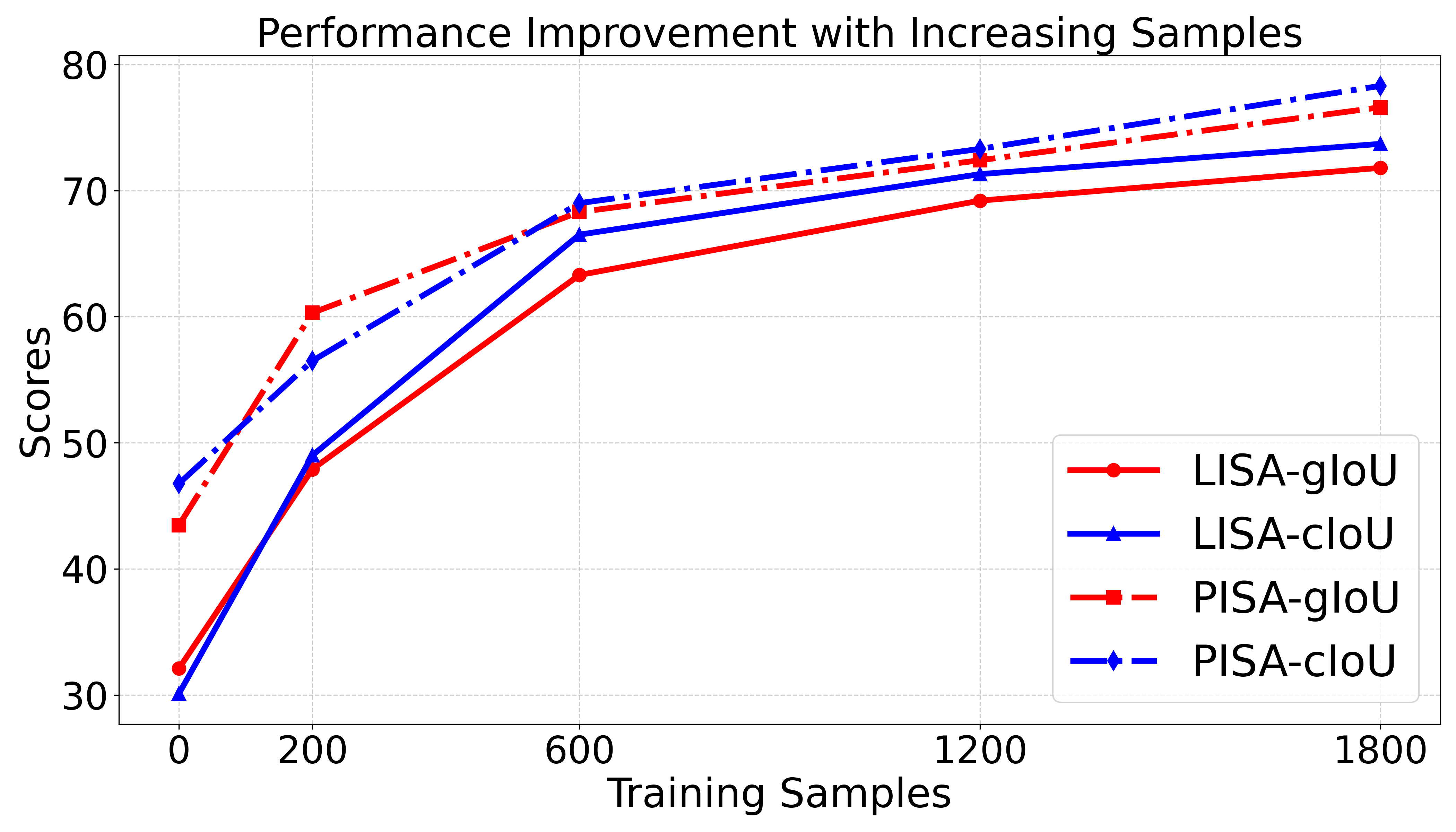}
    \vspace{-17pt}
    \caption{Performance improvement with increasing number of training samples. We gradually add training samples to 200, 600, 1,200, and 1,800.}
    % \vspace{-3pt}
    \label{fig:more samples higher iou} 
\end{figure}

To verify the quality and training potential of the PISA dataset, we gradually increase the number of training samples from 200 to 1,800 and observe the performance improvement. Specifically, we start with 200 samples for training, then gradually increase the number of training samples to 600, 1,200, and finally 1,800. Each increment includes all the previously used training samples. As shown in Fig.~\ref{fig:more samples higher iou}, with the increasing number of training samples, the IoU metric gradually increases and exhibits a logarithmic convergence tendency. This indicates that our high-quality data significantly boosts performance, even with just 200 samples. The performance of both models improves substantially from the outset.

\section{Does object recognition hinder part segmentation?}
To explore whether the bottleneck lies in current VLMs' object recognition ability, we use the object classes as instruction and obtain the results in Tab.~\ref{tab:recall}.
Since we do not have object-level labels, we use the recall rate as a reflection of whether the model can find the entire object.
From the results in Tab.~\ref{tab:recall}, the precision is much lower compared to the recall rate, and the recall rate is close to 1 after the third quartile (75th percentile).
This indicates that the predicted masks can generally cover the part labels, so the poor performance of TRPS cannot derive from the object recognition ability.
\begin{table}[h]
  \centering
  \footnotesize
  % \vspace{-5pt}
  % \vspace{-3.6mm}
  \centering
     \setlength{\tabcolsep}{1.0mm}{
     \resizebox{\linewidth}{!}{
     \begin{tabular}{c|ccccc}
     \toprule
      % \specialrule{.1em}{.05em}{.05em} 
      \multirow{2}{*}[-0.8ex]{Methods}& \multicolumn{5}{c}{Object-Level} \\
      \cmidrule{2-6}
              & Prec.  & Rec.@A  & Rec.@25\%   & Rec.@50\% & Rec.@75\%  \\
        \midrule\midrule
        OVSeg   & 20.93 & 81.80 & 85.00 & 99.26 & 100.00 \\
        SAN  & 19.46 & 73.55 & 56.37 & 98.49 & 100.00 \\
        G-SAM  & 25.53 & 89.83 & 92.59 & 96.99 & 99.37 \\
        % \specialrule{.1em}{.05em}{.05em} 
        \bottomrule
\end{tabular}}}
\vspace{-4pt}
  \caption{Precision and recall rate on object-level segmentation results. The five metrics refer to precision (Prec.), average recall (Rec.@A), first quartile recall (Rec.@25\%), median recall (Rec.@50\%), and third quartile recall (Rec.@75\%), respectively.}
  \label{tab:recall}
\end{table}

\vspace{20mm}
\section{GPT-4V Qualitative Results}
\begin{figure}[h]
    \centering
    % \vspace{-3mm}
    \includegraphics[width=\linewidth]{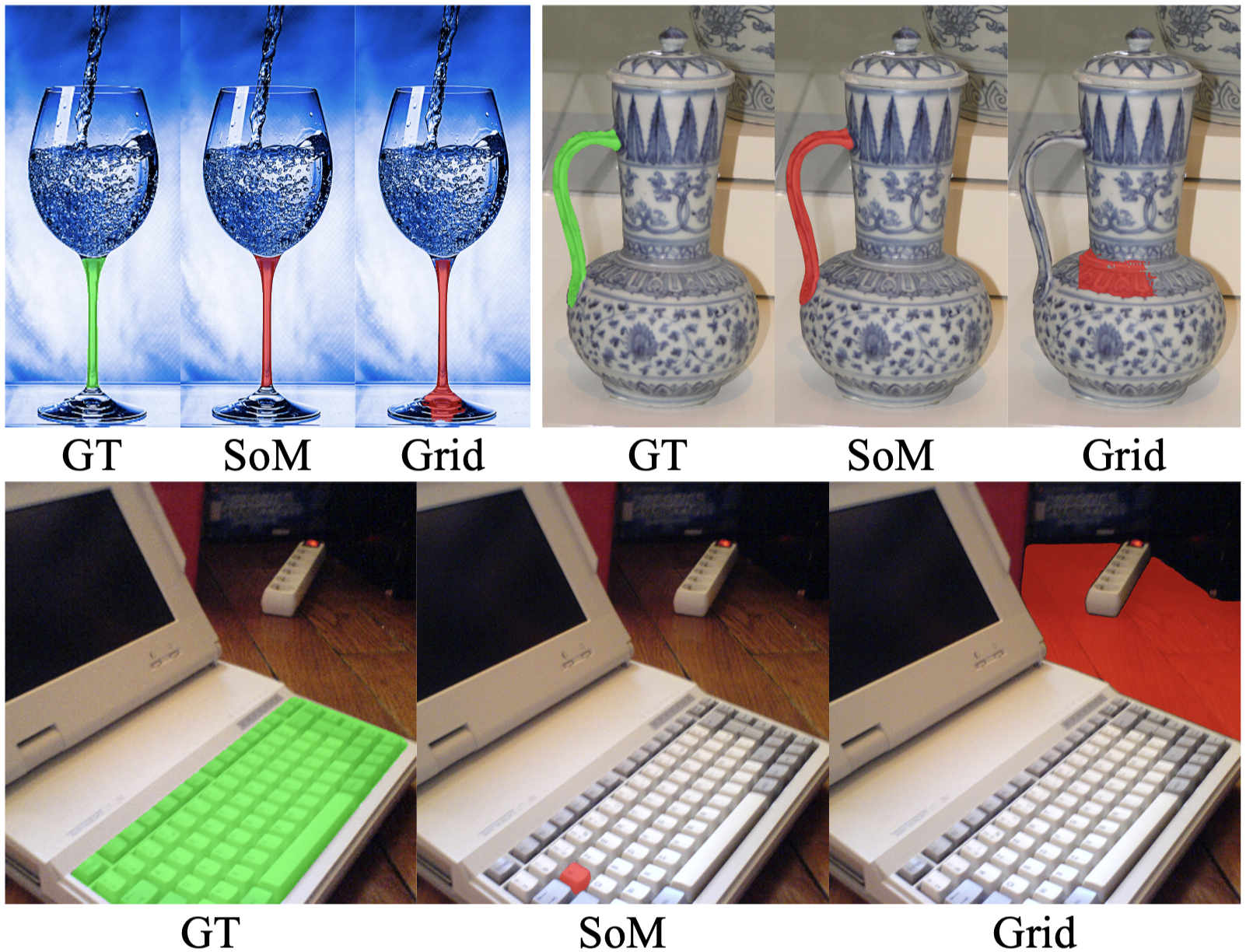} % 插入图像，并调整宽度
    \vspace{-17pt}
    \caption{GPT4-V based methods.} % 图像的标题
    \label{fig:gpt4v} % 用于引用的标签
   % \vspace{-10pt}
\end{figure}

We show the results of GPT-4V-based methods, namely SoM-based GPT-4V and Grid-based GPT-4V, in Fig.~\ref{fig:gpt4v}. While GPT-4V-based methods deliver clear boundaries, they sometimes select the wrong segments from SAM~\cite{kirillov2023segment}, leading to poor overall performance.

\clearpage
\clearpage
\section{More Qualitative Results}
\label{appendix: more qualitative}
\begin{figure*}[h]
	\centering
	% \vspace{-4mm}
	\resizebox{1\textwidth}{!}
	{
		\begin{tabular}{@{}c@{}c@{}c@{}c@{}c@{}c@{}c@{}c@{}c@{}c@{}c@{}c}

\vspace{-2mm}
\includegraphics[width=1.8cm,height=1.8cm]{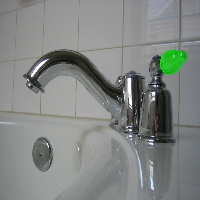}\hspace{1mm} &
\includegraphics[width=1.8cm,height=1.8cm]{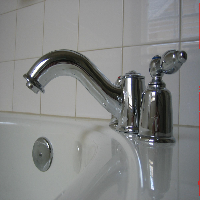}\hspace{1mm} &
\includegraphics[width=1.8cm,height=1.8cm]{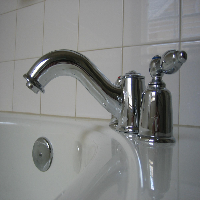}\hspace{1mm} &
\includegraphics[width=1.8cm,height=1.8cm]{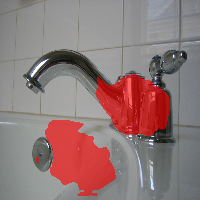}\hspace{1mm} &
\includegraphics[width=1.8cm,height=1.8cm]{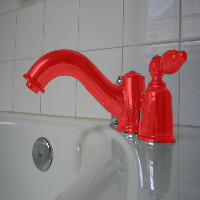}\hspace{1mm} &
\includegraphics[width=1.8cm,height=1.8cm]{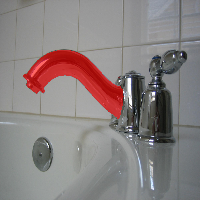}\hspace{1mm} &
\includegraphics[width=1.8cm,height=1.8cm]{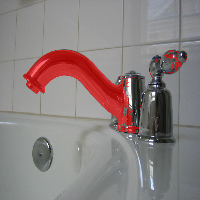}\hspace{1mm} &
\includegraphics[width=1.8cm,height=1.8cm]{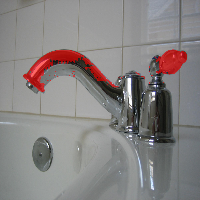}\hspace{1mm} &
\includegraphics[width=1.8cm,height=1.8cm]{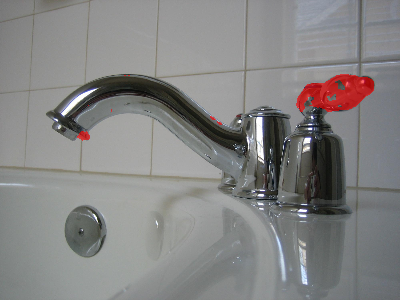}\hspace{1mm} \\
\vspace{-2mm}
\includegraphics[width=1.8cm,height=1.8cm]{figs/selected_imgs_neurips/gt/2329134125_8a71be7470_o-kettle-handle.png}\hspace{1mm} &
\includegraphics[width=1.8cm,height=1.8cm]{figs/selected_imgs_neurips/pred_mask_human_xdecoder/2329134125_8a71be7470_o-kettle-handle.png}\hspace{1mm} &
\includegraphics[width=1.8cm,height=1.8cm]{figs/selected_imgs_neurips/pred_mask_human_seem/2329134125_8a71be7470_o-kettle-handle.png}\hspace{1mm} &
\includegraphics[width=1.8cm,height=1.8cm]{figs/selected_imgs_neurips/pred_mask_human_tris/2329134125_8a71be7470_o-kettle-handle.png}\hspace{1mm} &
\includegraphics[width=1.8cm,height=1.8cm]{figs/selected_imgs_neurips/pred_mask_human_groundedsam/2329134125_8a71be7470_o-kettle-handle.png}\hspace{1mm} &
\includegraphics[width=1.8cm,height=1.8cm]{figs/selected_imgs_neurips/pred_mask_human_minigpt/2329134125_8a71be7470_o-kettle-handle.png}\hspace{1mm} &
\includegraphics[width=1.8cm,height=1.8cm]{figs/selected_imgs_neurips/pred_lisa-untrain-test/2329134125_8a71be7470_o-kettle-handle.png}\hspace{1mm} &
\includegraphics[width=1.8cm,height=1.8cm]{figs/selected_imgs_neurips/pred_lisa-train1800/2329134125_8a71be7470_o-kettle-handle.png}\hspace{1mm} &
\includegraphics[width=1.8cm,height=1.8cm]{figs/selected_imgs_neurips/pred_pisa_pretrain-v1-dinodecoder-train1800/2329134125_8a71be7470_o-kettle-handle.png}\hspace{1mm} \\
 \vspace{-2mm}

\includegraphics[width=1.8cm,height=1.8cm]{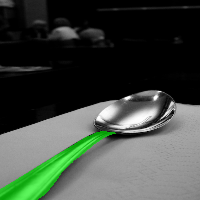}\hspace{1mm} &
\includegraphics[width=1.8cm,height=1.8cm]{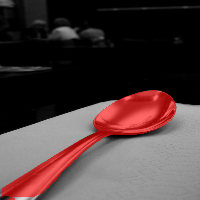}\hspace{1mm} &
\includegraphics[width=1.8cm,height=1.8cm]{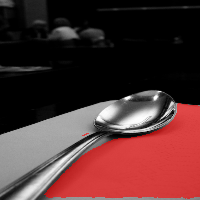}\hspace{1mm} &
\includegraphics[width=1.8cm,height=1.8cm]{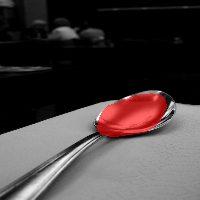}\hspace{1mm} &
\includegraphics[width=1.8cm,height=1.8cm]{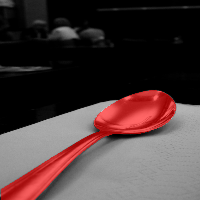}\hspace{1mm} &
\includegraphics[width=1.8cm,height=1.8cm]{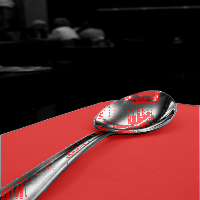}\hspace{1mm} &
\includegraphics[width=1.8cm,height=1.8cm]{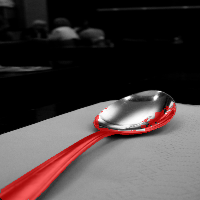}\hspace{1mm} &
\includegraphics[width=1.8cm,height=1.8cm]{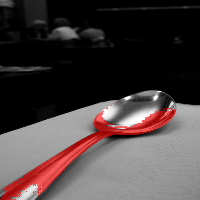}\hspace{1mm} &
\includegraphics[width=1.8cm,height=1.8cm]{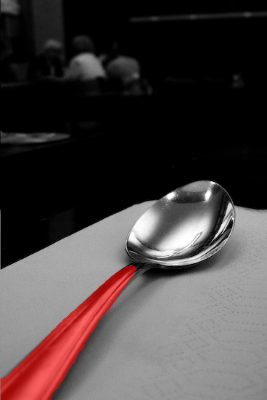}\hspace{1mm} \\
 \vspace{-2mm}

\includegraphics[width=1.8cm,height=1.8cm]{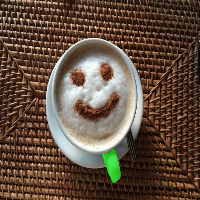}\hspace{1mm} &
\includegraphics[width=1.8cm,height=1.8cm]{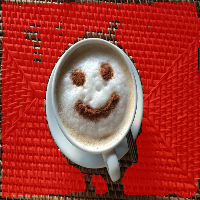}\hspace{1mm} &
\includegraphics[width=1.8cm,height=1.8cm]{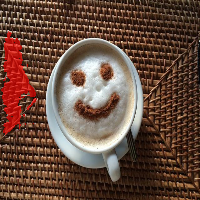}\hspace{1mm} &
\includegraphics[width=1.8cm,height=1.8cm]{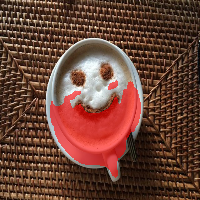}\hspace{1mm} &
\includegraphics[width=1.8cm,height=1.8cm]{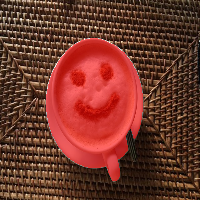}\hspace{1mm} &
\includegraphics[width=1.8cm,height=1.8cm]{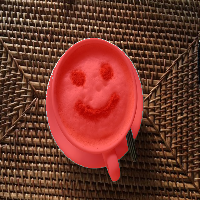}\hspace{1mm} &
\includegraphics[width=1.8cm,height=1.8cm]{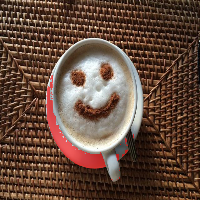}\hspace{1mm} &
\includegraphics[width=1.8cm,height=1.8cm]{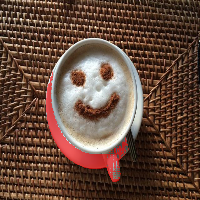}\hspace{1mm} &
\includegraphics[width=1.8cm,height=1.8cm]{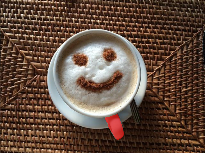}\hspace{1mm} \\
 \vspace{-2mm}

\includegraphics[width=1.8cm,height=1.8cm]{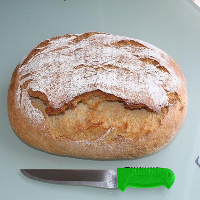}\hspace{1mm} &
\includegraphics[width=1.8cm,height=1.8cm]{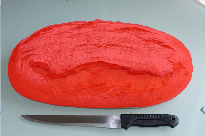}\hspace{1mm} &
\includegraphics[width=1.8cm,height=1.8cm]{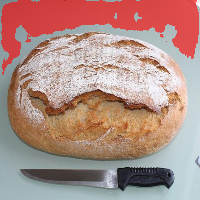}\hspace{1mm} &
\includegraphics[width=1.8cm,height=1.8cm]{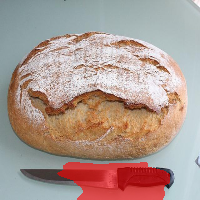}\hspace{1mm} &
\includegraphics[width=1.8cm,height=1.8cm]{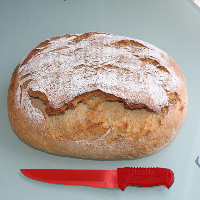}\hspace{1mm} &
\includegraphics[width=1.8cm,height=1.8cm]{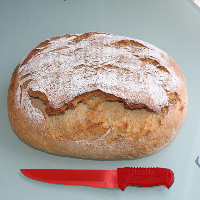}\hspace{1mm} &
\includegraphics[width=1.8cm,height=1.8cm]{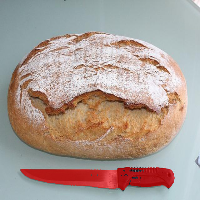}\hspace{1mm} &
\includegraphics[width=1.8cm,height=1.8cm]{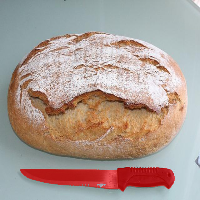}\hspace{1mm} &
\includegraphics[width=1.8cm,height=1.8cm]{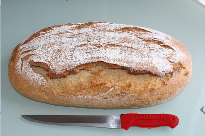}\hspace{1mm} \\
 \vspace{-2mm}

\includegraphics[width=1.8cm,height=1.8cm]{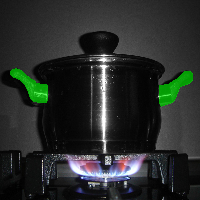}\hspace{1mm} &
\includegraphics[width=1.8cm,height=1.8cm]{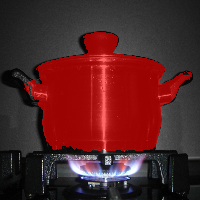}\hspace{1mm} &
\includegraphics[width=1.8cm,height=1.8cm]{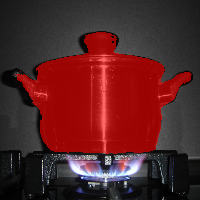}\hspace{1mm} &
\includegraphics[width=1.8cm,height=1.8cm]{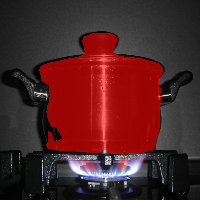}\hspace{1mm} &
\includegraphics[width=1.8cm,height=1.8cm]{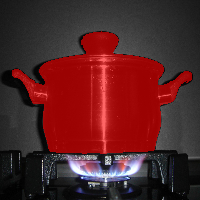}\hspace{1mm} &
\includegraphics[width=1.8cm,height=1.8cm]{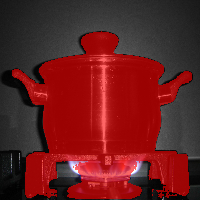}\hspace{1mm} &
\includegraphics[width=1.8cm,height=1.8cm]{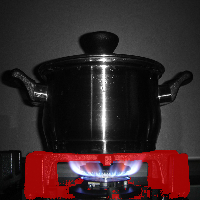}\hspace{1mm} &
\includegraphics[width=1.8cm,height=1.8cm]{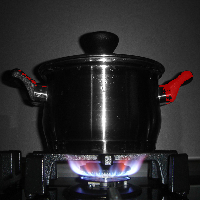}\hspace{1mm} &
\includegraphics[width=1.8cm,height=1.8cm]{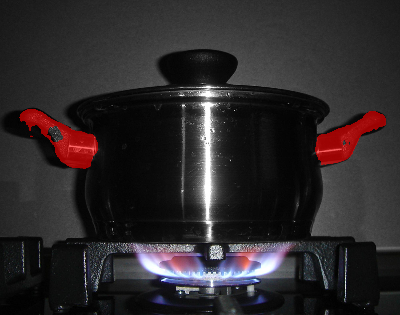}\hspace{1mm} \\
 \vspace{-2mm}

\includegraphics[width=1.8cm,height=1.8cm]{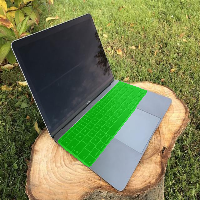}\hspace{1mm} &
\includegraphics[width=1.8cm,height=1.8cm]{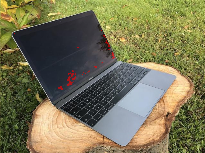}\hspace{1mm} &
\includegraphics[width=1.8cm,height=1.8cm]{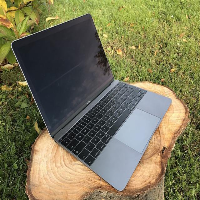}\hspace{1mm} &
\includegraphics[width=1.8cm,height=1.8cm]{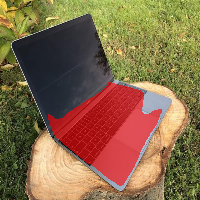}\hspace{1mm} &
\includegraphics[width=1.8cm,height=1.8cm]{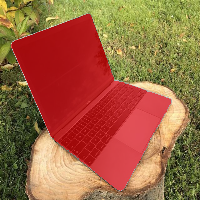}\hspace{1mm} &
\includegraphics[width=1.8cm,height=1.8cm]{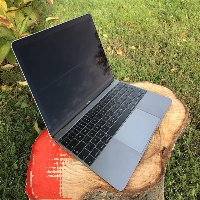}\hspace{1mm} &
\includegraphics[width=1.8cm,height=1.8cm]{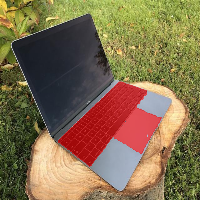}\hspace{1mm} &
\includegraphics[width=1.8cm,height=1.8cm]{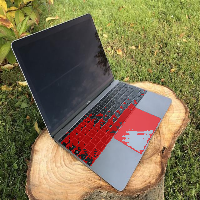}\hspace{1mm} &
\includegraphics[width=1.8cm,height=1.8cm]{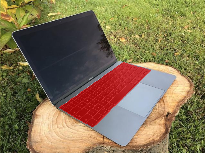}\hspace{1mm} \\
\vspace{-2mm}

 \includegraphics[width=1.8cm,height=1.8cm]{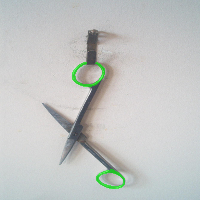}\hspace{1mm} &
			\includegraphics[width=1.8cm,height=1.8cm]{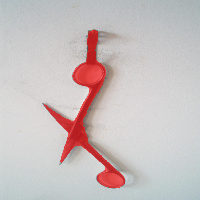}\hspace{1mm} &
			\includegraphics[width=1.8cm,height=1.8cm]{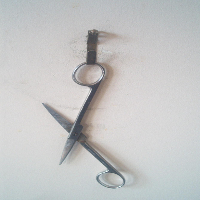}\hspace{1mm} &
			\includegraphics[width=1.8cm,height=1.8cm]{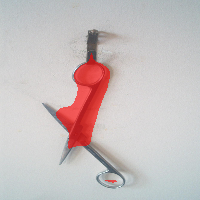}\hspace{1mm} &
			\includegraphics[width=1.8cm,height=1.8cm]{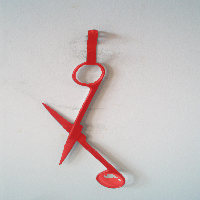}\hspace{1mm} &
			\includegraphics[width=1.8cm,height=1.8cm]{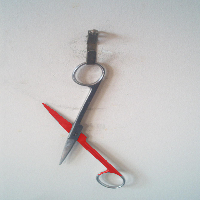}\hspace{1mm} &
			\includegraphics[width=1.8cm,height=1.8cm]{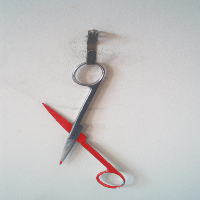}\hspace{1mm} &
                \includegraphics[width=1.8cm,height=1.8cm]{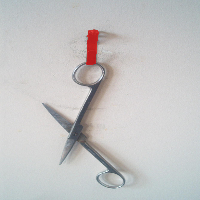}\hspace{1mm} &
			\includegraphics[width=1.8cm,height=1.8cm]{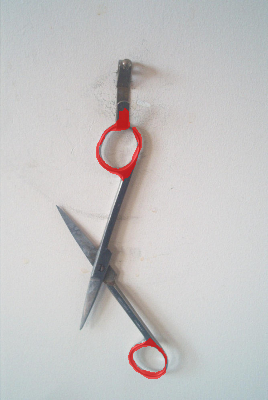}\hspace{1mm} \\
   \vspace{-2mm}
			%-----------------------------------------------------------------------------------------
			\includegraphics[width=1.8cm,height=1.8cm]{figs/selected_imgs_neurips/gt/knife_002845-knife-handle.png}\hspace{1mm} &
                \includegraphics[width=1.8cm,height=1.8cm]{figs/selected_imgs_neurips/pred_mask_human_xdecoder/knife_002845-knife-handle.png}\hspace{1mm} &
			\includegraphics[width=1.8cm,height=1.8cm]{figs/selected_imgs_neurips/pred_mask_human_seem/knife_002845-knife-handle.png}\hspace{1mm} &
			\includegraphics[width=1.8cm,height=1.8cm]{figs/selected_imgs_neurips/pred_mask_human_tris/knife_002845-knife-handle.png}\hspace{1mm} &
			\includegraphics[width=1.8cm,height=1.8cm]{figs/selected_imgs_neurips/pred_mask_human_groundedsam/knife_002845-knife-handle.png}\hspace{1mm} &
			\includegraphics[width=1.8cm,height=1.8cm]{figs/selected_imgs_neurips/pred_mask_human_minigpt/knife_002845-knife-handle.png}\hspace{1mm} &
			\includegraphics[width=1.8cm,height=1.8cm]{figs/selected_imgs_neurips/pred_lisa-untrain-test/knife_002845-knife-handle.png}\hspace{1mm} &
   			\includegraphics[width=1.8cm,height=1.8cm]{figs/selected_imgs_neurips/pred_lisa-train1800/knife_002845-knife-handle.png}\hspace{1mm} &
			\includegraphics[width=1.8cm,height=1.8cm]{figs/selected_imgs_neurips/pred_pisa_pretrain-v1-dinodecoder-train1800/knife_002845-knife-handle.png}\hspace{1mm} \\
   
\vspace{-2mm}

\includegraphics[width=1.8cm,height=1.8cm]{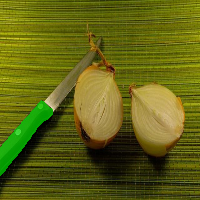}\hspace{1mm} &
\includegraphics[width=1.8cm,height=1.8cm]{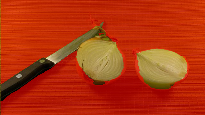}\hspace{1mm} &
\includegraphics[width=1.8cm,height=1.8cm]{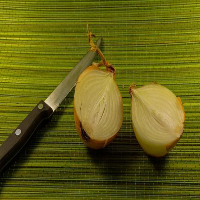}\hspace{1mm} &
\includegraphics[width=1.8cm,height=1.8cm]{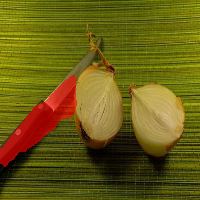}\hspace{1mm} &
\includegraphics[width=1.8cm,height=1.8cm]{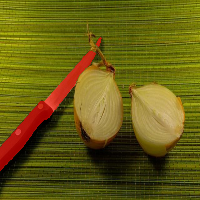}\hspace{1mm} &
\includegraphics[width=1.8cm,height=1.8cm]{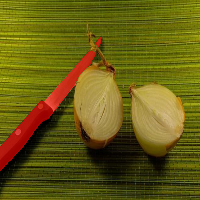}\hspace{1mm} &
\includegraphics[width=1.8cm,height=1.8cm]{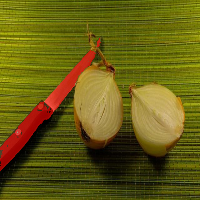}\hspace{1mm} &
\includegraphics[width=1.8cm,height=1.8cm]{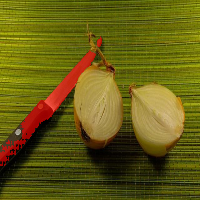}\hspace{1mm} &
\includegraphics[width=1.8cm,height=1.8cm]{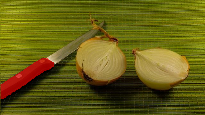}\hspace{1mm} &\\[1pt]
                %-----------------------------------------------------------------------------------------
			{\scriptsize Ground Truth} & {\scriptsize X-Decoder} & {\scriptsize SEEM} & {\scriptsize TRIS} & {\scriptsize G-SAM} & {\scriptsize MiniGPT-v2} & {\scriptsize LISA-Pretrain} & {\scriptsize LISA-Finetune} & {\scriptsize \textcolor{red}{PISA-Finetune}}\ \\
		\end{tabular}
	}
	\vspace{-3mm}
	\caption{Qualitative comparison of different VLMs and the fine-tuned models. In these examples, the pre-trained LISA falls short of recognizing the correct part. After fine-tuning, PISA shows better potential for part understanding than LISA.}
	\label{fig:qualitative supplementary results 3}
	\vspace{-5mm}
\end{figure*}

\begin{figure*}[h]
	\centering
	\vspace{-4mm}
	\resizebox{1\textwidth}{!}
	{
		\begin{tabular}{@{}c@{}c@{}c@{}c@{}c@{}c@{}c@{}c@{}c@{}c@{}c@{}c}
 \vspace{-2mm}
\includegraphics[width=1.8cm,height=1.8cm]{figs/selected_imgs_neurips/gt/4178009615_ed8921d0d1_k-kettle-spout.png}\hspace{1mm} &
\includegraphics[width=1.8cm,height=1.8cm]{figs/selected_imgs_neurips/pred_mask_human_xdecoder/4178009615_ed8921d0d1_k-kettle-spout.png}\hspace{1mm} &
\includegraphics[width=1.8cm,height=1.8cm]{figs/selected_imgs_neurips/pred_mask_human_seem/4178009615_ed8921d0d1_k-kettle-spout.png}\hspace{1mm} &
\includegraphics[width=1.8cm,height=1.8cm]{figs/selected_imgs_neurips/pred_mask_human_tris/4178009615_ed8921d0d1_k-kettle-spout.png}\hspace{1mm} &
\includegraphics[width=1.8cm,height=1.8cm]{figs/selected_imgs_neurips/pred_mask_human_groundedsam/4178009615_ed8921d0d1_k-kettle-spout.png}\hspace{1mm} &
\includegraphics[width=1.8cm,height=1.8cm]{figs/selected_imgs_neurips/pred_mask_human_minigpt/4178009615_ed8921d0d1_k-kettle-spout.png}\hspace{1mm} &
\includegraphics[width=1.8cm,height=1.8cm]{figs/selected_imgs_neurips/pred_lisa-untrain-test/4178009615_ed8921d0d1_k-kettle-spout.png}\hspace{1mm} &
\includegraphics[width=1.8cm,height=1.8cm]{figs/selected_imgs_neurips/pred_lisa-train1800/4178009615_ed8921d0d1_k-kettle-spout.png}\hspace{1mm} &
\includegraphics[width=1.8cm,height=1.8cm]{figs/selected_imgs_neurips/pred_pisa_pretrain-v1-dinodecoder-train1800/4178009615_ed8921d0d1_k-kettle-spout.png}\hspace{1mm} \\
 \vspace{-2mm}

\includegraphics[width=1.8cm,height=1.8cm]{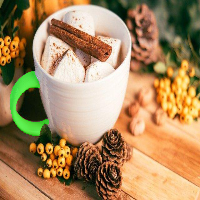}\hspace{1mm} &
\includegraphics[width=1.8cm,height=1.8cm]{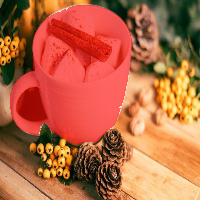}\hspace{1mm} &
\includegraphics[width=1.8cm,height=1.8cm]{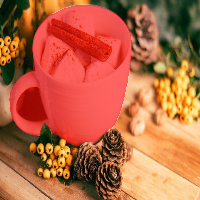}\hspace{1mm} &
\includegraphics[width=1.8cm,height=1.8cm]{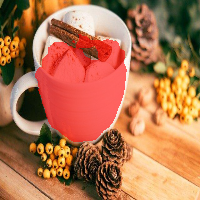}\hspace{1mm} &
\includegraphics[width=1.8cm,height=1.8cm]{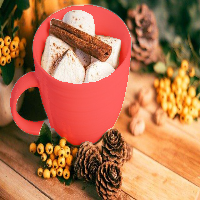}\hspace{1mm} &
\includegraphics[width=1.8cm,height=1.8cm]{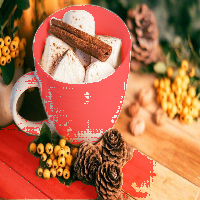}\hspace{1mm} &
\includegraphics[width=1.8cm,height=1.8cm]{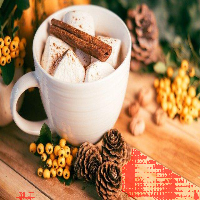}\hspace{1mm} &
\includegraphics[width=1.8cm,height=1.8cm]{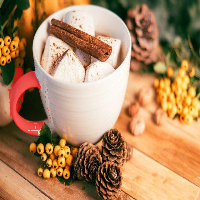}\hspace{1mm} &
\includegraphics[width=1.8cm,height=1.8cm]{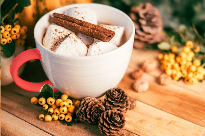}\hspace{1mm} \\
 
\vspace{-2mm}
\includegraphics[width=1.8cm,height=1.8cm]{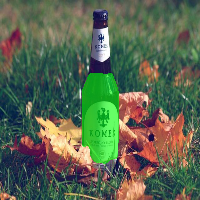}\hspace{1mm} &
\includegraphics[width=1.8cm,height=1.8cm]{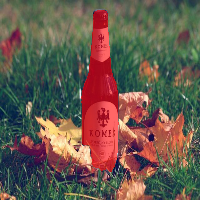}\hspace{1mm} &
\includegraphics[width=1.8cm,height=1.8cm]{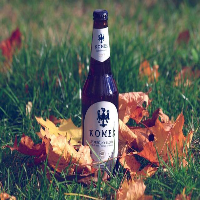}\hspace{1mm} &
\includegraphics[width=1.8cm,height=1.8cm]{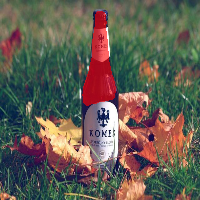}\hspace{1mm} &
\includegraphics[width=1.8cm,height=1.8cm]{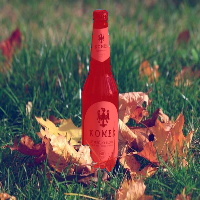}\hspace{1mm} &
\includegraphics[width=1.8cm,height=1.8cm]{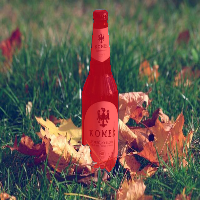}\hspace{1mm} &
\includegraphics[width=1.8cm,height=1.8cm]{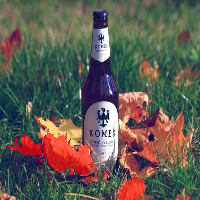}\hspace{1mm} &
\includegraphics[width=1.8cm,height=1.8cm]{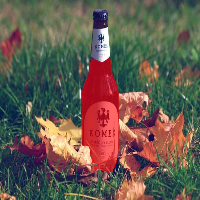}\hspace{1mm} &
\includegraphics[width=1.8cm,height=1.8cm]{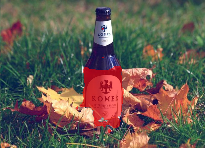}\hspace{1mm} \\
 
\vspace{-2mm}
\includegraphics[width=1.8cm,height=1.8cm]{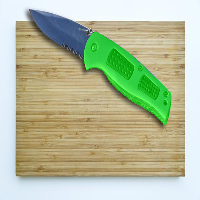}\hspace{1mm} &
\includegraphics[width=1.8cm,height=1.8cm]{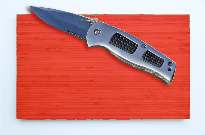}\hspace{1mm} &
\includegraphics[width=1.8cm,height=1.8cm]{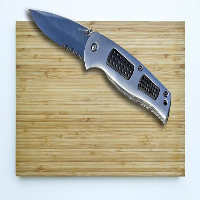}\hspace{1mm} &
\includegraphics[width=1.8cm,height=1.8cm]{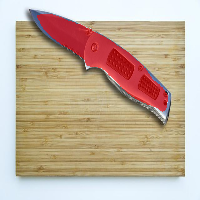}\hspace{1mm} &
\includegraphics[width=1.8cm,height=1.8cm]{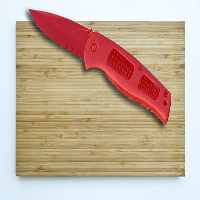}\hspace{1mm} &
\includegraphics[width=1.8cm,height=1.8cm]{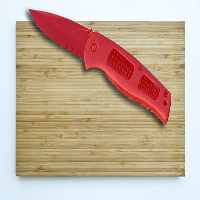}\hspace{1mm} &
\includegraphics[width=1.8cm,height=1.8cm]{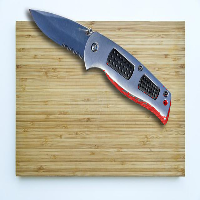}\hspace{1mm} &
\includegraphics[width=1.8cm,height=1.8cm]{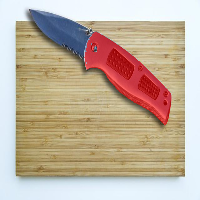}\hspace{1mm} &
\includegraphics[width=1.8cm,height=1.8cm]{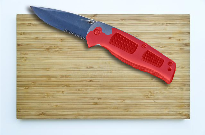}\hspace{1mm} \\
 
\vspace{-2mm}
\includegraphics[width=1.8cm,height=1.8cm]{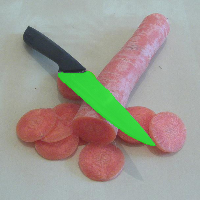}\hspace{1mm} &
\includegraphics[width=1.8cm,height=1.8cm]{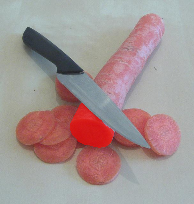}\hspace{1mm} &
\includegraphics[width=1.8cm,height=1.8cm]{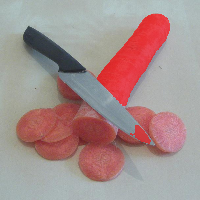}\hspace{1mm} &
\includegraphics[width=1.8cm,height=1.8cm]{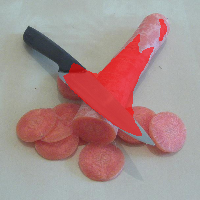}\hspace{1mm} &
\includegraphics[width=1.8cm,height=1.8cm]{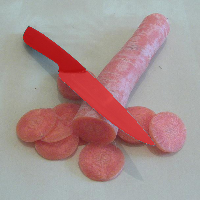}\hspace{1mm} &
\includegraphics[width=1.8cm,height=1.8cm]{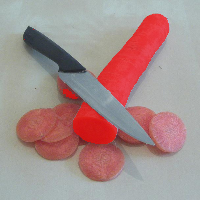}\hspace{1mm} &
\includegraphics[width=1.8cm,height=1.8cm]{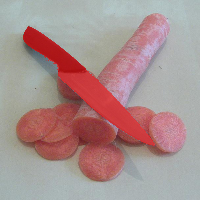}\hspace{1mm} &
\includegraphics[width=1.8cm,height=1.8cm]{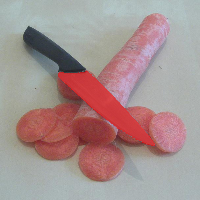}\hspace{1mm} &
\includegraphics[width=1.8cm,height=1.8cm]{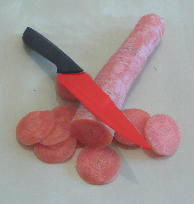}\hspace{1mm} \\
 
\vspace{-2mm}
\includegraphics[width=1.8cm,height=1.8cm]{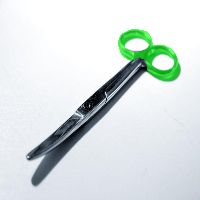}\hspace{1mm} &
\includegraphics[width=1.8cm,height=1.8cm]{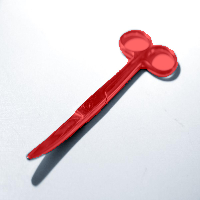}\hspace{1mm} &
\includegraphics[width=1.8cm,height=1.8cm]{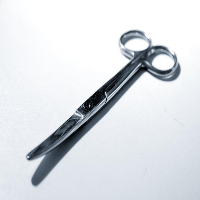}\hspace{1mm} &
\includegraphics[width=1.8cm,height=1.8cm]{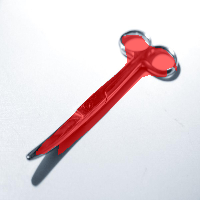}\hspace{1mm} &
\includegraphics[width=1.8cm,height=1.8cm]{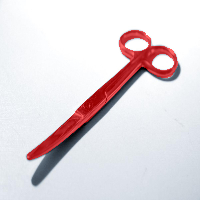}\hspace{1mm} &
\includegraphics[width=1.8cm,height=1.8cm]{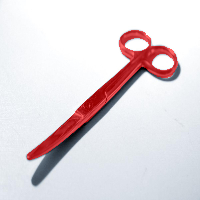}\hspace{1mm} &
\includegraphics[width=1.8cm,height=1.8cm]{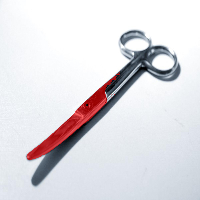}\hspace{1mm} &
\includegraphics[width=1.8cm,height=1.8cm]{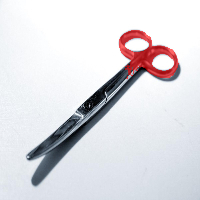}\hspace{1mm} &
\includegraphics[width=1.8cm,height=1.8cm]{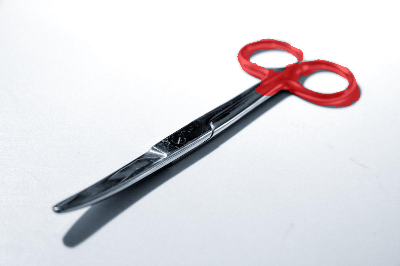}\hspace{1mm} \\
 \vspace{-2mm}

\includegraphics[width=1.8cm,height=1.8cm]{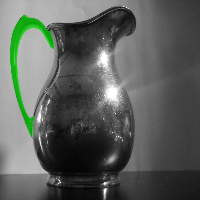}\hspace{1mm} &
\includegraphics[width=1.8cm,height=1.8cm]{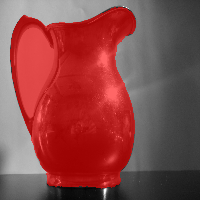}\hspace{1mm} &
\includegraphics[width=1.8cm,height=1.8cm]{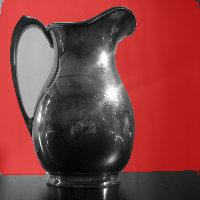}\hspace{1mm} &
\includegraphics[width=1.8cm,height=1.8cm]{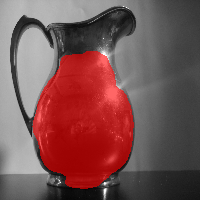}\hspace{1mm} &
\includegraphics[width=1.8cm,height=1.8cm]{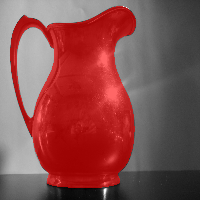}\hspace{1mm} &
\includegraphics[width=1.8cm,height=1.8cm]{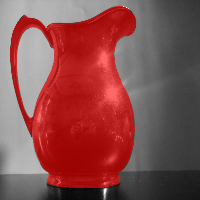}\hspace{1mm} &
\includegraphics[width=1.8cm,height=1.8cm]{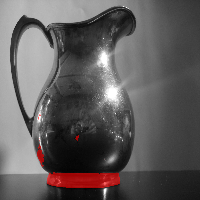}\hspace{1mm} &
\includegraphics[width=1.8cm,height=1.8cm]{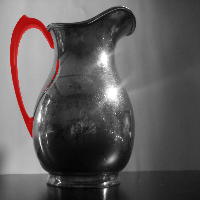}\hspace{1mm} &
\includegraphics[width=1.8cm,height=1.8cm]{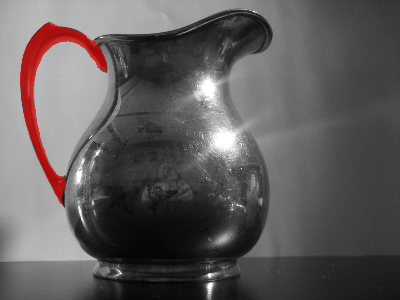}\hspace{1mm} \\
 \vspace{-2mm}

\includegraphics[width=1.8cm,height=1.8cm]{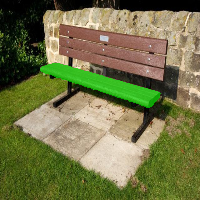}\hspace{1mm} &
\includegraphics[width=1.8cm,height=1.8cm]{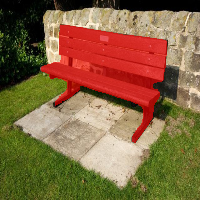}\hspace{1mm} &
\includegraphics[width=1.8cm,height=1.8cm]{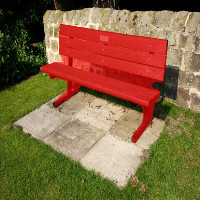}\hspace{1mm} &
\includegraphics[width=1.8cm,height=1.8cm]{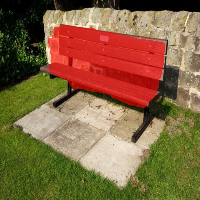}\hspace{1mm} &
\includegraphics[width=1.8cm,height=1.8cm]{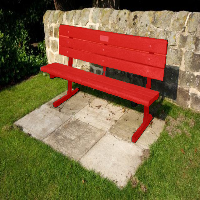}\hspace{1mm} &
\includegraphics[width=1.8cm,height=1.8cm]{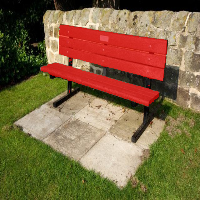}\hspace{1mm} &
\includegraphics[width=1.8cm,height=1.8cm]{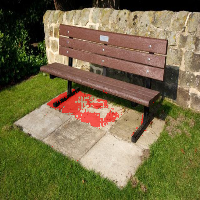}\hspace{1mm} &
\includegraphics[width=1.8cm,height=1.8cm]{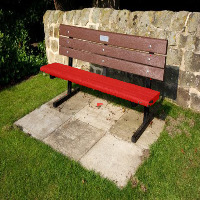}\hspace{1mm} &
\includegraphics[width=1.8cm,height=1.8cm]{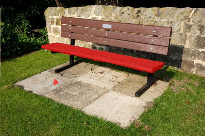}\hspace{1mm} \\
 \vspace{-2mm}

\includegraphics[width=1.8cm,height=1.8cm]{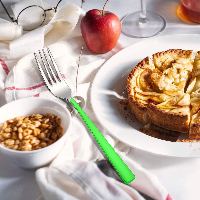}\hspace{1mm} &
\includegraphics[width=1.8cm,height=1.8cm]{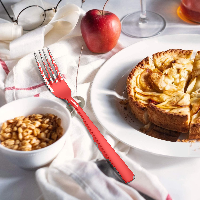}\hspace{1mm} &
\includegraphics[width=1.8cm,height=1.8cm]{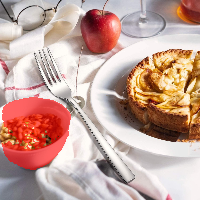}\hspace{1mm} &
\includegraphics[width=1.8cm,height=1.8cm]{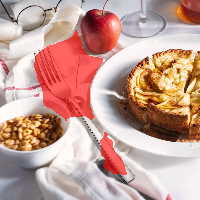}\hspace{1mm} &
\includegraphics[width=1.8cm,height=1.8cm]{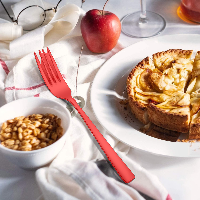}\hspace{1mm} &
\includegraphics[width=1.8cm,height=1.8cm]{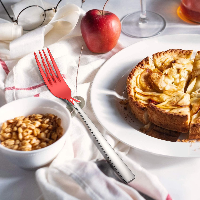}\hspace{1mm} &
\includegraphics[width=1.8cm,height=1.8cm]{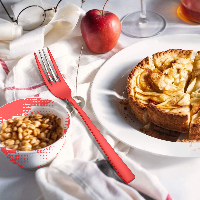}\hspace{1mm} &
\includegraphics[width=1.8cm,height=1.8cm]{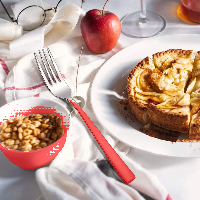}\hspace{1mm} &
\includegraphics[width=1.8cm,height=1.8cm]{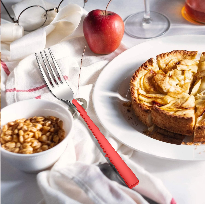}\hspace{1mm} \\
 \vspace{-2mm}

\includegraphics[width=1.8cm,height=1.8cm]{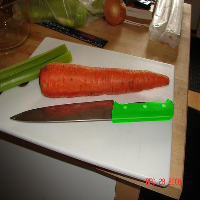}\hspace{1mm} &
\includegraphics[width=1.8cm,height=1.8cm]{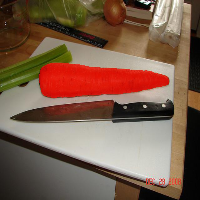}\hspace{1mm} &
\includegraphics[width=1.8cm,height=1.8cm]{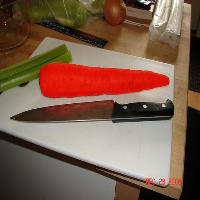}\hspace{1mm} &
\includegraphics[width=1.8cm,height=1.8cm]{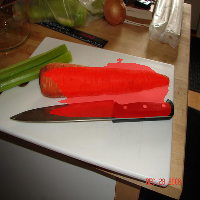}\hspace{1mm} &
\includegraphics[width=1.8cm,height=1.8cm]{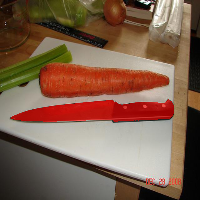}\hspace{1mm} &
\includegraphics[width=1.8cm,height=1.8cm]{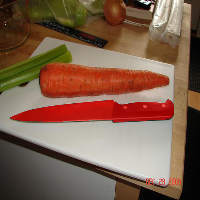}\hspace{1mm} &
\includegraphics[width=1.8cm,height=1.8cm]{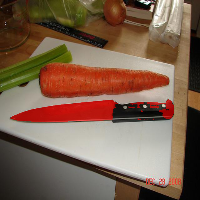}\hspace{1mm} &
\includegraphics[width=1.8cm,height=1.8cm]{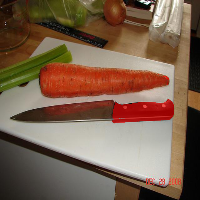}\hspace{1mm} &
\includegraphics[width=1.8cm,height=1.8cm]{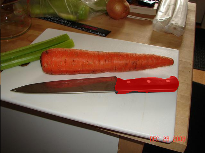}\hspace{1mm} \\
 
\vspace{-2mm}
\includegraphics[width=1.8cm,height=1.8cm]{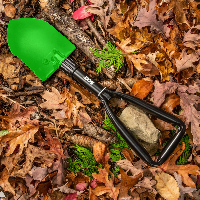}\hspace{1mm} &
\includegraphics[width=1.8cm,height=1.8cm]{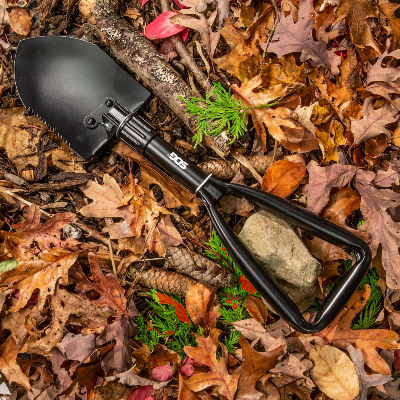}\hspace{1mm} &
\includegraphics[width=1.8cm,height=1.8cm]{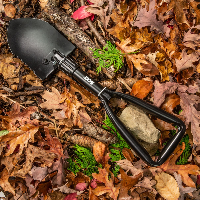}\hspace{1mm} &
\includegraphics[width=1.8cm,height=1.8cm]{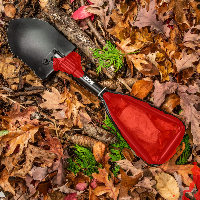}\hspace{1mm} &
\includegraphics[width=1.8cm,height=1.8cm]{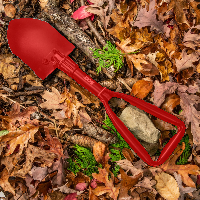}\hspace{1mm} &
\includegraphics[width=1.8cm,height=1.8cm]{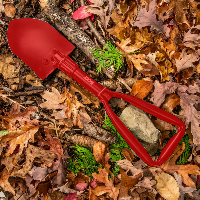}\hspace{1mm} &
\includegraphics[width=1.8cm,height=1.8cm]{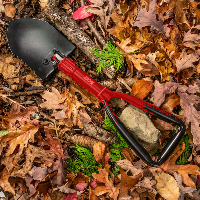}\hspace{1mm} &
\includegraphics[width=1.8cm,height=1.8cm]{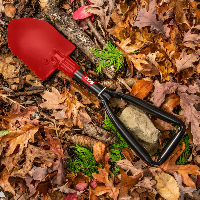}\hspace{1mm} &
\includegraphics[width=1.8cm,height=1.8cm]{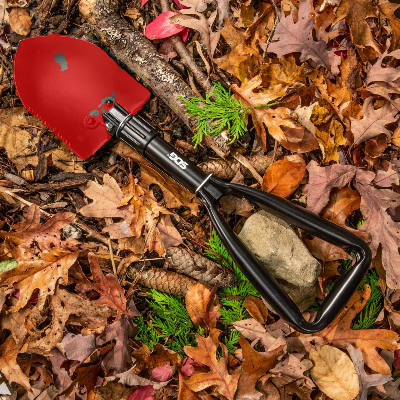}\hspace{1mm} \\
 
\vspace{-2mm}
\includegraphics[width=1.8cm,height=1.8cm]{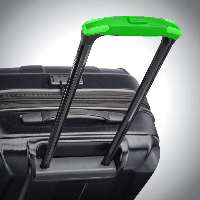}\hspace{1mm} &
\includegraphics[width=1.8cm,height=1.8cm]{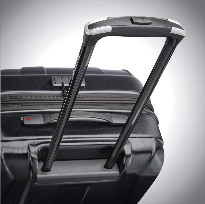}\hspace{1mm} &
\includegraphics[width=1.8cm,height=1.8cm]{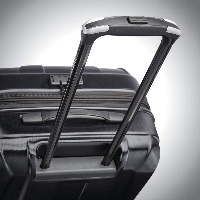}\hspace{1mm} &
\includegraphics[width=1.8cm,height=1.8cm]{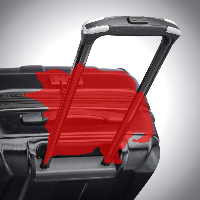}\hspace{1mm} &
\includegraphics[width=1.8cm,height=1.8cm]{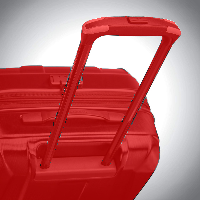}\hspace{1mm} &
\includegraphics[width=1.8cm,height=1.8cm]{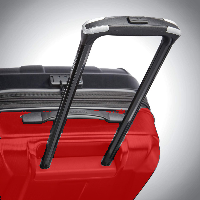}\hspace{1mm} &
\includegraphics[width=1.8cm,height=1.8cm]{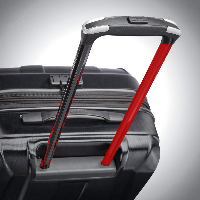}\hspace{1mm} &
\includegraphics[width=1.8cm,height=1.8cm]{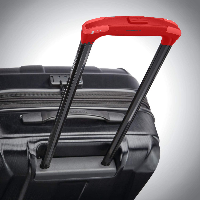}\hspace{1mm} &
\includegraphics[width=1.8cm,height=1.8cm]{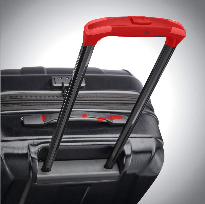}\hspace{1mm} \\
 \vspace{-2mm}

\includegraphics[width=1.8cm,height=1.8cm]{figs/selected_imgs_neurips/gt/wine_glass_001774-wine_glass-stem.png}\hspace{1mm} &
\includegraphics[width=1.8cm,height=1.8cm]{figs/selected_imgs_neurips/pred_mask_human_xdecoder/wine_glass_001774-wine_glass-stem.png}\hspace{1mm} &
\includegraphics[width=1.8cm,height=1.8cm]{figs/selected_imgs_neurips/pred_mask_human_seem/wine_glass_001774-wine_glass-stem.png}\hspace{1mm} &
\includegraphics[width=1.8cm,height=1.8cm]{figs/selected_imgs_neurips/pred_mask_human_tris/wine_glass_001774-wine_glass-stem.png}\hspace{1mm} &
\includegraphics[width=1.8cm,height=1.8cm]{figs/selected_imgs_neurips/pred_mask_human_groundedsam/wine_glass_001774-wine_glass-stem.png}\hspace{1mm} &
\includegraphics[width=1.8cm,height=1.8cm]{figs/selected_imgs_neurips/pred_mask_human_minigpt/wine_glass_001774-wine_glass-stem.png}\hspace{1mm} &
\includegraphics[width=1.8cm,height=1.8cm]{figs/selected_imgs_neurips/pred_lisa-untrain-test/wine_glass_001774-wine_glass-stem.png}\hspace{1mm} &
\includegraphics[width=1.8cm,height=1.8cm]{figs/selected_imgs_neurips/pred_lisa-train1800/wine_glass_001774-wine_glass-stem.png}\hspace{1mm} &
\includegraphics[width=1.8cm,height=1.8cm]{figs/selected_imgs_neurips/pred_pisa_pretrain-v1-dinodecoder-train1800/wine_glass_001774-wine_glass-stem.png}\hspace{1mm} \\
 \vspace{-2mm}

\includegraphics[width=1.8cm,height=1.8cm]{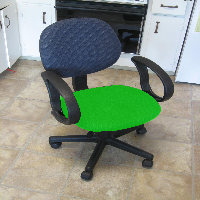}\hspace{1mm} &
\includegraphics[width=1.8cm,height=1.8cm]{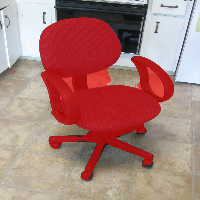}\hspace{1mm} &
\includegraphics[width=1.8cm,height=1.8cm]{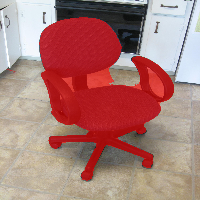}\hspace{1mm} &
\includegraphics[width=1.8cm,height=1.8cm]{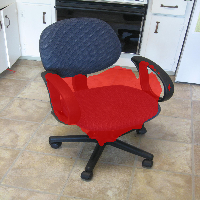}\hspace{1mm} &
\includegraphics[width=1.8cm,height=1.8cm]{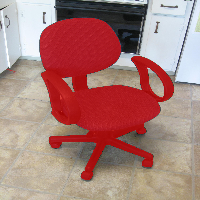}\hspace{1mm} &
\includegraphics[width=1.8cm,height=1.8cm]{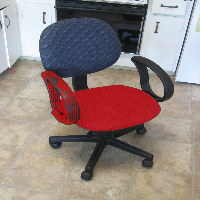}\hspace{1mm} &
\includegraphics[width=1.8cm,height=1.8cm]{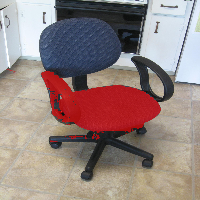}\hspace{1mm} &
\includegraphics[width=1.8cm,height=1.8cm]{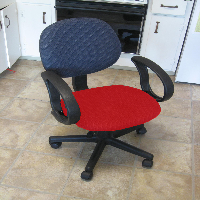}\hspace{1mm} &
\includegraphics[width=1.8cm,height=1.8cm]{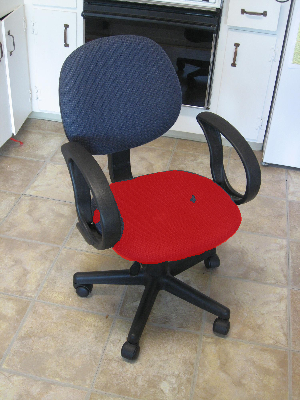}\hspace{1mm} &\\[1pt]
                %-----------------------------------------------------------------------------------------
			{\scriptsize Ground Truth} & {\scriptsize X-Decoder} & {\scriptsize SEEM} & {\scriptsize TRIS} & {\scriptsize G-SAM} & {\scriptsize MiniGPT-v2} & {\scriptsize LISA-Pretrain} & {\scriptsize \textcolor{red}{LISA-Finetune}} & {\scriptsize \textcolor{red}{PISA-Finetune}}\ \\
		\end{tabular}
	}
	\vspace{-3mm}
	\caption{Qualitative comparison of different VLMs and the fine-tuned models. In these examples, the pre-trained LISA falls short of recognizing the correct part. After fine-tuning, both LISA and PISA perform well on the part identification.}
	\label{fig:qualitative supplementary results 2}
	\vspace{-5mm}
\end{figure*}

\begin{figure*}[h]
	\centering
	\vspace{-5mm}
	\resizebox{1\textwidth}{!}
	{
		\begin{tabular}{@{}c@{}c@{}c@{}c@{}c@{}c@{}c@{}c@{}c@{}c@{}c@{}c}
  %%%%%%%%%%%%%lisa already good
			 
\includegraphics[width=1.8cm,height=1.8cm]{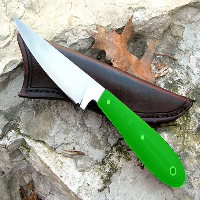}\hspace{1mm} &
\includegraphics[width=1.8cm,height=1.8cm]{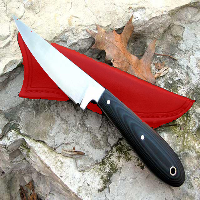}\hspace{1mm} &
\includegraphics[width=1.8cm,height=1.8cm]{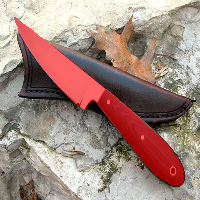}\hspace{1mm} &
\includegraphics[width=1.8cm,height=1.8cm]{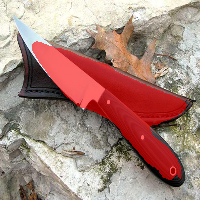}\hspace{1mm} &
\includegraphics[width=1.8cm,height=1.8cm]{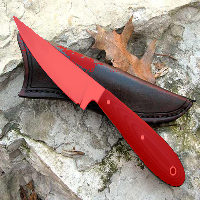}\hspace{1mm} &
\includegraphics[width=1.8cm,height=1.8cm]{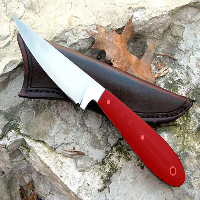}\hspace{1mm} &
\includegraphics[width=1.8cm,height=1.8cm]{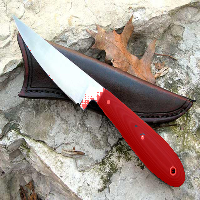}\hspace{1mm} &
\includegraphics[width=1.8cm,height=1.8cm]{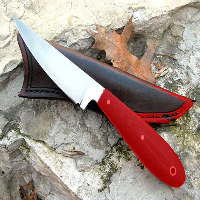}\hspace{1mm} &
\includegraphics[width=1.8cm,height=1.8cm]{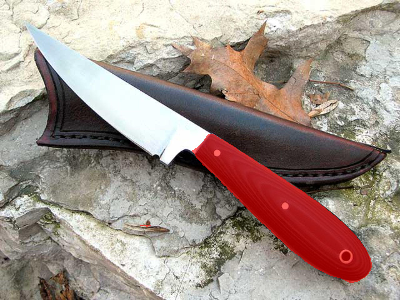}\hspace{1mm} \\
 \vspace{-2mm}

\includegraphics[width=1.8cm,height=1.8cm]{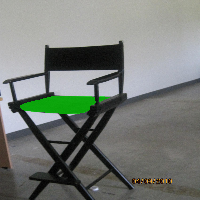}\hspace{1mm} &
\includegraphics[width=1.8cm,height=1.8cm]{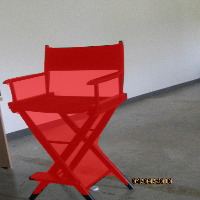}\hspace{1mm} &
\includegraphics[width=1.8cm,height=1.8cm]{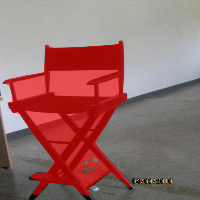}\hspace{1mm} &
\includegraphics[width=1.8cm,height=1.8cm]{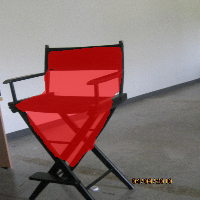}\hspace{1mm} &
\includegraphics[width=1.8cm,height=1.8cm]{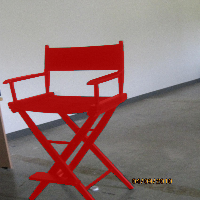}\hspace{1mm} &
\includegraphics[width=1.8cm,height=1.8cm]{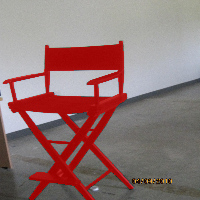}\hspace{1mm} &
\includegraphics[width=1.8cm,height=1.8cm]{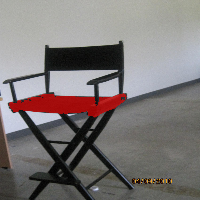}\hspace{1mm} &
\includegraphics[width=1.8cm,height=1.8cm]{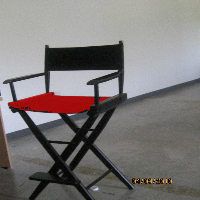}\hspace{1mm} &
\includegraphics[width=1.8cm,height=1.8cm]{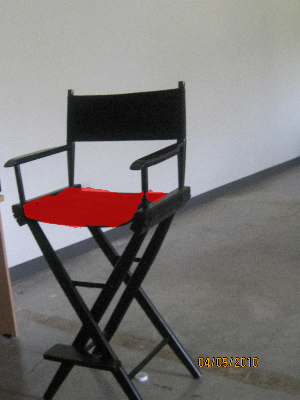}\hspace{1mm} \\
 \vspace{-2mm}

\includegraphics[width=1.8cm,height=1.8cm]{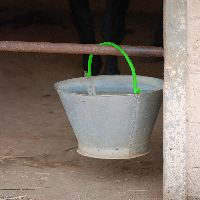}\hspace{1mm} &
\includegraphics[width=1.8cm,height=1.8cm]{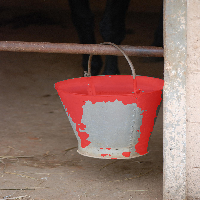}\hspace{1mm} &
\includegraphics[width=1.8cm,height=1.8cm]{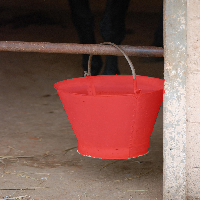}\hspace{1mm} &
\includegraphics[width=1.8cm,height=1.8cm]{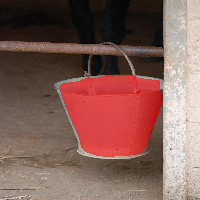}\hspace{1mm} &
\includegraphics[width=1.8cm,height=1.8cm]{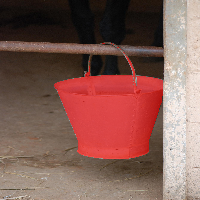}\hspace{1mm} &
\includegraphics[width=1.8cm,height=1.8cm]{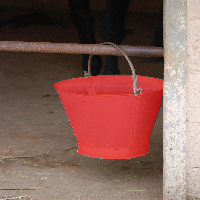}\hspace{1mm} &
\includegraphics[width=1.8cm,height=1.8cm]{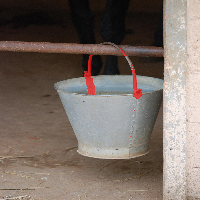}\hspace{1mm} &
\includegraphics[width=1.8cm,height=1.8cm]{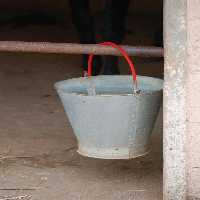}\hspace{1mm} &
\includegraphics[width=1.8cm,height=1.8cm]{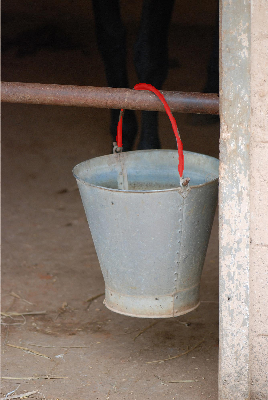}\hspace{1mm} \\
 
\vspace{-2mm}
\includegraphics[width=1.8cm,height=1.8cm]{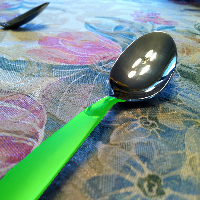}\hspace{1mm} &
\includegraphics[width=1.8cm,height=1.8cm]{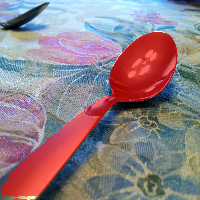}\hspace{1mm} &
\includegraphics[width=1.8cm,height=1.8cm]{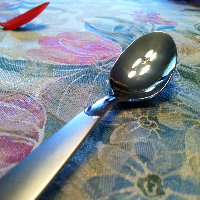}\hspace{1mm} &
\includegraphics[width=1.8cm,height=1.8cm]{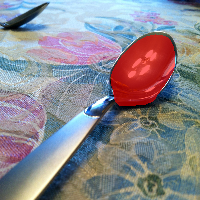}\hspace{1mm} &
\includegraphics[width=1.8cm,height=1.8cm]{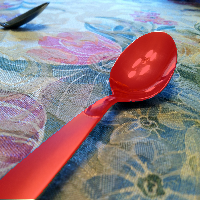}\hspace{1mm} &
\includegraphics[width=1.8cm,height=1.8cm]{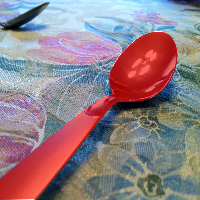}\hspace{1mm} &
\includegraphics[width=1.8cm,height=1.8cm]{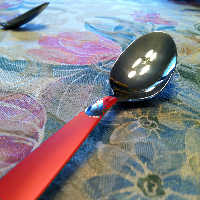}\hspace{1mm} &
\includegraphics[width=1.8cm,height=1.8cm]{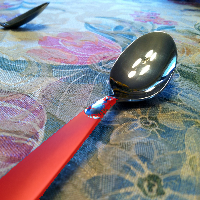}\hspace{1mm} &
\includegraphics[width=1.8cm,height=1.8cm]{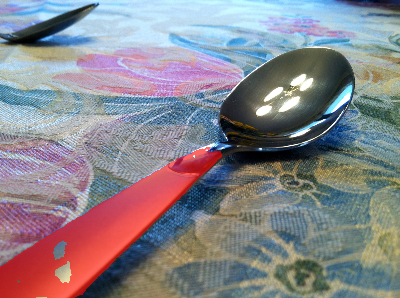}\hspace{1mm} \\
 
\vspace{-2mm}
\includegraphics[width=1.8cm,height=1.8cm]{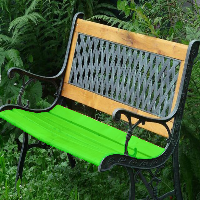}\hspace{1mm} &
\includegraphics[width=1.8cm,height=1.8cm]{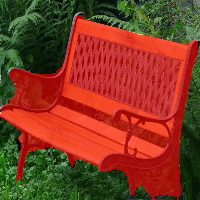}\hspace{1mm} &
\includegraphics[width=1.8cm,height=1.8cm]{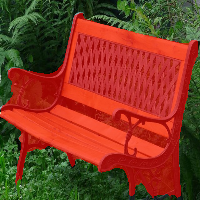}\hspace{1mm} &
\includegraphics[width=1.8cm,height=1.8cm]{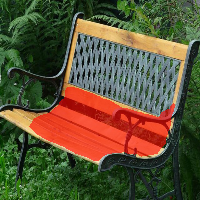}\hspace{1mm} &
\includegraphics[width=1.8cm,height=1.8cm]{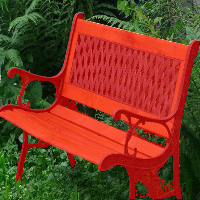}\hspace{1mm} &
\includegraphics[width=1.8cm,height=1.8cm]{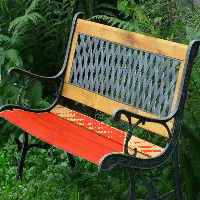}\hspace{1mm} &
\includegraphics[width=1.8cm,height=1.8cm]{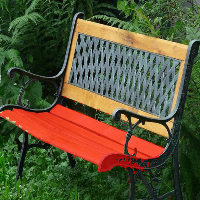}\hspace{1mm} &
\includegraphics[width=1.8cm,height=1.8cm]{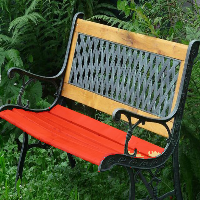}\hspace{1mm} &
\includegraphics[width=1.8cm,height=1.8cm]{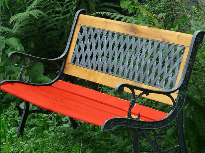}\hspace{1mm} \\
 \vspace{-2mm}

\includegraphics[width=1.8cm,height=1.8cm]{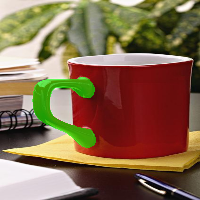}\hspace{1mm} &
\includegraphics[width=1.8cm,height=1.8cm]{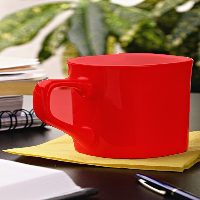}\hspace{1mm} &
\includegraphics[width=1.8cm,height=1.8cm]{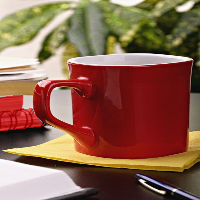}\hspace{1mm} &
\includegraphics[width=1.8cm,height=1.8cm]{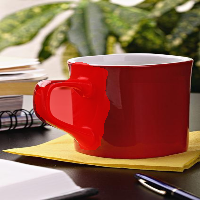}\hspace{1mm} &
\includegraphics[width=1.8cm,height=1.8cm]{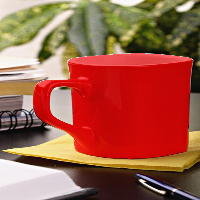}\hspace{1mm} &
\includegraphics[width=1.8cm,height=1.8cm]{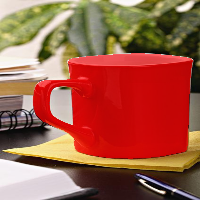}\hspace{1mm} &
\includegraphics[width=1.8cm,height=1.8cm]{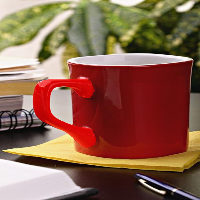}\hspace{1mm} &
\includegraphics[width=1.8cm,height=1.8cm]{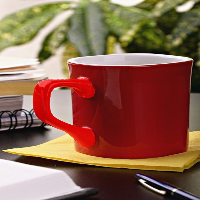}\hspace{1mm} &
\includegraphics[width=1.8cm,height=1.8cm]{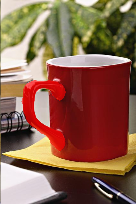}\hspace{1mm} \\
 \vspace{-2mm}

\includegraphics[width=1.8cm,height=1.8cm]{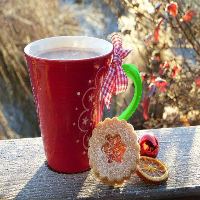}\hspace{1mm} &
\includegraphics[width=1.8cm,height=1.8cm]{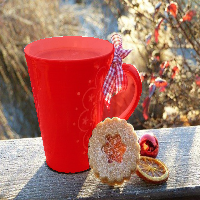}\hspace{1mm} &
\includegraphics[width=1.8cm,height=1.8cm]{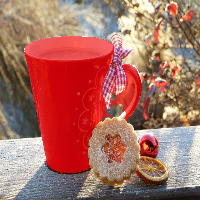}\hspace{1mm} &
\includegraphics[width=1.8cm,height=1.8cm]{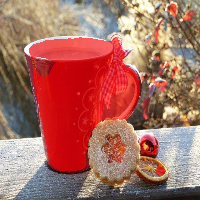}\hspace{1mm} &
\includegraphics[width=1.8cm,height=1.8cm]{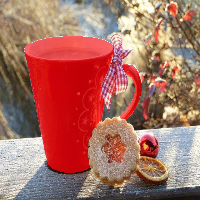}\hspace{1mm} &
\includegraphics[width=1.8cm,height=1.8cm]{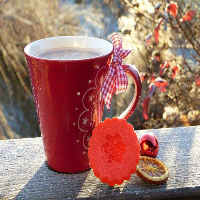}\hspace{1mm} &
\includegraphics[width=1.8cm,height=1.8cm]{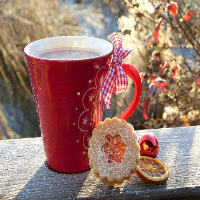}\hspace{1mm} &
\includegraphics[width=1.8cm,height=1.8cm]{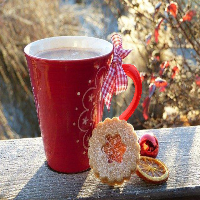}\hspace{1mm} &
\includegraphics[width=1.8cm,height=1.8cm]{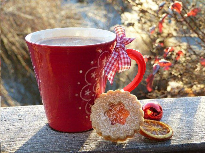}\hspace{1mm} \\
 \vspace{-2mm}

\includegraphics[width=1.8cm,height=1.8cm]{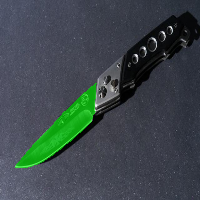}\hspace{1mm} &
\includegraphics[width=1.8cm,height=1.8cm]{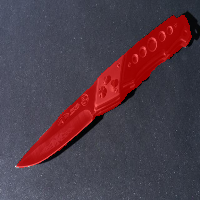}\hspace{1mm} &
\includegraphics[width=1.8cm,height=1.8cm]{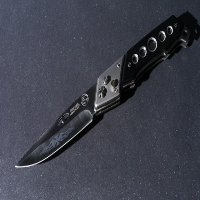}\hspace{1mm} &
\includegraphics[width=1.8cm,height=1.8cm]{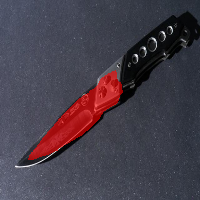}\hspace{1mm} &
\includegraphics[width=1.8cm,height=1.8cm]{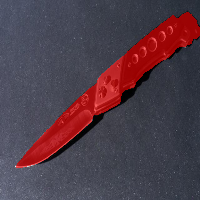}\hspace{1mm} &
\includegraphics[width=1.8cm,height=1.8cm]{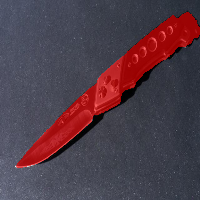}\hspace{1mm} &
\includegraphics[width=1.8cm,height=1.8cm]{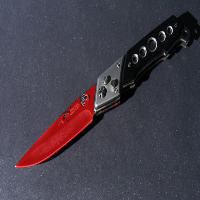}\hspace{1mm} &
\includegraphics[width=1.8cm,height=1.8cm]{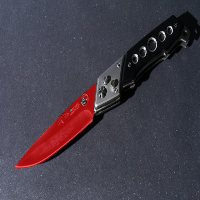}\hspace{1mm} &
\includegraphics[width=1.8cm,height=1.8cm]{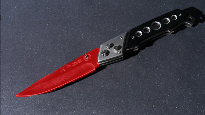}\hspace{1mm} \\
 \vspace{-2mm}

\includegraphics[width=1.8cm,height=1.8cm]{figs/selected_imgs_neurips/gt/scissors_001402-scissors-handle.png}\hspace{1mm} &
\includegraphics[width=1.8cm,height=1.8cm]{figs/selected_imgs_neurips/pred_mask_human_xdecoder/scissors_001402-scissors-handle.png}\hspace{1mm} &
\includegraphics[width=1.8cm,height=1.8cm]{figs/selected_imgs_neurips/pred_mask_human_seem/scissors_001402-scissors-handle.png}\hspace{1mm} &
\includegraphics[width=1.8cm,height=1.8cm]{figs/selected_imgs_neurips/pred_mask_human_tris/scissors_001402-scissors-handle.png}\hspace{1mm} &
\includegraphics[width=1.8cm,height=1.8cm]{figs/selected_imgs_neurips/pred_mask_human_groundedsam/scissors_001402-scissors-handle.png}\hspace{1mm} &
\includegraphics[width=1.8cm,height=1.8cm]{figs/selected_imgs_neurips/pred_mask_human_minigpt/scissors_001402-scissors-handle.png}\hspace{1mm} &
\includegraphics[width=1.8cm,height=1.8cm]{figs/selected_imgs_neurips/pred_lisa-untrain-test/scissors_001402-scissors-handle.png}\hspace{1mm} &
\includegraphics[width=1.8cm,height=1.8cm]{figs/selected_imgs_neurips/pred_lisa-train1800/scissors_001402-scissors-handle.png}\hspace{1mm} &
\includegraphics[width=1.8cm,height=1.8cm]{figs/selected_imgs_neurips/pred_pisa_pretrain-v1-dinodecoder-train1800/scissors_001402-scissors-handle.png}\hspace{1mm} \\
 \vspace{-2mm}

\includegraphics[width=1.8cm,height=1.8cm]{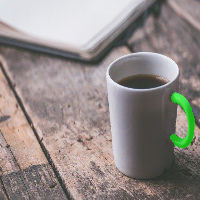}\hspace{1mm} &
\includegraphics[width=1.8cm,height=1.8cm]{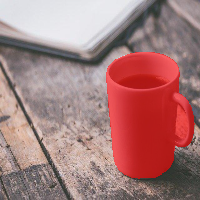}\hspace{1mm} &
\includegraphics[width=1.8cm,height=1.8cm]{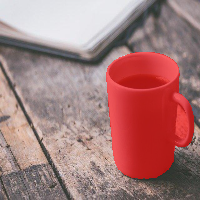}\hspace{1mm} &
\includegraphics[width=1.8cm,height=1.8cm]{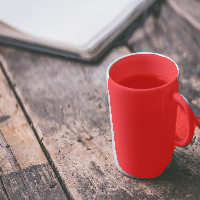}\hspace{1mm} &
\includegraphics[width=1.8cm,height=1.8cm]{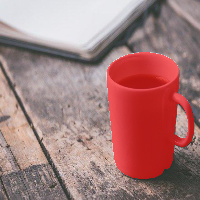}\hspace{1mm} &
\includegraphics[width=1.8cm,height=1.8cm]{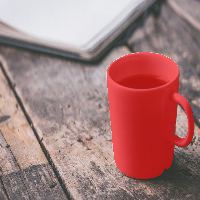}\hspace{1mm} &
\includegraphics[width=1.8cm,height=1.8cm]{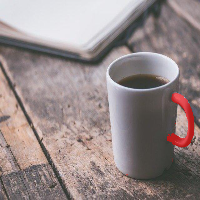}\hspace{1mm} &
\includegraphics[width=1.8cm,height=1.8cm]{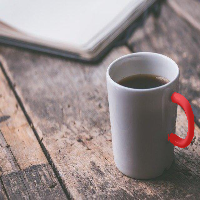}\hspace{1mm} &
\includegraphics[width=1.8cm,height=1.8cm]{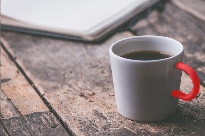}\hspace{1mm} \\
 \vspace{-2mm}

\includegraphics[width=1.8cm,height=1.8cm]{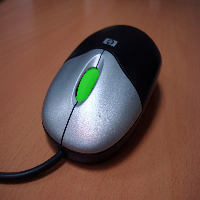}\hspace{1mm} &
\includegraphics[width=1.8cm,height=1.8cm]{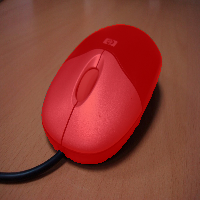}\hspace{1mm} &
\includegraphics[width=1.8cm,height=1.8cm]{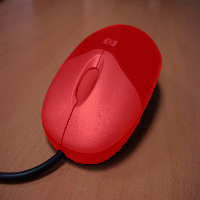}\hspace{1mm} &
\includegraphics[width=1.8cm,height=1.8cm]{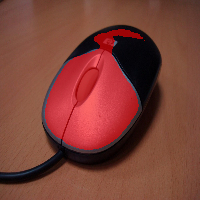}\hspace{1mm} &
\includegraphics[width=1.8cm,height=1.8cm]{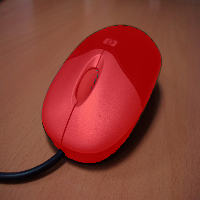}\hspace{1mm} &
\includegraphics[width=1.8cm,height=1.8cm]{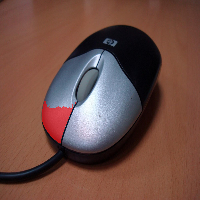}\hspace{1mm} &
\includegraphics[width=1.8cm,height=1.8cm]{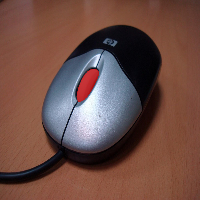}\hspace{1mm} &
\includegraphics[width=1.8cm,height=1.8cm]{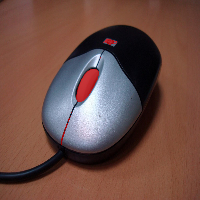}\hspace{1mm} &
\includegraphics[width=1.8cm,height=1.8cm]{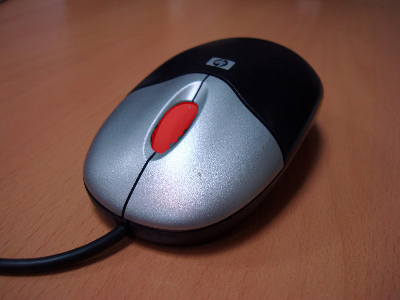}\hspace{1mm} \\
 \vspace{-2mm}

\includegraphics[width=1.8cm,height=1.8cm]{figs/selected_imgs_neurips/gt/6217625873_411169d784_o-laptop-keyboard.png}\hspace{1mm} &
\includegraphics[width=1.8cm,height=1.8cm]{figs/selected_imgs_neurips/pred_mask_human_xdecoder/6217625873_411169d784_o-laptop-keyboard.png}\hspace{1mm} &
\includegraphics[width=1.8cm,height=1.8cm]{figs/selected_imgs_neurips/pred_mask_human_seem/6217625873_411169d784_o-laptop-keyboard.png}\hspace{1mm} &
\includegraphics[width=1.8cm,height=1.8cm]{figs/selected_imgs_neurips/pred_mask_human_tris/6217625873_411169d784_o-laptop-keyboard.png}\hspace{1mm} &
\includegraphics[width=1.8cm,height=1.8cm]{figs/selected_imgs_neurips/pred_mask_human_groundedsam/6217625873_411169d784_o-laptop-keyboard.png}\hspace{1mm} &
\includegraphics[width=1.8cm,height=1.8cm]{figs/selected_imgs_neurips/pred_mask_human_minigpt/6217625873_411169d784_o-laptop-keyboard.png}\hspace{1mm} &
\includegraphics[width=1.8cm,height=1.8cm]{figs/selected_imgs_neurips/pred_lisa-untrain-test/6217625873_411169d784_o-laptop-keyboard.png}\hspace{1mm} &
\includegraphics[width=1.8cm,height=1.8cm]{figs/selected_imgs_neurips/pred_lisa-train1800/6217625873_411169d784_o-laptop-keyboard.png}\hspace{1mm} &
\includegraphics[width=1.8cm,height=1.8cm]{figs/selected_imgs_neurips/pred_pisa_pretrain-v1-dinodecoder-train1800/6217625873_411169d784_o-laptop-keyboard.png}\hspace{1mm} \\
 \vspace{-2mm}

\includegraphics[width=1.8cm,height=1.8cm]{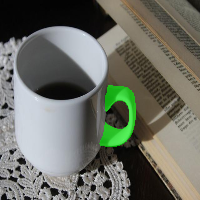}\hspace{1mm} &
\includegraphics[width=1.8cm,height=1.8cm]{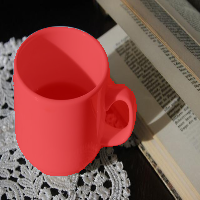}\hspace{1mm} &
\includegraphics[width=1.8cm,height=1.8cm]{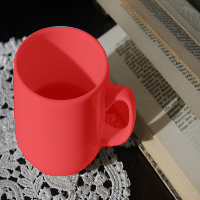}\hspace{1mm} &
\includegraphics[width=1.8cm,height=1.8cm]{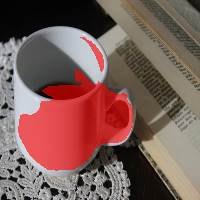}\hspace{1mm} &
\includegraphics[width=1.8cm,height=1.8cm]{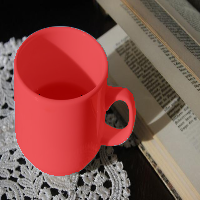}\hspace{1mm} &
\includegraphics[width=1.8cm,height=1.8cm]{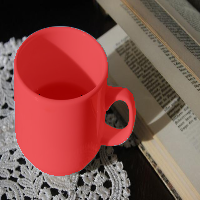}\hspace{1mm} &
\includegraphics[width=1.8cm,height=1.8cm]{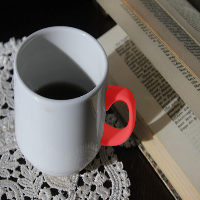}\hspace{1mm} &
\includegraphics[width=1.8cm,height=1.8cm]{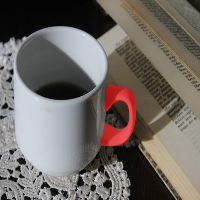}\hspace{1mm} &
\includegraphics[width=1.8cm,height=1.8cm]{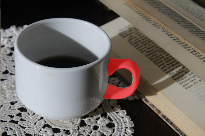}\hspace{1mm} \\
 \vspace{-2mm}

\includegraphics[width=1.8cm,height=1.8cm]{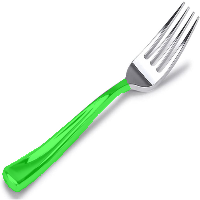}\hspace{1mm} &
\includegraphics[width=1.8cm,height=1.8cm]{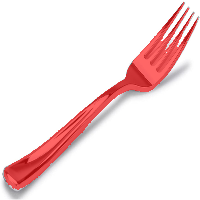}\hspace{1mm} &
\includegraphics[width=1.8cm,height=1.8cm]{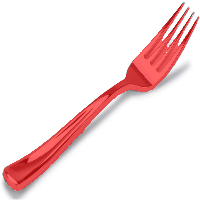}\hspace{1mm} &
\includegraphics[width=1.8cm,height=1.8cm]{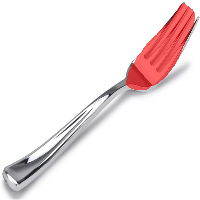}\hspace{1mm} &
\includegraphics[width=1.8cm,height=1.8cm]{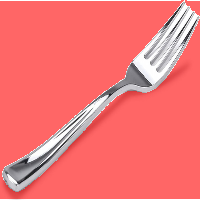}\hspace{1mm} &
\includegraphics[width=1.8cm,height=1.8cm]{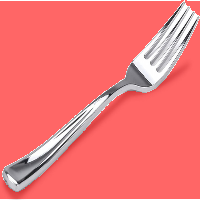}\hspace{1mm} &
\includegraphics[width=1.8cm,height=1.8cm]{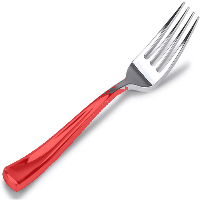}\hspace{1mm} &
\includegraphics[width=1.8cm,height=1.8cm]{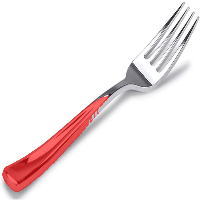}\hspace{1mm} &
\includegraphics[width=1.8cm,height=1.8cm]{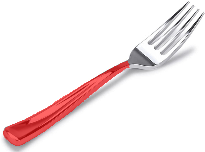}\hspace{1mm} &\\[1pt]
                %-----------------------------------------------------------------------------------------
			{\scriptsize Ground Truth} & {\scriptsize X-Decoder} & {\scriptsize SEEM} & {\scriptsize TRIS} & {\scriptsize G-SAM} & {\scriptsize MiniGPT-v2} & {\scriptsize \textcolor{red}{LISA-Pretrain}} & {\scriptsize LISA-Finetune} & {\scriptsize PISA-Finetune}\ \\
		\end{tabular}
	}
	\vspace{-3mm}
	\caption{Qualitative comparison of different VLMs and the fine-tuned models. In these examples, the pre-trained LISA already delivers good identification of the target parts.}
	\label{fig:qualitative supplementary results 1}
	\vspace{-5mm}
\end{figure*}

In Figure 3-5 of the main paper, we only include six qualitative results due to space limitations. In Fig.~\ref{fig:qualitative supplementary results 3}, we present more examples where the fine-tuned PISA shows superior visual part segmentation results, demonstrating the effectiveness of our proposed method. Besides, both the pre-trained and fine-tuned LISA models also demonstrate great potential in part grounding. Here, we visualize additional results of the VLMs and fine-tuned models. As shown in Fig.~\ref{fig:qualitative supplementary results 1}, the pre-trained LISA\cite{lai2023lisa} can better identify desired parts compared to other VLMs. This indicates the evaluation usage of our \Title{} dataset, where all the advanced VLMs can be evaluated and compared. Furthermore, in Fig.~\ref{fig:qualitative supplementary results 2}, the pre-trained LISA fails to recognize target parts, similar to other VLMs, while both fine-tuned models significantly improve the results. 

In Tab.~\ref{tab:qualitative image index}, we provide a list containing the name of each sample we evaluate so that their language input can be easily retrieved from our dataset.

\begin{table*}[t]
\centering
\begin{tabular}{|c|c|c|}
\hline
\textbf{Fig.~\ref{fig:qualitative supplementary results 3}} & \textbf{Fig.~\ref{fig:qualitative supplementary results 2}} & \textbf{Fig.~\ref{fig:qualitative supplementary results 1}} \\
\hline
\begin{minipage}[t]{5cm}
\begin{flushleft}
1. 1009786005\_d4a02fd811\_o-faucet-handle \\
2. 2329134125\_8a71be7470\_o-kettle-handle \\
3. 3088942376\_8681bb276f\_o-spoon-handle \\
4. cup\_000294-cup-handle \\
5. knife\_000911-knife-handle \\
6. 410044558\_6145ff0aaa\_o-pot-handle \\
7. laptop\_000445-laptop-keyboard \\
8. knife\_000691-knife-handle
\end{flushleft}
\end{minipage}
&
\begin{minipage}[t]{5cm}
\begin{flushleft}
1. 4178009615\_ed8921d0d1\_k-kettle-spout \\
2. cup\_000324-cup-handle \\
3. bottle\_002805-bottle-body \\
4. knife\_000568-knife-handle \\
5. knife\_000953-knife-blade \\
6. 34465720\_f8f20ee31a\_c-scissors-handle \\
7. 381204305\_e5e937fccc\_h-pitcher-handle \\
8. bench\_001273-bench-seat \\
9. fork\_002954-fork-handle \\
10. knife\_000154-knife-handle \\
11. shovel\_1-shovel-blade \\
12. suitcase\_001098-suitcase-telescopic\_handle \\
13. wine\_glass\_001774-wine\_glass-stem \\
14. dining\_4-chair-seat
\end{flushleft}
\end{minipage}
&
\begin{minipage}[t]{5cm}
\begin{flushleft}
1. 2491323916\_a05ac3648f\_o-knife-handle \\
2. 4580224808\_1194613deb\_o-chair-seat \\
3. 4471021242\_b9d855f193\_k-bucket-handle \\
4. 8607578325\_25221a7726\_h-spoon-handle \\
5. bench\_002898-bench-seat \\
6. cup\_001798-cup-handle \\
7. cup\_002055-cup-handle \\
8. knife\_000530-knife-blade \\
9. scissors\_001402-scissors-handle \\
10. cup\_002062-cup-handle \\
11. 2939090254\_2f01ebed6d\_o-computer\_mouse-scroll\_wheel \\
12. 6217625873\_411169d784\_o-laptop-keyboard \\
13. cup\_001104-cup-handle \\
14. fork\_001529-fork-handle 
\vspace{2mm}
\end{flushleft}
\end{minipage}
\\
\hline
\end{tabular}
\caption{Index name for samples in Fig.~\ref{fig:qualitative supplementary results 3}, Fig.~\ref{fig:qualitative supplementary results 2}, and Fig.~\ref{fig:qualitative supplementary results 1}.}
\label{tab:qualitative image index}
\end{table*}

%%%%%%%%%%%%%%%%%%%%%%%%%%%%%%%%%%%5
%% both tune good
%%%%%%%%%%%%%%%%%%%%%%%%%%%%%%%%%%%%

%%%%%%%%%%%%%%%%%%%%%%%%%%%%%%%
% page 3
%%%%%%%%%%%%%%%%%%%%%%%%%%%%%%%

\clearpage
\clearpage
\section{More Annotation Samples}
\label{appendix: more annotations}
\begin{figure*}[t]
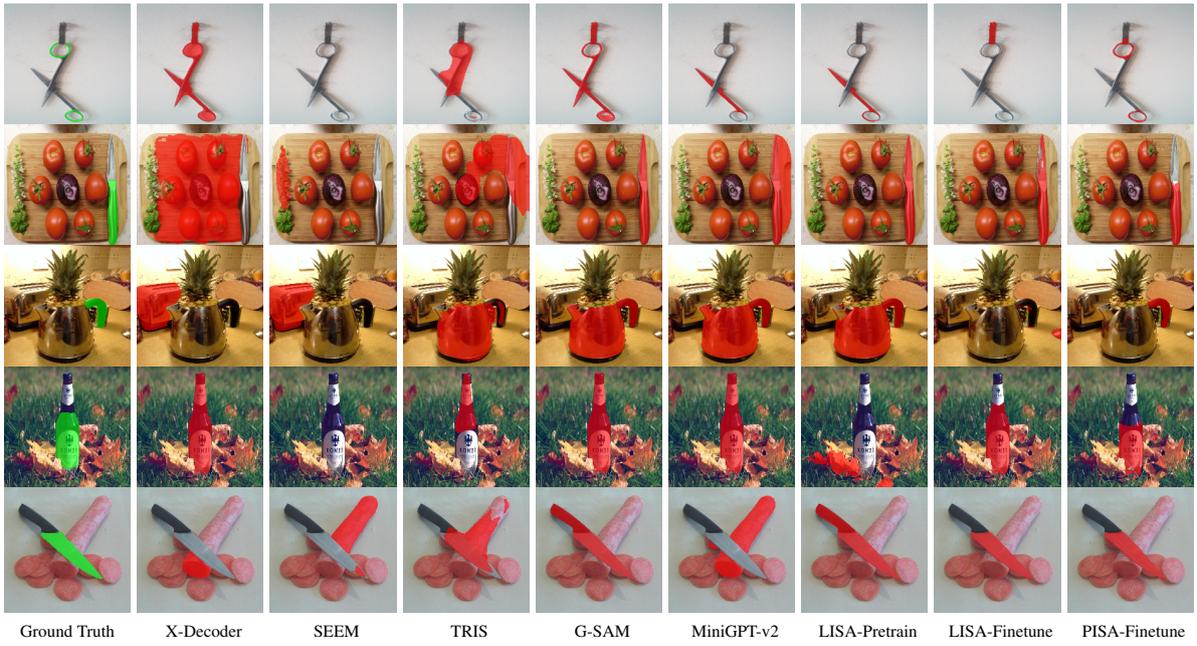

	\centering
	% \vspace{-4mm}
	\resizebox{1\textwidth}{!}
	{
		\begin{tabular}{@{}c@{}c@{}c@{}c@{}c@{}c@{}c@{}c@{}c@{}c@{}c@{}c}
			\vspace{-2mm}
			\includegraphics[width=1.8cm,height=1.8cm]{figs/selected_imgs_neurips/gt/538210619_c4def94c9b_o-scissors-handle.png}\hspace{1mm} &
			\includegraphics[width=1.8cm,height=1.8cm]{figs/selected_imgs_neurips/pred_mask_human_xdecoder/538210619_c4def94c9b_o-scissors-handle.png}\hspace{1mm} &
			\includegraphics[width=1.8cm,height=1.8cm]{figs/selected_imgs_neurips/pred_mask_human_seem/538210619_c4def94c9b_o-scissors-handle}\hspace{1mm} &
			\includegraphics[width=1.8cm,height=1.8cm]{figs/selected_imgs_neurips/pred_mask_human_tris/538210619_c4def94c9b_o-scissors-handle}\hspace{1mm} &
			\includegraphics[width=1.8cm,height=1.8cm]{figs/selected_imgs_neurips/pred_mask_human_groundedsam/538210619_c4def94c9b_o-scissors-handle}\hspace{1mm} &
			\includegraphics[width=1.8cm,height=1.8cm]{figs/selected_imgs_neurips/pred_mask_human_minigpt/538210619_c4def94c9b_o-scissors-handle}\hspace{1mm} &
			\includegraphics[width=1.8cm,height=1.8cm]{figs/selected_imgs_neurips/pred_lisa-untrain-test/538210619_c4def94c9b_o-scissors-handle}\hspace{1mm} &
                \includegraphics[width=1.8cm,height=1.8cm]{figs/selected_imgs_neurips/pred_lisa-train1800/538210619_c4def94c9b_o-scissors-handle}\hspace{1mm} &
			\includegraphics[width=1.8cm,height=1.8cm]{figs/selected_imgs_neurips/pred_pisa_pretrain-v1-dinodecoder-train1800/538210619_c4def94c9b_o-scissors-handle}\hspace{1mm} \\
			\vspace{-2mm}
			%-----------------------------------------------------------------------------------------
			\includegraphics[width=1.8cm,height=1.8cm]{figs/selected_imgs_neurips/gt/knife_002845-knife-handle.png}\hspace{1mm} &
                \includegraphics[width=1.8cm,height=1.8cm]{figs/selected_imgs_neurips/pred_mask_human_xdecoder/knife_002845-knife-handle.png}\hspace{1mm} &
			\includegraphics[width=1.8cm,height=1.8cm]{figs/selected_imgs_neurips/pred_mask_human_seem/knife_002845-knife-handle.png}\hspace{1mm} &
			\includegraphics[width=1.8cm,height=1.8cm]{figs/selected_imgs_neurips/pred_mask_human_tris/knife_002845-knife-handle.png}\hspace{1mm} &
			\includegraphics[width=1.8cm,height=1.8cm]{figs/selected_imgs_neurips/pred_mask_human_groundedsam/knife_002845-knife-handle.png}\hspace{1mm} &
			\includegraphics[width=1.8cm,height=1.8cm]{figs/selected_imgs_neurips/pred_mask_human_minigpt/knife_002845-knife-handle.png}\hspace{1mm} &
			\includegraphics[width=1.8cm,height=1.8cm]{figs/selected_imgs_neurips/pred_lisa-untrain-test/knife_002845-knife-handle.png}\hspace{1mm} &
   			\includegraphics[width=1.8cm,height=1.8cm]{figs/selected_imgs_neurips/pred_lisa-train1800/knife_002845-knife-handle.png}\hspace{1mm} &
			\includegraphics[width=1.8cm,height=1.8cm]{figs/selected_imgs_neurips/pred_pisa_pretrain-v1-dinodecoder-train1800/knife_002845-knife-handle.png}\hspace{1mm} \\
                \vspace{-2mm}
                %-----------------------------------------------------------------------------------------
			\includegraphics[width=1.8cm,height=1.8cm]{figs/selected_imgs_neurips/gt/2329134125_8a71be7470_o-kettle-handle.png}\hspace{1mm} &
                \includegraphics[width=1.8cm,height=1.8cm]{figs/selected_imgs_neurips/pred_mask_human_xdecoder/2329134125_8a71be7470_o-kettle-handle.png}\hspace{1mm} &
			\includegraphics[width=1.8cm,height=1.8cm]{figs/selected_imgs_neurips/pred_mask_human_seem/2329134125_8a71be7470_o-kettle-handle.png}\hspace{1mm} &
			\includegraphics[width=1.8cm,height=1.8cm]{figs/selected_imgs_neurips/pred_mask_human_tris/2329134125_8a71be7470_o-kettle-handle.png}\hspace{1mm} &
			\includegraphics[width=1.8cm,height=1.8cm]{figs/selected_imgs_neurips/pred_mask_human_groundedsam/2329134125_8a71be7470_o-kettle-handle.png}\hspace{1mm} &
			\includegraphics[width=1.8cm,height=1.8cm]{figs/selected_imgs_neurips/pred_mask_human_minigpt/2329134125_8a71be7470_o-kettle-handle.png}\hspace{1mm} &
			\includegraphics[width=1.8cm,height=1.8cm]{figs/selected_imgs_neurips/pred_lisa-untrain-test/2329134125_8a71be7470_o-kettle-handle.png}\hspace{1mm} &
   			\includegraphics[width=1.8cm,height=1.8cm]{figs/selected_imgs_neurips/pred_lisa-train1800/2329134125_8a71be7470_o-kettle-handle.png}\hspace{1mm} &
			\includegraphics[width=1.8cm,height=1.8cm]{figs/selected_imgs_neurips/pred_pisa_pretrain-v1-dinodecoder-train1800/2329134125_8a71be7470_o-kettle-handle.png}\hspace{1mm} \\
			\vspace{-2mm}
			%-----------------------------------------------------------------------------------------
			\includegraphics[width=1.8cm,height=1.8cm]{figs/selected_imgs_neurips/gt/bottle_002805-bottle-body.png}\hspace{1mm} &
                \includegraphics[width=1.8cm,height=1.8cm]{figs/selected_imgs_neurips/pred_mask_human_xdecoder/bottle_002805-bottle-body.png}\hspace{1mm} &
			\includegraphics[width=1.8cm,height=1.8cm]{figs/selected_imgs_neurips/pred_mask_human_seem/bottle_002805-bottle-body.png}\hspace{1mm} &
			\includegraphics[width=1.8cm,height=1.8cm]{figs/selected_imgs_neurips/pred_mask_human_tris/bottle_002805-bottle-body.png}\hspace{1mm} &
			\includegraphics[width=1.8cm,height=1.8cm]{figs/selected_imgs_neurips/pred_mask_human_groundedsam/bottle_002805-bottle-body.png}\hspace{1mm} &
			\includegraphics[width=1.8cm,height=1.8cm]{figs/selected_imgs_neurips/pred_mask_human_minigpt/bottle_002805-bottle-body.png}\hspace{1mm} &
			\includegraphics[width=1.8cm,height=1.8cm]{figs/selected_imgs_neurips/pred_lisa-untrain-test/bottle_002805-bottle-body.png}\hspace{1mm} &
   			\includegraphics[width=1.8cm,height=1.8cm]{figs/selected_imgs_neurips/pred_lisa-train1800/bottle_002805-bottle-body.png}\hspace{1mm} &
			\includegraphics[width=1.8cm,height=1.8cm]{figs/selected_imgs_neurips/pred_pisa_pretrain-v1-dinodecoder-train1800/bottle_002805-bottle-body.png}\hspace{1mm} \\
                \vspace{-2mm}
                %-----------------------------------------------------------------------------------------
			\includegraphics[width=1.8cm,height=1.8cm]{figs/selected_imgs_neurips/gt/knife_000953-knife-blade.png}\hspace{1mm} &
                \includegraphics[width=1.8cm,height=1.8cm]{figs/selected_imgs_neurips/pred_mask_human_xdecoder/knife_000953-knife-blade.png}\hspace{1mm} &
			\includegraphics[width=1.8cm,height=1.8cm]{figs/selected_imgs_neurips/pred_mask_human_seem/knife_000953-knife-blade.png}\hspace{1mm} &
			\includegraphics[width=1.8cm,height=1.8cm]{figs/selected_imgs_neurips/pred_mask_human_tris/knife_000953-knife-blade.png}\hspace{1mm} &
			\includegraphics[width=1.8cm,height=1.8cm]{figs/selected_imgs_neurips/pred_mask_human_groundedsam/knife_000953-knife-blade.png}\hspace{1mm} &
			\includegraphics[width=1.8cm,height=1.8cm]{figs/selected_imgs_neurips/pred_mask_human_minigpt/knife_000953-knife-blade.png}\hspace{1mm} &
			\includegraphics[width=1.8cm,height=1.8cm]{figs/selected_imgs_neurips/pred_lisa-untrain-test/knife_000953-knife-blade.png}\hspace{1mm} &
   			\includegraphics[width=1.8cm,height=1.8cm]{figs/selected_imgs_neurips/pred_lisa-train1800/knife_000953-knife-blade.png}\hspace{1mm} &
			\includegraphics[width=1.8cm,height=1.8cm]{figs/selected_imgs_neurips/pred_pisa_pretrain-v1-dinodecoder-train1800/knife_000953-knife-blade.png}\hspace{1mm} &\\[2pt]
                %-----------------------------------------------------------------------------------------
			{\scriptsize Ground Truth} & {\scriptsize X-Decoder} & {\scriptsize SEEM} & {\scriptsize TRIS} & {\scriptsize G-SAM} & {\scriptsize MiniGPT-v2} & {\scriptsize LISA-Pretrain} & {\scriptsize LISA-Finetune} & {\scriptsize PISA-Finetune}\ \\
		\end{tabular}
	}
	\vspace{-3mm}
\caption{More qualitative examples with corresponding annotations recorded in Table~\ref{tab:annotations for more examples}.}
	\label{fig:image example for annotation}
	\vspace{-5mm}
\end{figure*}

In addition to the annotation examples shown in Fig.~\ref{fig:annotation example}, we include five more annotations for the samples in Fig.~\ref{fig:image example for annotation} in Table~\ref{tab:annotations for more examples}. The listed annotations correspond to the order of the images.

\onecolumn

\begin{longtable}{|p{15cm}|}
\hline
\begin{lstlisting}
{
    "image_path": "538210619_c4def94c9b_o.jpg",
    "part_list": [
        {
            "object": "scissors",
            "part": "handle",
            "affordance": "hold",
            "action": "hold",
            "instruction": [
                "If I want to use the scissors, which part in the picture should I put my fingers in?",
                "Describe the part of the scissors in the picture where fingers should be placed.",
                "Where is the handle of the scissors in this image?",
                "Where is the handle of the scissors that can be held in this image?",
                "handle of the scissors",
                "handle of the scissors that can be held"
            ]
        }
    ]
}
\end{lstlisting}
\\ \hline
\begin{lstlisting}
{
    "image_path": "knife_002845.jpg",
    "part_list": [
        {
            "object": "knife",
            "part": "handle",
            "affordance": "hold",
            "action": "pick up",
            "instruction": [
                "If I want to pick up the knife, which part in the picture can be used?",
                "Which part of the knife is safe to hold when picking it up?",
                "Where is the handle of the knife in this image?",
                "Where is the handle of the knife that can be held in this image?",
                "handle of the knife",
                "handle of the knife that can be held"
            ]
        }
    ]
}
\end{lstlisting}
\\ \hline
\begin{lstlisting}
{
    "image_path": "2329134125_8a71be7470_o.jpg",
    "part_list": [
        {
            "object": "kettle",
            "part": "handle",
            "affordance": "hold",
            "action": "hold",
            "instruction": [
                "Which part in the picture can be utilized to hold the kettle?",
                "In the image, identify the part of the kettle that's meant to be held.",
                "Where is the handle of the kettle in this image?",
                "Where is the handle of the kettle that can be held in this image?",
                "handle of the kettle",
                "handle of the kettle that can be held"
            ]
        }
    ]
}
\end{lstlisting}
\\ \hline
\begin{lstlisting}
{
    "image_path": "bottle_002805.jpg",
    "part_list": [
        {
            "object": "bottle",
            "part": "body",
            "affordance": "hold",
            "action": "hold",
            "instruction": [
                "If I want to hold the bottles, which parts in the picture can be utilized?",
                "To hold the bottles, which parts are designed for grip?",
                "Where is the body of the bottle in this image?",
                "Where is the body of the bottle that can be held in this image?",
                "body of the bottle",
                "body of the bottle that can be held"
            ]
        }
    ]
}
\end{lstlisting}
\\ \hline
\begin{lstlisting}
{
    "image_path": "knife_000953.jpg",
    "part_list": [
        {
            "object": "knife",
            "part": "blade",
            "affordance": "cut",
            "action": "cut",
            "instruction": [
                "If I want to use the knife to cut the carrots, which part in the picture should be used?",
                "Identify the part of the knife ideal for slicing the carrots.",
                "Where is the blade of the knife in this image?",
                "Where is the blade of the knife that can cut in this image?",
                "blade of the knife",
                "blade of the knife that can cut"
            ]
        }
    ]
}
\end{lstlisting}
\\ \hline
\caption{Corresponding annotations for images in Fig.~\ref{fig:image example for annotation}.}
\label{tab:annotations for more examples}
\end{longtable}

\twocolumn

\section{A Case Study on Real-world Grasping Data.}
\label{appendix: case study}
Grasping is one vital aspect that our \Title{} benchmark aims to facilitate. Consequently, we evaluate the model trained with our data in a real-world tabletop grasping environment. We use the table setup from ShapeGrasp~\cite{Li2024ShapeGraspZT}, which consists of 38 objects covering 12 general categories and 49 tasks. These categories and tasks are the same as those in LERF-TOGO~\cite{rashid2023language}. More details about the dataset are included in the supplementary material.
Our trained PISA model is evaluated on the zero-shot task-oriented grasping task, as described in~\cite{Li2024ShapeGraspZT, rashid2023language}. We compare the successful part selection rate, defining a successful part selection as our output segmentation mask accurately aligned with the target part.
As shown in Tab.~\ref{tab:grasp}, PISA's zero-shot part identification ability is comparable to state-of-the-art (SOTA) methods. Additionally, due to PISA's end-to-end advantage, its execution time significantly outperforms others.

\begin{table}[h]
  \centering
  \footnotesize
  \centering
     \setlength{\tabcolsep}{1.5mm}{
     \resizebox{\linewidth}{!}{
     \begin{tabular}{c|ccc}
          % \specialrule{.1em}{.05em}{.05em}
          \toprule
          & PISA & ShapeGrasp & LERF-TOGO \\
            \midrule\midrule
            Part Selection (\%) & 80 & 86 & 82$^*$ \\
            Time (s) & 2 & 25 & 120\\
            \bottomrule
    \end{tabular}}
}
  \vspace{-4pt}
  \caption{Comparison of part selection accuracy and execution time. $^*$ indicates that LERF-TOGO uses the same categories of objects, but not identical ones.}
  \label{tab:grasp}
\end{table}

\begin{figure*}[h]
    \centering
    \begin{minipage}{0.23\textwidth}
        \centering
        \includegraphics[width=3cm, height=3cm]{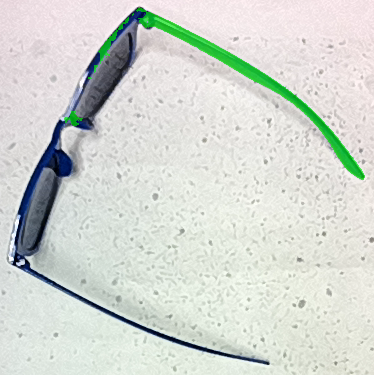}
        \parbox{0.9\textwidth}{\centering \small blue sunglasses - earhooks}
    \end{minipage}\hfill
    \begin{minipage}{0.23\textwidth}
        \centering
        \includegraphics[width=3cm, height=3cm]{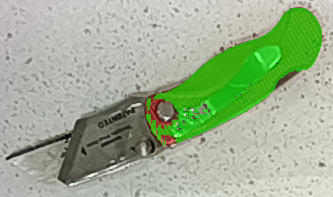}
        \parbox{0.9\textwidth}{\centering \small box cutter - handle}
    \end{minipage}\hfill
    \begin{minipage}{0.23\textwidth}
        \centering
        \includegraphics[width=3cm, height=3cm]{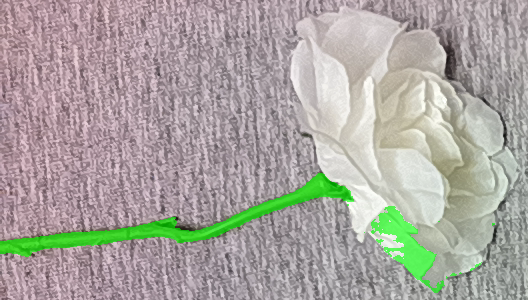}
        \parbox{0.9\textwidth}{\centering \small daisy - plant stem}
    \end{minipage}\hfill
    \begin{minipage}{0.23\textwidth}
        \centering
        \includegraphics[width=3cm, height=3cm]{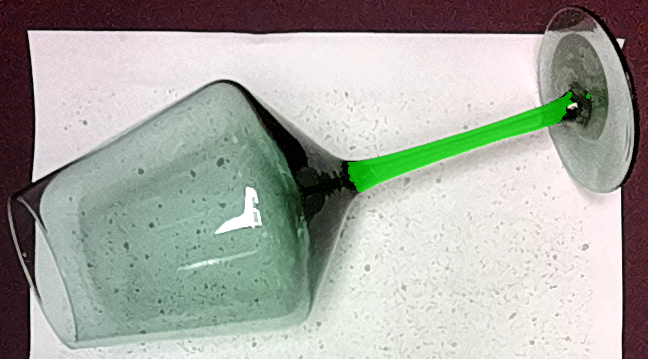}
        \parbox{0.9\textwidth}{\centering \small green martini glass - stem}
    \end{minipage}

    \vspace{0.3cm}

    \begin{minipage}{0.23\textwidth}
        \centering
        \includegraphics[width=3cm, height=3cm]{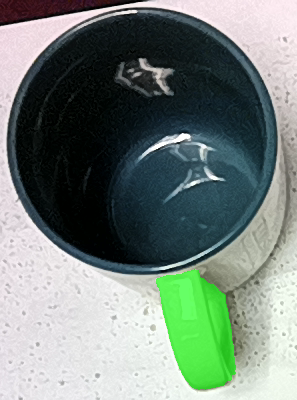}
        \parbox{0.9\textwidth}{\centering \small grey mug - handle}
    \end{minipage}\hfill
    \begin{minipage}{0.23\textwidth}
        \centering
        \includegraphics[width=3cm, height=3cm]{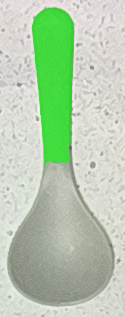}
        \parbox{0.9\textwidth}{\centering \small grey spoon - handle}
    \end{minipage}\hfill
    \begin{minipage}{0.23\textwidth}
        \centering
        \includegraphics[width=3cm, height=3cm]{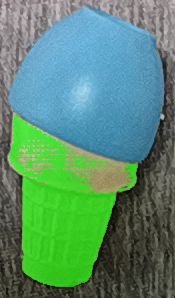}
        \parbox{0.9\textwidth}{\centering \small ice cream - cone}
    \end{minipage}\hfill
    \begin{minipage}{0.23\textwidth}
        \centering
        \includegraphics[width=3cm, height=3cm]{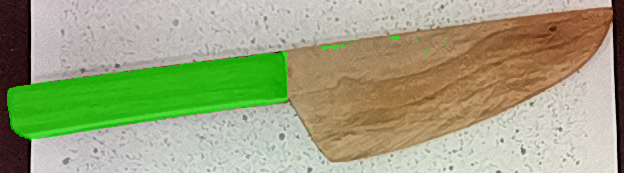}
        \parbox{0.9\textwidth}{\centering \small knife - handle}
    \end{minipage}

    \vspace{0.3cm}

    \begin{minipage}{0.23\textwidth}
        \centering
        \includegraphics[width=3cm, height=3cm]{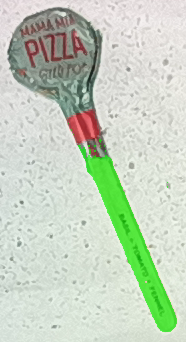}
        \parbox{0.9\textwidth}{\centering \small lollipop - stick}
    \end{minipage}\hfill
    \begin{minipage}{0.23\textwidth}
        \centering
        \includegraphics[width=3cm, height=3cm]{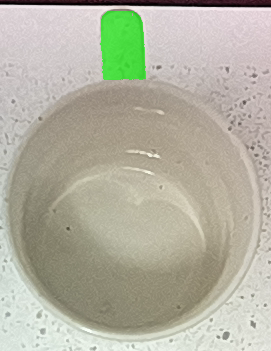}
        \parbox{0.9\textwidth}{\centering \small mug - handle}
    \end{minipage}\hfill
    \begin{minipage}{0.23\textwidth}
        \centering
        \includegraphics[width=3cm, height=3cm]{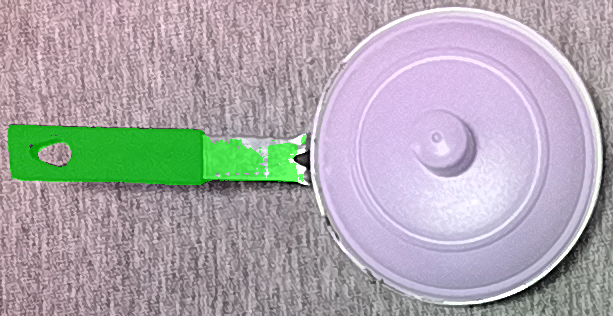}
        \parbox{0.9\textwidth}{\centering \small saucepan - handle}
    \end{minipage}\hfill
    \begin{minipage}{0.23\textwidth}
        \centering
        \includegraphics[width=3cm, height=3cm]{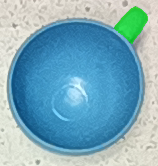}
        \parbox{0.9\textwidth}{\centering \small teacup - handle}
    \end{minipage}

    \caption{PISA zero-shot prediction on novel objects. Green masks represent the prediction, and the label below each image highlights the object-part name.}
    \label{fig:grasping results}
\end{figure*}

In Tab.~\ref{tab:grasping_tasks}, we list all the tasks~\cite{Li2024ShapeGraspZT} evaluated in our case study in the discussion section. In Fig.~\ref{fig:grasping results}, we showcase some results of our PISA model predicting in a zero-shot manner. It is evident that PISA, trained with our proposed dataset, demonstrates good generalization ability, successfully segmenting unseen parts like plant stems.

It is worth discussing that while the quantitative results shown in the discussion are not superior to ShapeGrasp~\cite{Li2024ShapeGraspZT} and LERF-TOGO~\cite{rashid2023language}, the entire real-world dataset contains only 49 tasks. Although LERF-TOGO achieves 6\% higher accuracy than us, this difference equates to just 3 images. Moreover, our method is significantly faster than others, and this novel end-to-end prediction approach can be beneficial for real-time robot grasping. Our methods can easily be integrated with existing grasping baselines such as GraspNet~\cite{fang2020graspnet}. With our dataset, researchers can focus more on applying segmentation methods to grasping, creating a good bridge between 2D perception and 3D grasping.
\begin{table*}[h]
    \centering
    
    \caption{\textbf{Complete list of tasks for each scene}}
    \begin{tabular}{lc}
    \toprule
         Scene & Tasks  \\
         \midrule
         Kitchen & `pick up the grey spoon', `pick up the teapot', \\
          & `scrub the dishes', `dust the books' \\
         \midrule
         Flowers & `give the daisy', `give the rose'\\
         \midrule
         Mugs & `pick up the mug', `pick up the blue mug', `pick up the grey mug' \\
         & `pick up the white mug', `pick up the teacup'\\
         \midrule
         Tools & `pick up the retractable tape measure', `pick up the screwdriver', `cut the wire'\\
         & `pick up the soldering iron', `swing the hammer'\\
         \midrule
         Knives & `cut the bread', `cut the steak', `cut the box' \\
         \midrule
         Martinis & `pick up the grey martini glass', `pick up the green martini glass' \\
         \midrule 
         Fragile & `hang the camera', `wear the blue sunglasses'\\
         & `wear the black sunglasses', `pick up the lightbulb'\\
         \midrule
         Cords & `pick up the power strip', `plug in the power strip', `pick up the usb dongle',  \\
         & `push in the connector'\\
         \midrule 
         Messy & `eat the ice cream', `eat the lollipop', `eat the red lollipop'\\
         \midrule
         Pasta & `pick up the wine bottle', `uncork the wine', `pick up the corkscrew', \\
         & `pick up the saucepan', `open the saucepan'\\
         \midrule
         Cleaning & `pick up the clorox box', `close the clorox box', `grab a wet towel'\\
         & `pick up the tissue box', `dispense a tissue'\\
         \midrule
         Bottles & `pick up the meyers cleaning spray', `open the meyers cleaning spray', \\ 
         & `spray the meyers cleaning spray', `pick up the purple cleaning spray', \\ 
         & `open the purple cleaning spray', `spray the purple cleaning spray' \\
        \bottomrule
    \end{tabular}
    \label{tab:grasping_tasks}
\end{table*}

% \bibliographystyle{plain}
% \bibliography{main}
\section{Distinctions between InstructPart and LISA.}
\label{appendix: lisa}
While both works fall under the category of \textit{reasoning-based segmentation}, the goal, task definition, benchmark scale, and downstream applicability are fundamentally different:
\begin{itemize}
    \item \textbf{Benchmarking Goals and Granularity:} \\
    LISA focuses primarily on \textit{object-level scene understanding}, where the objective is to semantically interpret an image and segment an object based on abstract instructions (e.g., ``segment the food with the most protein'' or ``segment the food that is not spicy''). In contrast, our work introduces \textit{task-oriented part-level segmentation}, aiming to understand the affordance and functionality of object components. This finer-grained understanding is essential for practical applications that require actionable perception and reasoning grounded in object structure.
    \item \textbf{Benchmark Scale and Usefulness:} \\
    While LISA introduces an important first step toward reasoning-based segmentation, its benchmark contains 1,218 samples, which may be insufficient for a comprehensive evaluation of vision-language models. In contrast, our \textit{InstructPart} benchmark includes 2,400 images, together with 9,600 diverse task instructions, making it more comprehensive and diverse. This enables a more thorough evaluation and offers greater potential for model training and fine-tuning.
    \item \textbf{Novelty and Research Opportunity:} \\
    We consider the reasoning-based segmentation task proposed by LISA as a combination of VQA and semantic segmentation—two tasks that have been well explored. However, \textit{task-oriented part understanding} remains significantly under-explored, as discussed in Section 2.1 of our paper. Our work goes further by introducing the use of instructions and affordances to refer to different object parts. This creates a more challenging and novel setting, which we believe will encourage research into part-level reasoning and grounding.
\end{itemize}

\section{Analysis on Sub-optimal Performance of Existing VLMs on InstructPart}

The sub-optimal performance of state-of-the-art VLMs on our benchmark can be attributed to both the lack of task-relevant training data and limitations in current model architectures for part-level understanding and affordance reasoning.

\begin{itemize}

    \item \textbf{Training Data Limitations:} \\
    Most existing VLMs are not trained with supervision at the part level, nor are they exposed to task-oriented instructions that require grounding specific object components. This leads to a gap in their ability to localize and reason about fine-grained object parts based on functional cues—capabilities that our task explicitly targets. We present two findings to support the claim that current VLMs lack suitable training data:
    \begin{itemize}
        \item In Section 4.5 (Figures 3–5), we show that many VLMs tend to either segment the entire object or miss the correct regions entirely—indicating difficulty in fine-grained localization.
        \item As shown in Appendix D, even simple fine-tuning on our dataset leads to a significant performance boost, suggesting that the models possess latent capability but lack the appropriate supervision signal.
    \end{itemize}

    \item \textbf{Architectural Limitations:} \\
    Most VLMs use a CLIP-based image encoder, which is optimized for object-level semantic understanding and lacks explicit mechanisms for part-level grounding or affordance reasoning. To address this, we incorporate a DINOv2 vision encoder in our baseline, which better captures part-level correspondences across diverse objects (e.g., the handle of a knife vs. the handle of scissors). As a result, our baseline outperforms state-of-the-art VLMs on the proposed task.

\end{itemize}

\section{Justification for Including ORPS}
Referring Expression Segmentation (RES) generally aims to generate segmentation masks from natural language expressions, and our ORPS task can indeed be viewed as a specialized form of RES. However, there are several important distinctions:

\begin{itemize}
    \item Existing RES tasks primarily focus on using expressions to identify entire entities (e.g., ``the woman in the red shirt''). In contrast, ORPS focuses on identifying specific object parts, using a consistent and controlled format: ``\texttt{[part name] of [object name]}''.
    
    \item ORPS can be considered the ``optimal condition'' of TRPS — that is, it strips away complex instruction reasoning and isolates the challenge of part-level visual grounding. This enables us to more precisely understand a model’s bottleneck: is it struggling with language reasoning or with part segmentation?
    
    \item As shown in Table 2, by comparing the performance gap between ORPS and TRPS:
    \begin{itemize}
        \item Reasoning segmentation (RS) methods show a smaller drop in performance from ORPS to TRPS, indicating stronger generalization to complex instructions.
        \item In contrast, Open-Vocabulary Segmentation (OVS) and Referring Expression Segmentation (RES) baselines show a larger drop, highlighting limited ability to handle task-oriented reasoning.
    \end{itemize}
    
    \item This analysis demonstrates that ORPS complements TRPS by offering a controlled setting for part-level grounding, and jointly, they allow us to better characterize the strengths and limitations of different segmentation approaches — especially when comparing models with or without integrated language reasoning.
\end{itemize}

\end{document}